\pdfoutput=1

\documentclass[11pt]{article}

\usepackage[preprint]{acl}

\usepackage{times}
\usepackage{latexsym}

\usepackage[T1]{fontenc}

\usepackage[utf8]{inputenc}

\usepackage{microtype}

\usepackage{inconsolata}

\usepackage{graphicx}
\usepackage{amsfonts}
\usepackage{amsmath}
\usepackage{amsthm}
\usepackage{amssymb}
\usepackage{booktabs}
\usepackage{multirow}
\usepackage{xspace}
\usepackage{caption}
\usepackage{subcaption}
\usepackage[dvipsnames]{xcolor}
\usepackage{float}

\usepackage{cleveref}
\crefname{section}{\S}{\S\S}
\Crefname{section}{\S}{\S\S}
\crefformat{section}{\S#2#1#3}
\crefname{figure}{Fig.}{Fig.}
\crefname{algorithm}{Alg.}{Alg.}
\crefname{line}{line}{lines}
\crefname{appendix}{App.}{}
\crefname{equation}{eq.}{eqs.}
\crefname{table}{Table}{Tables}
\crefname{proposition}{Proposition}{Propositions}
\crefname{assumption}{Assump.}{Assumps.}
\crefname{nenition}{Definition}{Definitions}
\crefname{task}{Task}{Tasks}
\crefname{model}{DNN}{DNNs}
\crefname{causalgraph}{CG}{CGs}
\crefname{thm}{Theorem}{Theorems}

\newcommand{\defn}[1]{\textbf{#1}}

\makeatletter
\DeclareRobustCommand*{\escapeus}[1]{%
  \begingroup\@activeus\scantokens{#1\endinput}\endgroup}
\begingroup\lccode`\~=`\_\relax
   \lowercase{\endgroup\def\@activeus{\catcode`\_=\active \let~\_}}
\makeatother

\usepackage{tabularray}
\usepackage{cellspace}
\usepackage{setspace}

\newcommand{\olmo}{\texttt{OLMo2}\xspace}
\newcommand{\opt}{\texttt{OPT}\xspace}

\newcommand{\olmothirty}{\texttt{OLMo2 32B}\xspace}
\newcommand{\optthirty}{\texttt{OPT 30B}\xspace}

\definecolor{snli}{HTML}{000083}
\definecolor{mnli}{HTML}{8332A8}
\definecolor{anli}{HTML}{00BB00}
\definecolor{rte}{HTML}{009BFE}
\definecolor{hans}{HTML}{CCCC00}
\definecolor{wnli}{HTML}{FE6230}
\definecolor{paws}{HTML}{DD0000}
\definecolor{scitail}{HTML}{873E23}

\newcommand{\colouredsnli}{\textcolor{snli}{\texttt{SNLI}}\xspace}
\newcommand{\colouredmnli}{\textcolor{mnli}{\texttt{MNLI}}\xspace}
\newcommand{\colouredwnli}{\textcolor{wnli}{\texttt{WNLI}}\xspace}
\newcommand{\colouredrte}{\textcolor{rte}{\texttt{RTE}}\xspace}
\newcommand{\colouredscitail}{\textcolor{scitail}{\texttt{SciTail}}\xspace}
\newcommand{\colouredanli}{\textcolor{anli}{\texttt{ANLI}}\xspace}
\newcommand{\colouredhans}{\textcolor{hans}{\texttt{HANS}}\xspace}
\newcommand{\colouredpaws}{\textcolor{paws}{\texttt{PAWS}}\xspace}

\newcommand{\colouredsnlin}{\textcolor{snli}{\texttt{1b.SNLI}}\xspace}
\newcommand{\colouredmnlin}{\textcolor{mnli}{\texttt{1a.MNLI}}\xspace}
\newcommand{\colouredwnlin}{\textcolor{wnli}{\texttt{2.WNLI}}\xspace}
\newcommand{\colouredscitailn}{\textcolor{scitail}{\texttt{3.SciTail}}\xspace}
\newcommand{\colouredrten}{\textcolor{rte}{\texttt{4.RTE}}\xspace}
\newcommand{\colouredhansn}{\textcolor{hans}{\texttt{5.HANS}}\xspace}
\newcommand{\colouredanlin}{\textcolor{anli}{\texttt{6.ANLI}}\xspace}
\newcommand{\colouredpawsn}{\textcolor{paws}{\texttt{7.PAWS}}\xspace}

\title{Do Generalisation Results Generalise?}

\newcommand{\insteth}{1}
\newcommand{\instmila}{2}
\newcommand{\instmcgill}{3}

\author{Matteo Boglioni,$^{\insteth,\instmila}$ ~~~
        Andrea Sgobbi,$^{\insteth}$ ~~~
        Gabriel Tavernini,$^{\insteth}$ ~~~
        Francesco Rita,$^{\insteth}$ \\
        \textbf{
        Marius Mosbach,$^{\instmila,\instmcgill}$ ~~~
        Tiago Pimentel$^{\insteth}$} \\
        ~\\
  $^{\insteth}$ETH Zürich, ~~~
  $^{\instmila}$Mila - Quebec Artificial Intelligence Institute, ~~~
  $^{\instmcgill}$McGill University \\
  \textsf{\{}%
  \href{mailto:mboglioni@ethz.ch}{\textsf{mboglioni}}, 
  \href{mailto:asgobbi@ethz.ch}{\textsf{asgobbi}}, 
  \href{mailto:gtavernini@ethz.ch}{\textsf{gtavernini}},
  \href{mailto:frita01@ethz.ch}{\textsf{frita01}}%
  \textsf{\}@ethz.ch}, \\
  \href{mailto:marius.mosbach@mila.quebec}{\textsf{marius.mosbach}}\textsf{@mila.quebec},
  \href{mailto:tiago.pimentel@inf.ethz.ch}{\textsf{tiago.pimentel}}\textsf{@inf.ethz.ch}
  }

\begin{document}
\maketitle

\begin{abstract}
A large language model's (LLM's) out-of-distribution (OOD) generalisation ability is crucial to its deployment.
Previous work assessing LLMs' generalisation performance, however, typically focuses on a single out-of-distribution dataset.
This approach may fail to precisely evaluate the capabilities of the model, as the data shifts encountered once a model is deployed are much more diverse.
In this work, we investigate whether OOD generalisation results generalise.
More specifically, we evaluate a model's performance across multiple OOD testsets throughout a finetuning run; we then evaluate the partial correlation of performances across these testsets, regressing out in-domain performance.
This allows us to assess how correlated are generalisation performances once in-domain performance is controlled for.
Analysing \olmo and \opt, we observe no overarching trend in generalisation results: the existence of a positive or negative correlation between any two OOD testsets depends strongly on the specific choice of model analysed.
\end{abstract}

\section{Introduction}

A large language model's (LLM's) out-of-distribution (OOD) generalisation\footnote{Throughout this work, generalisation always refers to \emph{out-of-distribution} generalisation.} performance is an essential property for its deployment in the wild. 
Not surprisingly, it has received increased attention from the community \citep{xu2021raise, Hupkes_2023, yang2023glue, yang-etal-2024-unveiling, yang2025evaluating, yuan2023revisiting, ye-2024-cross}.
Most work evaluating generalisation, however, relies on a single out-of-distribution testset per task \citep{mosbach2023few, joshi-he-2022-investigation, bhargava-etal-2021-generalization}.\footnote{Two notable exceptions are discussed in \cref{sec:related}.}

When a model achieves high scores on such an out-of-distribution testset, authors typically assume the model has found a good solution for the task and that the model does not rely on spurious features to solve it.
There is, however, no \textit{a priori} reason why a model which generalises in one OOD testset should also generalise in testsets created under different distribution shifts.
Furthermore, \citet{mosbach2023few} show that generalisation performance can be quite unstable across training, reinforcing the need for its more precise assessment.

 \begin{figure}[t]
  \centering
  \begin{subfigure}[b]{0.47\textwidth}
    \centering
    \includegraphics[width=\textwidth]{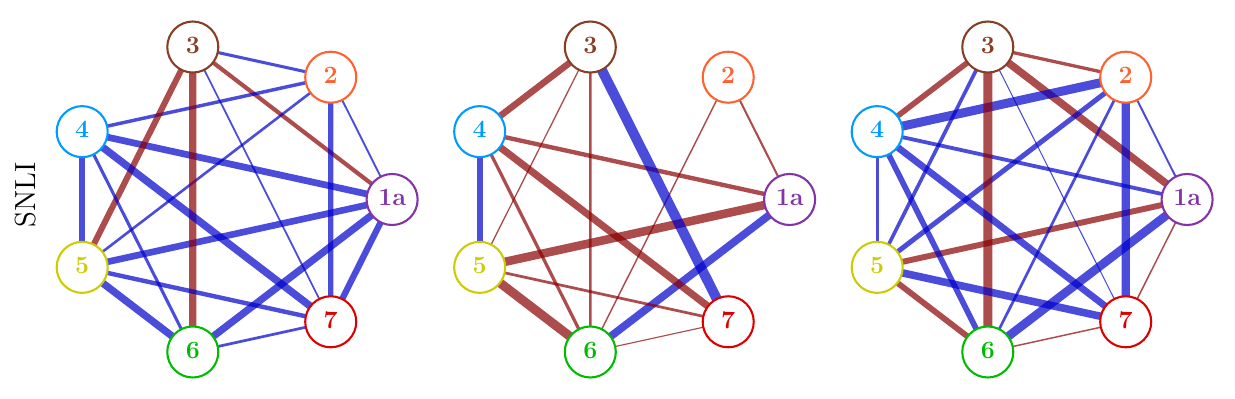}
  \end{subfigure}
  
  \begin{subfigure}[b]{0.47\textwidth}
    \centering
    \includegraphics[width=\textwidth]{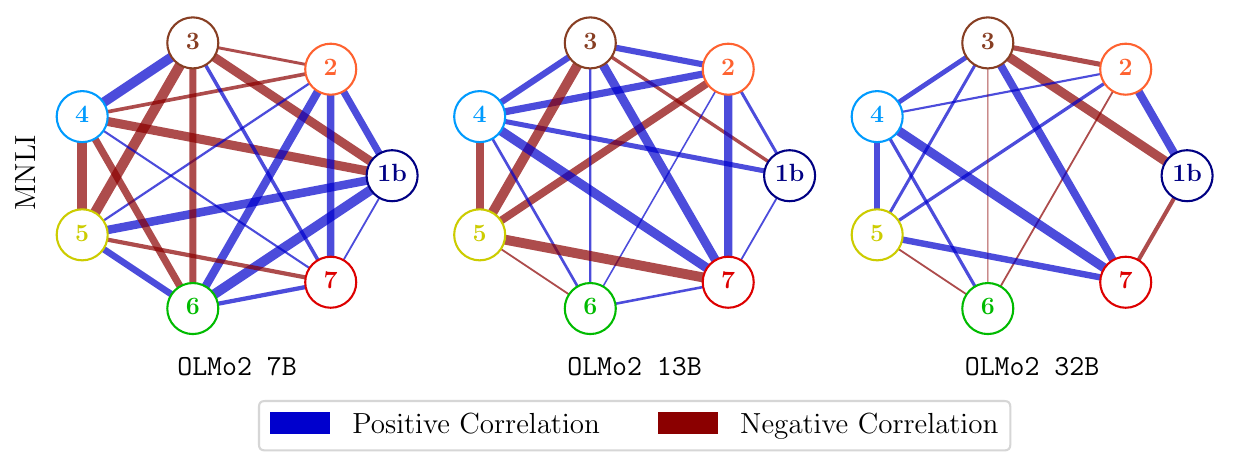}
  \end{subfigure}
  \vspace{-5pt}
  \caption{\olmo's partial OOD correlations on SNLI (top) and MNLI (bottom). No clear trends are observed. Edge thickness increases with absolute correlation value. Legend: \colouredmnlin, \colouredsnlin, \colouredwnlin, \colouredscitailn, \colouredrten, \colouredhansn, \colouredanlin, and \colouredpawsn.}
  \label{fig:intro_graph}
  \vspace{-5pt}
\end{figure}

In this paper, we investigate whether generalisation results generalise.
To this end, we analyse how a model's generalisation performances in different OOD testsets correlate across a single finetuning run.
However, generalisation performances are bound to be trivially correlated due to their dependency on a common factor: the model's in-domain performance.
We control for that factor by computing \defn{partial OOD correlations} instead, regressing out in-domain performance.
These partial correlations quantify how strongly generalisation performances correlate beyond their dependence on in-domain performance.\looseness=-1

Empirically, we show that whether generalisation performance will transfer across OOD testsets is a complex phenomenon. 
While a specific model's generalisation performance may be strongly correlated across two OOD testsets, it might present negative correlations under another pair.
Similarly, while two OOD testsets may present positive partial correlations under one model, they may present negative correlations under another.
This large variance in generalisation performance highlights that fair generalisation evaluation must span multiple OOD testsets.

\section{Out-of-distribution Generalisation in Language Models}
\label{sec:related}

Robust generalisation beyond the training distribution has long been a challenge in natural language processing (NLP).
In the quest to improve OOD generalisation, researchers face an important problem: how do we evaluate it in the first place?

\paragraph{How to evaluate OOD generalisation?}
Assessing generalisation performance is an intricate game of cat-and-mouse: as models tend to saturate on existing benchmarks, new ones are released to expose new weaknesses.
\citet{mccoy2019right}, for instance, adversarially constructed an OOD testset (HANS) to reveal LLMs' reliance on superficial cues to solve a natural language inference (NLI) task.
Similarly, \citet{nie2020adversarial} constructed an OOD dataset (ANLI) in rounds to continually fool NLI models. 
Further, \citet{liu-etal-2022-wanli} used models trained on MNLI \cite{williams-etal-2018-broad} to generate their own synthetic (adversarial) datasets. 

\paragraph{Do finetuned models generalise?}
Given all these benchmarks, we should have a good idea about how well language models generalise.
However, the effect of  finetuning on a model's OOD generalisation remains a little-understood topic.
\citet{kumar2022finetuningdistortpretrainedfeatures}, for instance, show that finetuning models with randomly initialised classifier heads can lead to distorted features and hence poor generalisation.
Recent empirical work \citep{mosbach2023few,yang-etal-2024-unveiling}, however, show strong generalisation of finetuned models on OOD data.
Both these works, however, use pattern-based fine-tuning \citep{schick-schutze-2021-exploiting} instead of a randomly initialised classifier head, being thus not directly comparable to the findings of \citeauthor{kumar2022finetuningdistortpretrainedfeatures}.
A few prior works investigate multiple OOD testsets.
E.g., \citet{gupta2024whispers} evaluate which OOD testsets still represent a challenge for finetuned models.
Closest to our work is \citet{sun-etal-2023-validity}, who compare the rankings achieved by finetuned models on a number of OOD testsets.
More specifically, they compute the correlations across OOD testsets of the rankings achieved by several pretrained models when finetuned to perform NLI.
Their analyses, however, do not control for either the models' in-domain performance, or the used pretrained models' size and quality.
Instead, we will analyse partial correlations within each training run, controlling for both factors.

\section{Measuring Correlations between OOD Generalisations}
\label{sec:methodology}

We aim to assess how robust generalisation results are to a specific choice of OOD testset.
We can quantify this by analysing how correlated generalisation results are across testsets.
Language models with better in-domain performance, however, are also likely to perform better out-of-domain \citep{yang2023glue}. 
Naively computing OOD correlations, thus, is likely to mostly capture this trivial (and arguably uninteresting) source of correlation.
To control for this, we measure \defn{partial OOD correlations} instead: the correlation between two OOD performances once in-domain performance has been regressed out.
How does this work in practice?

\newcommand{\btheta}{\boldsymbol{\theta}}
\newcommand{\ptheta}{p_{\btheta}}
\newcommand{\score}{s}
\newcommand{\scores}{\mathbf{s}}
\newcommand{\scoreid}{\score^{\mathtt{ind}}}
\newcommand{\scoresid}{\scores^{\mathtt{ind}}}
\newcommand{\datasetood}{d}
\newcommand{\scoreoodbase}{\score}
\newcommand{\scoresoodbase}{\scores}
\newcommand{\scoreood}{\scoreoodbase^{\mathtt{ood}:\datasetood}}
\newcommand{\scoresood}{\scoresoodbase^{\mathtt{ood}:\datasetood}}
\newcommand{\scoresoodone}{\scoresoodbase^{\mathtt{ood}:\datasetood_1}}
\newcommand{\scoresoodtwo}{\scoresoodbase^{\mathtt{ood}:\datasetood_2}}
\newcommand{\correl}{\mathtt{corr}}
\newcommand{\scorecorr}{\rho}
\newcommand{\regressor}{f^{\datasetood}}
\newcommand{\R}{\mathbb{R}}
\newcommand{\residual}{e^{\datasetood}}
\newcommand{\residualsbase}{\mathbf{e}}
\newcommand{\residuals}{\residualsbase^{\datasetood}}
\newcommand{\residualsone}{\residualsbase^{\datasetood_1}}
\newcommand{\residualstwo}{\residualsbase^{\datasetood_2}}
\newcommand{\cov}{\mathtt{Cov}}

\begin{table*}[ht]
    \centering
    \resizebox{\textwidth}{!}{
        \begin{tabular}{lccccccccc}
        \toprule
        && \multicolumn{8}{c}{\textbf{\colouredmnli}} \\
        \cmidrule(lr){3-10}
        \textbf{Model} & \textbf{Size} & \textbf{\colouredmnli}$^{\ddagger}$ & \textbf{\colouredsnli} & \textbf{\colouredwnli}& \textbf{\colouredscitail} & \textbf{\colouredrte} & \textbf{\colouredhans} & \textbf{\colouredanli} & \textbf{\colouredpaws} \\
\midrule
\multirow{4}{*}{\texttt{OPT}}
& \multicolumn{1}{c|}{2.7b} & \textit{81.6 $\pm$ 5.9} & 72.7 $\pm$ 17.8 & 49.9 $\pm$ 0.7 & 65.8 $\pm$ 5.9 & 62.5 $\pm$ 2.1 & 51.7 $\pm$ 2.9 & 50.5 $\pm$ 0.5 & 46.3 $\pm$ 1.2 \\
& \multicolumn{1}{c|}{6.7b} & \textit{84.7 $\pm$ 6.7} & 83.7 $\pm$ 12.6 & 50.7 $\pm$ 1.8 & 70.7 $\pm$ 10.9 & 64.3 $\pm$ 1.0 & 55.5 $\pm$ 7.8 & 49.2 $\pm$ 1.9 & 47.3 $\pm$ 0.8 \\
& \multicolumn{1}{c|}{13b} & \textit{87.3 $\pm$ 5.6} & 83.9 $\pm$ 13.4 & 50.9 $\pm$ 2.6 & 71.3 $\pm$ 7.6 & 67.9 $\pm$ 1.6 & 57.0 $\pm$ 7.3 & 52.3 $\pm$ 1.5 & 48.5 $\pm$ 2.5 \\
& \multicolumn{1}{c|}{30b} & \textit{\textbf{89.0 $\pm$ 5.9}} & \textbf{86.8 $\pm$ 14.2} & 50.7 $\pm$ 1.1 & \textbf{74.7 $\pm$ 3.2} & \textbf{71.2 $\pm$ 3.2} & 59.3 $\pm$ 6.2 & 53.0 $\pm$ 1.0 & 48.6 $\pm$ 1.5 \\
\midrule
\multirow{3}{*}{\texttt{OLMo2}}
& \multicolumn{1}{c|}{7B} & \textit{75.1 $\pm$ 1.9} & 67.1 $\pm$ 15.8 & 55.7 $\pm$ 2.6 & 55.0 $\pm$ 8.1 & 63.1 $\pm$ 5.6 & 52.7 $\pm$ 2.2 & 55.7 $\pm$ 4.7 & 59.1 $\pm$ 5.9 \\
& \multicolumn{1}{c|}{13B} & \textit{61.2 $\pm$ 1.5} & 56.7 $\pm$ 5.9 & 52.4 $\pm$ 1.0 & 57.0 $\pm$ 4.8 & 54.0 $\pm$ 1.9 & 51.3 $\pm$ 2.5 & 50.9 $\pm$ 0.5 & 54.7 $\pm$ 2.5 \\
& \multicolumn{1}{c|}{32B} & \textit{87.4 $\pm$ 12.0} & 81.8 $\pm$ 24.8 & \textbf{68.2 $\pm$ 12.5} & 53.7 $\pm$ 9.5 & 69.2 $\pm$ 4.0 & \textbf{61.5 $\pm$ 7.4} & \textbf{68.3 $\pm$ 5.4} & \textbf{66.9 $\pm$ 2.6} \\
\midrule
\multicolumn{2}{l}{\texttt{Chance performance}} & 50.0& 50.0& 50.0& 50.0& 50.0&50.0 & 50.0 & 50.0 \\
        \bottomrule
        \end{tabular}
    }
    \vspace{-3pt}
    \caption{Accuracy on each OOD dataset for models trained on MNLI with 128 examples over 3 independent runs. Measurements are taken using the checkpoint with the highest in-domain performance. $^{\ddagger}$ in-domain dataset.}
    \label{tab:acc_nli_mnli}
    \vspace{-5pt}
\end{table*}

\begin{table*}[ht]
    \centering
    \resizebox{\textwidth}{!}{
        \begin{tabular}{lccccccccc}
        \toprule
        && \multicolumn{8}{c}{\textbf{\colouredsnli}} \\
        \cmidrule(lr){3-10}
        \textbf{Model} & \textbf{Size} &
        \textbf{\colouredsnli}$^{\ddagger}$ & \textbf{\colouredmnli} & \textbf{\colouredwnli}& \textbf{\colouredscitail} & \textbf{\colouredrte} & \textbf{\colouredhans} & \textbf{\colouredanli} & \textbf{\colouredpaws} \\
\midrule
\multirow{4}{*}{\texttt{OPT}}
& \multicolumn{1}{c|}{2.7b} & \textit{94.2 $\pm$ 0.2}& 78.6 $\pm$ 3.1 & 50.4 $\pm$ 0.5 & 74.1 $\pm$ 2.8 & 66.4 $\pm$ 0.7 & 51.4 $\pm$ 1.4 & 50.6 $\pm$ 1.1 & 50.6 $\pm$ 3.8 \\
& \multicolumn{1}{c|}{6.7b} & \textit{94.3 $\pm$ 1.3}& 78.2 $\pm$ 6.4 & 52.2 $\pm$ 0.6 & 68.8 $\pm$ 13.3 & 66.0 $\pm$ 0.9 & 54.9 $\pm$ 2.4 & 51.8 $\pm$ 3.1 & 49.5 $\pm$ 2.3 \\
& \multicolumn{1}{c|}{13b} & \textit{95.3 $\pm$ 0.4}& 82.0 $\pm$ 4.0 & 49.9 $\pm$ 0.4 & 70.2 $\pm$ 4.2 & 65.3 $\pm$ 1.3 & 53.4 $\pm$ 3.0 & 51.8 $\pm$ 1.0 & 48.6 $\pm$ 2.3 \\
& \multicolumn{1}{c|}{30b} & \textit{96.1 $\pm$ 0.1}& \textbf{86.0 $\pm$ 3.9} & 52.0 $\pm$ 1.4 & \textbf{76.8 $\pm$ 2.3} & 70.7 $\pm$ 2.5 & 58.8 $\pm$ 5.6 & 53.0 $\pm$ 1.3 & 49.7 $\pm$ 4.1 \\
\midrule
\multirow{3}{*}{\texttt{OLMo2}}
& \multicolumn{1}{c|}{7B} & \textit{90.6 $\pm$ 5.0}& 70.7 $\pm$ 11.3 & 59.3 $\pm$ 3.4 & 56.9 $\pm$ 4.3 & 61.0 $\pm$ 4.3 & 61.8 $\pm$ 3.1 & 56.3 $\pm$ 3.7 & 64.6 $\pm$ 3.9 \\
& \multicolumn{1}{c|}{13B} & \textit{80.4 $\pm$ 2.3}& 61.4 $\pm$ 2.8 & 54.9 $\pm$ 0.2 & 54.7 $\pm$ 2.7 & 55.8 $\pm$ 0.7 & 57.3 $\pm$ 3.2 & 52.3 $\pm$ 1.9 & 54.8 $\pm$ 1.7 \\
& \multicolumn{1}{c|}{32B} & \textit{\textbf{98.0 $\pm$ 0.1}} & 84.1 $\pm$ 5.1 & \textbf{73.9 $\pm$ 1.0} & 60.7 $\pm$ 5.8 & \textbf{70.9 $\pm$ 1.2} & \textbf{66.7 $\pm$ 2.7} & \textbf{65.4 $\pm$ 2.6} & \textbf{69.6 $\pm$ 0.9} \\
\midrule
\multicolumn{2}{l}{\texttt{Chance performance}} & 50.0& 50.0& 50.0& 50.0& 50.0&50.0 & 50.0 & 50.0 \\
        \bottomrule
        \end{tabular}
    }
    \vspace{-3pt}
    \caption{Accuracy on each OOD dataset for models trained on SNLI with 128 examples over 3 independent runs. Measurements are taken using the checkpoint with the highest in-domain performance. $^{\ddagger}$ in-domain dataset.}
    \label{tab:acc_nli_snli}
    \vspace{-5pt}
\end{table*}

Let $\ptheta$ be a language model, which we finetune on a specific (in-domain) training set.
While fine-tuning this model, we measure its \defn{in-domain performance}, denoted $\scoreid_t$, on an in-domain testset at several checkpoints $t$; this gives us a vector of performances $\scoresid$.
Simultaneously, we measure this model's \defn{out-of-domain performance}, denoted $\scoreood_t$, on several out-of-domain testsets, $\datasetood$, using the same checkpoints; getting a vector of performances $\scoresood$.
If we simply wanted to examine the correlation between OOD performances, we would evaluate: $\correl(\scoresoodone, \scoresoodtwo)$.

Computing the partial correlation between two OOD datasets, however, requires regressing out in-domain performance.
To do this, we train a regression model $\regressor:\R \rightarrow \R$ for each OOD dataset $\datasetood$, which, given an in-domain performance measurement, predicts that checkpoints' OOD performance: $\scoreood_t \approx \regressor(\scoreid_t)$.
Given this model, 
we compute a residual
$\residual_t = \scoreood_t - \regressor(\scoreid_t)$, which quantifies how much better or worse a model performs on $\datasetood$ than what would be expected given its in-domain performance.
Doing this for all checkpoint steps $t$, we get a vector of residuals $\residuals$.
Finally, we compute the \defn{partial correlations} we are interested in as:\looseness=-1
\begin{align}
    \scorecorr^{\datasetood_1, \datasetood_2} &= \correl(\residualsone, \residualstwo)~.
\end{align}
Since our focus is on observing joint improvement, we choose to capture simple linear correlations between these residuals, measuring Pearson's correlations as the $\correl(\cdot)$ function.
Throughout the paper, we present results using GAM regressors $\regressor$.\footnote{We experiment with both linear and GAM \cite{hastie1990gam} regressors, but find this choice has only a minor impact on results. \cref{fig:ood_regression_full_128shots} shows in-domain vs.\ out-of-domain curves learned by our regressors. We place partial correlation results using linear regressors in \cref{fig:corr_linear_4x4_128,fig:corr_linear_4x4_64,fig:corr_linear_4x4_32} in \cref{app:detailed_results}.\looseness=-1}

\section{Experimental Setup}

\paragraph{Task.}
As a test-bed for our experiments, we focus on natural language inference \citep[NLI;][]{dagan2004probabilistic, putra2024recognizing}, as generalisation performance on this task has received considerable interest \citep{bhargava-etal-2021-generalization, zhou2021investigatingeffectnaturallanguage,mosbach2023few,gupta2024whispers}.
This task consists in determining the logical relationship between a pair of sentences.
More specifically, each entry in this task consists of a pair of sentences, a premise and a hypothesis; the task is then to determine if the premise entails, contradicts, or is neutral about the hypothesis.

\begin{figure}
    \centering
    \includegraphics[width=\linewidth]{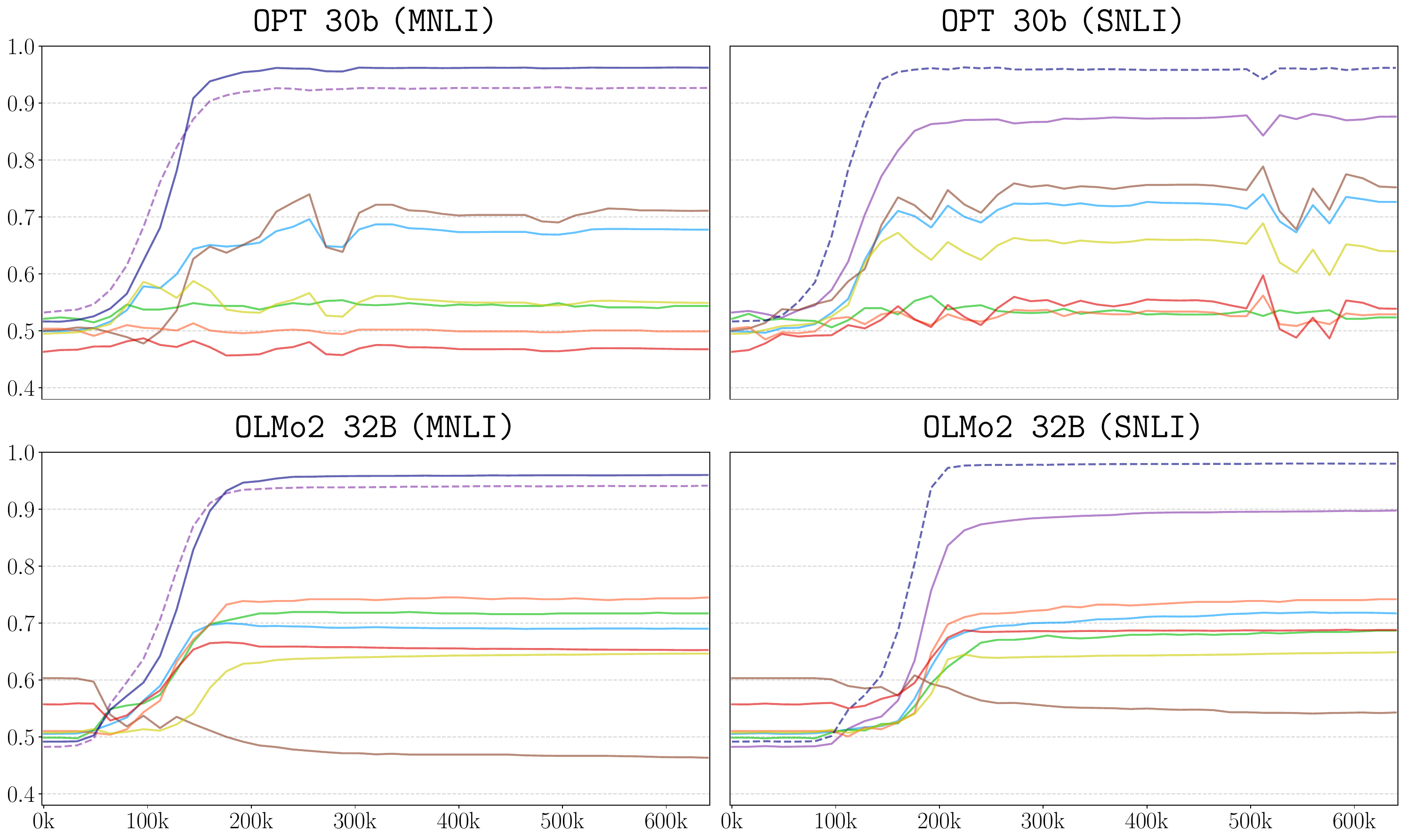}
    \vspace{-15pt}
    \caption{Accuracy ($y$-axis) across training steps ($x$-axis)  of \opt (top) and \olmo (bottom) for a single finetuning run on \colouredmnli (left) and \colouredsnli (right). Legend: \colouredmnli, \colouredsnli, \colouredwnli, \colouredrte, \colouredscitail, \colouredanli, \colouredhans and \colouredpaws. \looseness=-1}
    \label{fig:odd_performance_across_train}
    \vspace{-5pt}
\end{figure}

\begin{figure*}[t]
    \centering
    \includegraphics[width=\textwidth]{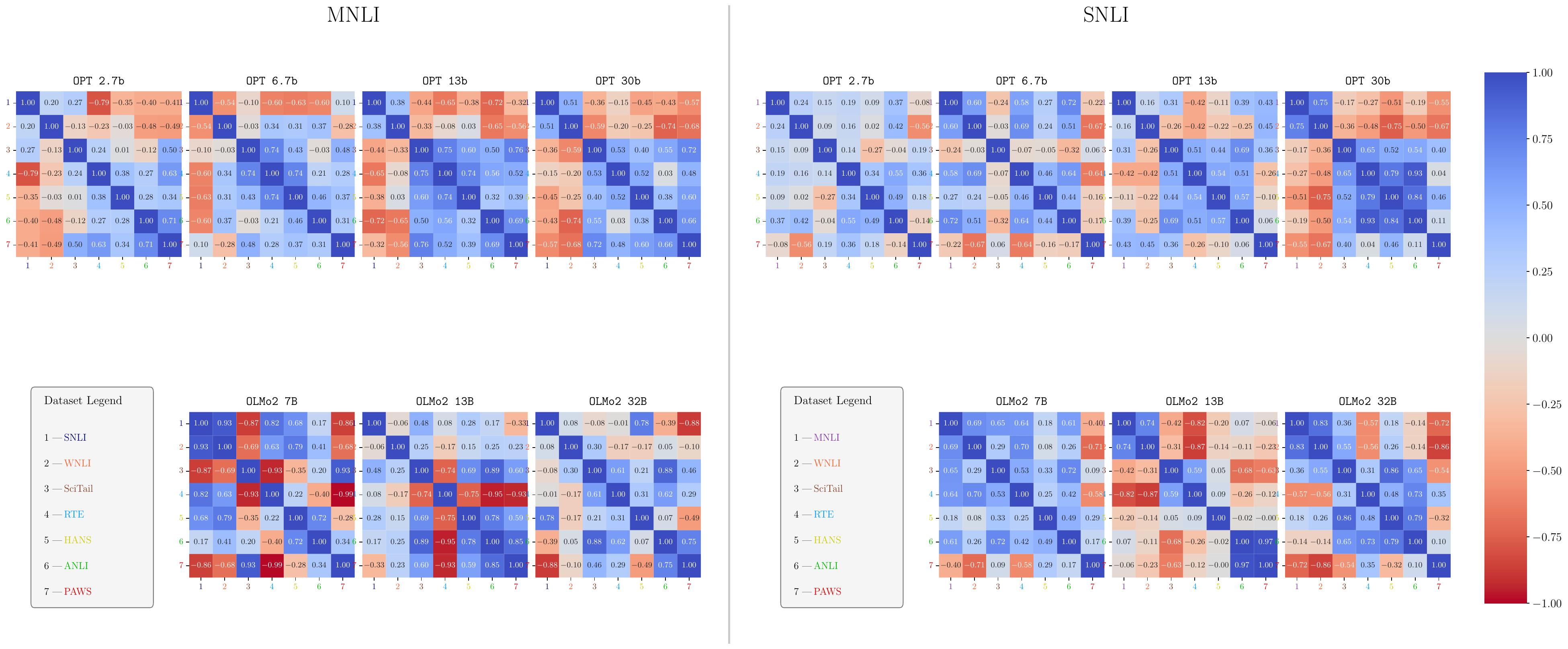}
    \vspace{-20pt}
    \caption{Partial correlations of \opt (top) and \olmo (bottom) across model sizes (ordered from left to right) trained on MNLI (left) and SNLI (right). All these correlations are obtained by fitting a GAM regressor over 3 independent training runs. See a larger version of this plot in \cref{fig:corr_gam_4x4_128} (in \cref{app:detailed_results}).}
    \label{fig:corr_gam_2x8_128}
    \vspace{-8pt}
\end{figure*}

\paragraph{Models.}
We rely here on two different model families: \opt \citep{zhang2022optopenpretrainedtransformer}, \olmo \citep{olmo20252olmo2furious}. 
We choose these models due to them being publicly available in multiple sizes, and due to their popularity in recent years for a broad range of NLP tasks.
Beyond that, \opt also makes our experiments more easily comparable to previous work \citep[e.g.,][]{mosbach2023few, srinivasan2024comparative}.
Following \citet{mosbach2023few}, we finetune these models using: a few-shot setting, with 128, 64 and 32 examples; low-rank adaptation \citep[LoRA;][]{hulora};
and pattern-based finetuning
\citep{schick-schutze-2021-exploiting, gao-etal-2021-making}, reusing the pre-trained LM head instead of using a randomly initialised one.
More details can be found in \cref{sec:pbftappendix}.

\paragraph{Data.}
For our experiments, we selected 8 different NLI datasets: \textbf{SNLI} \citep{bowman-etal-2015-large}, \textbf{MNLI} \citep{williams-etal-2018-broad}, \textbf{SciTail} \citep{scitail}, \textbf{WNLI} and \textbf{RTE} \citep{wang2018glue}, \textbf{PAWS} \cite{zhang2019paws}, \textbf{HANS} \citep{mccoy2019right}, \textbf{ANLI} \citep{nie2020adversarial}.
We run experiments while finetuning our models on either SNLI or MNLI, making that our in-domain dataset---and evaluate our model on the 7 other OOD datasets.
Details about the selected datasets can be found in \cref{sec:nliappendix}.

\section{Results}

\vspace{-1pt}
\paragraph{Finetuned models tend to generalise, but not everywhere.}
\Cref{tab:acc_nli_mnli,tab:acc_nli_snli} present the generalisation of our evaluated models across all analysed testsets.
These tables present performances for a single checkpoint per model, where checkpoints were selected based on having the best in-domain performance.
The tables show that no testset seems to be challenging for all models: every testset has at least one model that generalises successfully.
Furthermore, it also shows that finetuning produces models that often perform well across a range of OOD testsets.
However, for any given model, there is always at least one testset at which they underperform.
For instance, the same \optthirty checkpoints achieve 86.0\% accuracy on MNLI, but 49.7\% on PAWS.
This variability highlights a key limitation of single-testset evaluations.
Additionally, na\"ively looking at \cref{tab:acc_nli_mnli,tab:acc_nli_snli} might lead one to conclude that generalisation results are mostly robust: \optthirty trained on MNLI does better than all other models in most testsets, and both \optthirty and \olmothirty seem to consistently beat other models when trained on SNLI.
This conclusion, however, is not necessarily warranted, as both model size and in-domain performance act as strong confounders.
We now look at how the generalisation performance of each of these models fluctuates throughout training, as a way to control for the effect of model size on results.\looseness=-1

\vspace{-1pt}
\paragraph{OPT's generalisation performance oscillates, but OLMo2's doesn't.}
\Cref{fig:odd_performance_across_train} presents \optthirty's and \olmothirty's OOD generalisation performances across training.
(Results for smaller \opt and \olmo models are in \cref{fig:odd_performance_across_train_full_128shots}, in \cref{app:detailed_results}.)
Overall, these figures reproduce one of the key results in \citet{mosbach2023few}, showing that \opt's generalisation performance is unstable throughout training, presenting large (mostly unpredictable) oscillations.
Interestingly, \olmo's performance does not present the same oscillations.
Perhaps more important for our research question though, we see in this figure that generalisation in some OOD testsets seems to track the others; this is most obvious for the results of \optthirty trained on SNLI.
In-domain performance, however, also tracks OOD generalisation in these results---at least to some extent.
Next, we thus move to analysing partial correlations as introduced in \cref{sec:methodology}.

\begin{figure}
    \centering
    \includegraphics[width=1\linewidth]{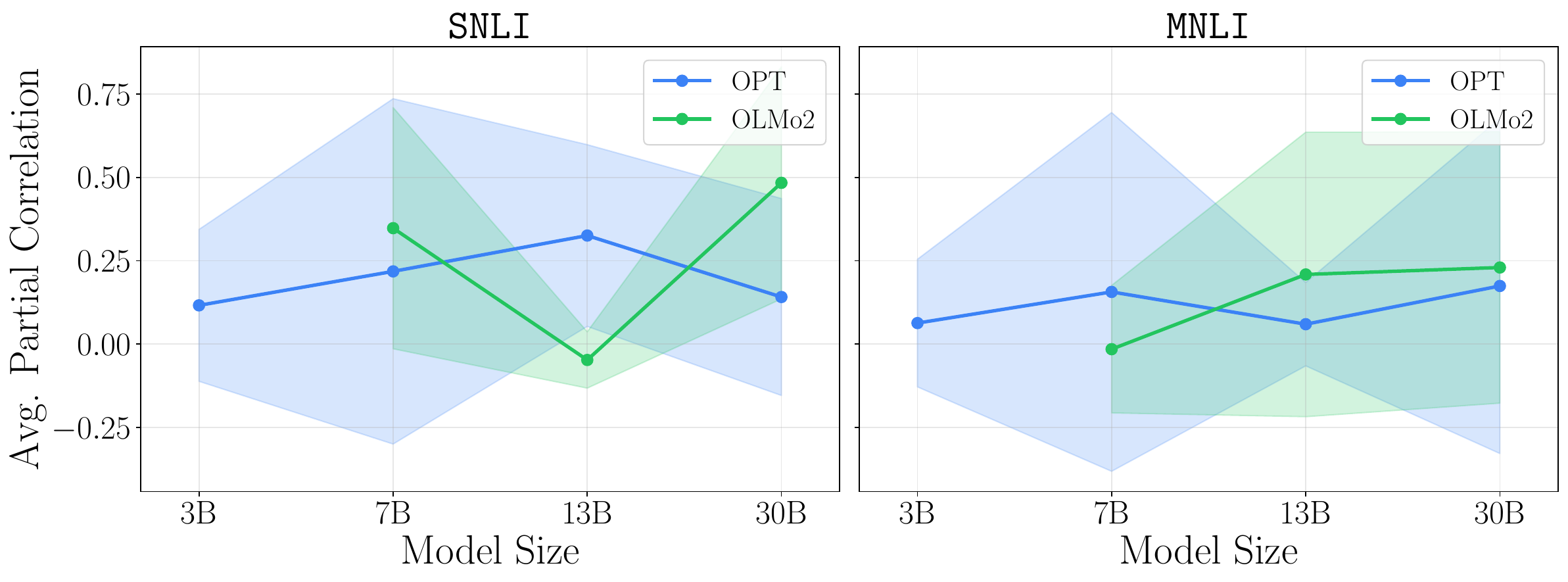}
    \caption{Partial correlations averaged across all OOD testset pairs for \opt and \olmo with different sizes.}
    \label{fig:average_correlations}
\end{figure}

\vspace{-1pt}
\paragraph{Generalisation's generalisation is complicated.}
\Cref{fig:corr_gam_2x8_128} presents the partial correlations across OOD testsets for both \opt and \olmo models.
In this figure, we observe that OOD generalisation is a highly complex property for which no clear trend emerges across testsets.
While for a model two OOD testsets might present strong postive partial correlations, for another model this correlation might be negative.
Additional intuition can also be drawn from \cref{fig:average_correlations}, which shows that partial correlations do not seem to strengthen with model size or with a particular choice of training dataset; partial correlations for models finetuned on \colouredmnli\ do not differ substantially from their corresponding \colouredsnli\ counterparts.
These findings underscore the importance of conducting a comprehensive evaluation when making claims about a model’s generalisation capabilities, an often-lacking aspect in the current literature.\looseness=-1

\section{Conclusions}

Our results highlight the need for generalisation research to rely on several OOD testsets to ensure fair evaluations. 
We do not observe clear trends when studying testset-to-testset performance correlations: no clear trends arise when comparing different training datasets, model families or sizes.
In fact, the partial correlation of performances on a pair of OOD testsets seems to not be an intrinsic property even of the testset pair itself, depending on the specific model and training dataset considered.

\section*{Limitations}

Due to limited compute resources, it was impractical for us to include models larger than 30B parameters in our analysis.  
However, it would be interesting to investigate if the inconsistent trends observed here would carry to other model families and to larger sizes.
Additionally, we are not sure if our studied models (\opt, \olmo) were exposed to the analysed testsets during pretraining.\footnote{Although \olmo is trained on open datasets, directly testing for contamination on such a big dataset was a bigger challenge than anticipated, with most solutions relying on massive indexes for lookups \citep{vu2023koala}.}
We conducted preliminary experiments using \texttt{Min-k\%++} \citep{zhang2025minkimprovedbaselinedetecting} and \texttt{Time Travel in LLMs} \citep{golchin2024timetravelllmstracing} to investigate such data contamination, but these experiments were inconclusive in most cases.
Despite the negative results, though, our tests suggest that the models have not outright memorised the OOD testsets, which would allow them to trivialise the task. %
Finally, our experiments focus exclusively on NLI. 
This limitation results from the lack of dedicated OOD testsets for other tasks, making it difficult to study the extent to which our findings are NLI-specific.\looseness=-1

\bibliography{custom}

\newpage
\appendix

\section{Pattern-based finetuning details}
\label{sec:pbftappendix}
Pattern-based finetuning requires us to specify an input pattern and define a mapping between the answer tokens and the actual labels \citep{schick-etal-2020-automatically}. 
Our experiments with NLI use the following pattern:\looseness=-1
\begin{quote}
    \texttt{\{premise\} Question: \{hypothesis\} Yes or No?} 
\end{quote}
The target tokens we consider are, respectively, \texttt{`\_Yes'} for entailment and \texttt{`\_No'} otherwise.\footnote{The underscores indicate a whitespace in the token. This is important to guarantee that the correct token-string representing this character-string is considered \citep{pimentel-meister-2024-compute,phan-etal-2024-understanding}, which may impact prompting performance \citep{sanz-guerrero-etal-2025-mind}.}

\section{NLI Datasets}
\label{sec:nliappendix}
 We use 2 main large-scale datasets for finetuning the models:
\textbf{SNLI} \citep{bowman-etal-2015-large}, which contains $570$K crowdsourced sentence-pairs based on image captions;
and \textbf{MNLI} \citep{williams-etal-2018-broad}, which is a set of $433$K sentence-pairs meant to cover a large range of genres of spoken and written text. 
Compared to SNLI, MNLI offers more linguistic diversity and difficulty as
it includes representative samples from 10 distinct genres of written
and spoken English.\looseness=-1
We assessed the generalisation capacity of fine-tuned models using 6 NLI testsets. 
These comprise 3 adversarial datasets---designed especially to evaluate the models' robustness to heuristics---%
as well as 3 more standard NLI datasets with various but comparable input distributions:
\begin{itemize}
    \item \textit{Standard}: \textbf{SciTail} \citep{scitail} is based on science multiple-choice exams, \textbf{WNLI} focuses on identifying the referent of a certain pronoun and \textbf{RTE} is a general entailment dataset. These last two are a part of the GLUE Benchmark \citep{wang2018glue}.
    \item \textit{Adversarial}: \textbf{PAWS} \citep{zhang2019paws} uses paraphrase adversaries, \textbf{HANS} \citep{mccoy2019right} tackles failure cases of $3$ simple heuristics and \textbf{ANLI} \citep{nie2020adversarial} finds adversaries via human feedback.
\end{itemize}
To avoid inconsistencies that can result from different annotation policies among datasets, we removed the neutral-labeled samples, enabling us to more effectively separate the impacts of domain shifts on model performance, and guaranteeing a more consistent assessment framework.

\section{Resource Usage}

We ran our experiments on various machines, depending on memory requirements. Small models were trained on 4x A5000 GPUs (with 24GB each), larger models were trained using 8x A6000 (with 48GB each) or 4x A100 (with 80GB). 
The total runtime for all the experiments presented here is of $5{,}500$ GPU hours.

\onecolumn

\section{Detailed Results}
\label{app:detailed_results}

\subsection{Average Performance for other Few-shot Settings}

\begin{table}[H]
    \centering
    \resizebox{\textwidth}{!}{
        \begin{tabular}{lccccccccc}
        \toprule
        && \multicolumn{8}{c}{\textbf{\colouredmnli}} \\
        \cmidrule(lr){3-10}
        \textbf{Model} & \textbf{Size} & \textbf{\colouredmnli}$^{\ddagger}$ & \textbf{\colouredsnli} & \textbf{\colouredwnli}& \textbf{\colouredscitail} & \textbf{\colouredrte} & \textbf{\colouredhans} & \textbf{\colouredanli} & \textbf{\colouredpaws} \\
\midrule
\multirow{3}{*}{\texttt{OPT}}
& \multicolumn{1}{c|}{2.7b} & \textit{67.2 $\pm$ 2.9} & 59.4 $\pm$ 7.7 & 50.9 $\pm$ 0.3 & 59.6 $\pm$ 6.0 & 54.5 $\pm$ 2.1 & 51.7 $\pm$ 1.5 & 50.0 $\pm$ 0.9 & 48.3 $\pm$ 3.7 \\
& \multicolumn{1}{c|}{6.7b} & \textit{74.4 $\pm$ 6.6} & 66.7 $\pm$ 15.0 & 50.3 $\pm$ 1.2 & 63.8 $\pm$ 4.0 & 57.5 $\pm$ 4.7 & 54.5 $\pm$ 3.5 & 50.6 $\pm$ 0.6 & 50.6 $\pm$ 4.7 \\
& \multicolumn{1}{c|}{13b} & \textit{79.7 $\pm$ 8.8} & 75.6 $\pm$ 22.6 & 51.0 $\pm$ 1.7 & \textbf{73.9 $\pm$ 3.1} & 63.5 $\pm$ 4.2 & 55.0 $\pm$ 2.5 & 50.1 $\pm$ 2.2 & 50.7 $\pm$ 3.1 \\
& \multicolumn{1}{c|}{30b} & \textit{\textbf{82.9 $\pm$ 8.7}} & 75.0 $\pm$ 23.0 & 51.8 $\pm$ 2.1 & 63.1 $\pm$ 7.5 & 62.0 $\pm$ 2.4 & 57.5 $\pm$ 1.7 & 52.9 $\pm$ 1.7 & 48.7 $\pm$ 2.8 \\
\midrule
\multirow{3}{*}{\texttt{OLMo2}}
& \multicolumn{1}{c|}{7B} & \textit{59.9 $\pm$ 3.2} & 54.0 $\pm$ 3.8 & 50.5 $\pm$ 0.1 & 52.1 $\pm$ 4.8 & 51.9 $\pm$ 1.5 & 51.0 $\pm$ 1.7 & 51.0 $\pm$ 1.7 & 54.2 $\pm$ 1.5 \\
& \multicolumn{1}{c|}{13B} & \textit{56.3 $\pm$ 4.5} & 52.4 $\pm$ 2.4 & 50.3 $\pm$ 0.6 & 57.0 $\pm$ 2.1 & 51.8 $\pm$ 1.0 & 51.6 $\pm$ 1.3 & 50.7 $\pm$ 2.1 & 51.4 $\pm$ 2.7 \\
& \multicolumn{1}{c|}{32B} & \textit{82.5 $\pm$ 12.5} & \textbf{76.6 $\pm$ 21.4} & \textbf{64.9 $\pm$ 11.1} & 55.5 $\pm$ 5.1 & \textbf{64.8 $\pm$ 9.4} & \textbf{60.4 $\pm$ 5.8} & \textbf{64.1 $\pm$ 9.2} & \textbf{64.2 $\pm$ 5.3} \\
\midrule
\multicolumn{2}{c}{\texttt{Chance performance}} & 50.0& 50.0& 50.0& 50.0& 50.0&50.0 & 50.0 & 50.0 \\
        \bottomrule
        \end{tabular}
    }
    \caption{Accuracy on each OOD dataset for models trained on MNLI with 64 examples. Measurements are taken using the checkpoint with the highest in-domain performance. $^{\ddagger}$ in-domain dataset.}
    \label{tab:acc_nli_64shot_mnli}
    \vspace{-5pt}
\end{table}

\begin{table}[H]
    \centering
    \resizebox{\textwidth}{!}{
        \begin{tabular}{lccccccccc}
        \toprule
        && \multicolumn{8}{c}{\textbf{\colouredsnli}} \\
        \cmidrule(lr){3-10}
        \textbf{Model} & \textbf{Size} &
        \textbf{\colouredsnli}$^{\ddagger}$ & \textbf{\colouredmnli} & \textbf{\colouredwnli}& \textbf{\colouredscitail} & \textbf{\colouredrte} & \textbf{\colouredhans} & \textbf{\colouredanli} & \textbf{\colouredpaws} \\
\midrule
\multirow{3}{*}{\texttt{OPT}}
& \multicolumn{1}{c|}{2.7b} & \textit{87.5 $\pm$ 6.2}& 71.8 $\pm$ 5.8 & 51.7 $\pm$ 0.5 & 69.9 $\pm$ 4.5 & 59.0 $\pm$ 5.7 & 52.5 $\pm$ 1.0 & 50.4 $\pm$ 1.7 & 51.5 $\pm$ 4.2 \\
& \multicolumn{1}{c|}{6.7b} & \textit{88.3 $\pm$ 4.7}& 72.7 $\pm$ 9.0 & 52.7 $\pm$ 1.9 & 62.1 $\pm$ 16.5 & 61.4 $\pm$ 2.8 & 54.3 $\pm$ 2.8 & 51.3 $\pm$ 2.4 & 49.4 $\pm$ 2.9 \\
& \multicolumn{1}{c|}{13b} & \textit{93.5 $\pm$ 0.9}& \textbf{80.8 $\pm$ 4.7} & 50.6 $\pm$ 1.0 & 72.4 $\pm$ 5.1 & 66.1 $\pm$ 0.3 & 54.4 $\pm$ 3.9 & 49.9 $\pm$ 0.9 & 52.1 $\pm$ 5.1 \\
& \multicolumn{1}{c|}{30b} & \textit{\textbf{94.5 $\pm$ 1.3}} & 78.8 $\pm$ 4.2 & 54.1 $\pm$ 1.7 & \textbf{76.3 $\pm$ 1.8} & \textbf{67.2 $\pm$ 6.4} & \textbf{64.7 $\pm$ 4.0} & 51.6 $\pm$ 2.2 & 53.0 $\pm$ 4.9 \\
\midrule
\multirow{3}{*}{\texttt{OLMo2}}
& \multicolumn{1}{c|}{7B} & \textit{70.3 $\pm$ 12.5}& 56.3 $\pm$ 5.1 & 52.8 $\pm$ 0.9 & 52.3 $\pm$ 5.6 & 53.5 $\pm$ 1.6 & 52.4 $\pm$ 2.1 & 51.8 $\pm$ 0.7 & 56.7 $\pm$ 2.9 \\
& \multicolumn{1}{c|}{13B} & \textit{59.7 $\pm$ 5.2}& 54.5 $\pm$ 5.0 & 52.8 $\pm$ 1.0 & 54.7 $\pm$ 4.1 & 52.2 $\pm$ 0.3 & 53.6 $\pm$ 1.0 & 50.9 $\pm$ 0.4 & 52.4 $\pm$ 1.7 \\
& \multicolumn{1}{c|}{32B} & \textit{92.7 $\pm$ 4.0}& 72.1 $\pm$ 8.7 & \textbf{61.8 $\pm$ 7.2} & 61.0 $\pm$ 4.4 & 61.1 $\pm$ 3.6 & 60.7 $\pm$ 1.7 & \textbf{57.7 $\pm$ 4.4} & \textbf{61.7 $\pm$ 3.6} \\
\midrule
\multicolumn{2}{c}{\texttt{Chance performance}} & 50.0& 50.0& 50.0& 50.0& 50.0&50.0 & 50.0 & 50.0 \\
        \bottomrule
        \end{tabular}
    }
    \caption{Accuracy on each OOD dataset for models trained on SNLI with 64 examples. Measurements are taken using the checkpoint with the highest in-domain performance. $^{\ddagger}$ in-domain dataset.}
    \label{tab:acc_nli_64shot_snli}
    \vspace{-5pt}
\end{table}

\begin{table}[H]
    \centering
    \resizebox{\textwidth}{!}{
        \begin{tabular}{lccccccccc}
        \toprule
        && \multicolumn{8}{c}{\textbf{\colouredmnli}} \\
        \cmidrule(lr){3-10}
        \textbf{Model} & \textbf{Size} & \textbf{\colouredmnli}$^{\ddagger}$ & \textbf{\colouredsnli} & \textbf{\colouredwnli}& \textbf{\colouredscitail} & \textbf{\colouredrte} & \textbf{\colouredhans} & \textbf{\colouredanli} & \textbf{\colouredpaws} \\
\midrule
\multirow{3}{*}{\texttt{OPT}}
& \multicolumn{1}{c|}{2.7b} & \textit{58.8 $\pm$ 4.6} & 52.8 $\pm$ 4.4 & 51.4 $\pm$ 0.6 & 59.1 $\pm$ 4.2 & 52.0 $\pm$ 2.8 & 51.4 $\pm$ 0.3 & 50.3 $\pm$ 2.3 & 49.7 $\pm$ 4.3 \\
& \multicolumn{1}{c|}{6.7b} & \textit{65.1 $\pm$ 6.9} & 58.0 $\pm$ 9.3 & 50.4 $\pm$ 1.4 & 59.3 $\pm$ 1.1 & 52.7 $\pm$ 3.4 & 52.1 $\pm$ 1.2 & 51.0 $\pm$ 2.0 & 51.4 $\pm$ 5.1 \\
& \multicolumn{1}{c|}{13b} & \textit{68.1 $\pm$ 10.0} & 59.6 $\pm$ 15.8 & 49.9 $\pm$ 0.3 & \textbf{64.1 $\pm$ 6.4} & 55.4 $\pm$ 8.0 & 53.4 $\pm$ 0.9 & 50.0 $\pm$ 1.9 & 52.1 $\pm$ 6.0 \\
& \multicolumn{1}{c|}{30b} & \textit{68.2 $\pm$ 6.3} & \textbf{60.5 $\pm$ 15.3} & 51.2 $\pm$ 0.5 & 57.5 $\pm$ 4.2 & 54.5 $\pm$ 4.6 & 52.3 $\pm$ 3.8 & 52.4 $\pm$ 3.2 & 52.9 $\pm$ 4.2 \\
\midrule
\multirow{3}{*}{\texttt{OLMo2}}
& \multicolumn{1}{c|}{7B} & \textit{57.9 $\pm$ 4.7} & 50.8 $\pm$ 1.6 & 50.7 $\pm$ 0.6 & 52.6 $\pm$ 3.9 & 51.0 $\pm$ 0.8 & 51.2 $\pm$ 0.6 & 50.9 $\pm$ 0.8 & 53.3 $\pm$ 0.4 \\
& \multicolumn{1}{c|}{13B} & \textit{54.0 $\pm$ 5.5} & 50.3 $\pm$ 0.5 & 51.4 $\pm$ 0.4 & 56.1 $\pm$ 0.7 & 52.0 $\pm$ 0.9 & 52.2 $\pm$ 0.5 & 48.6 $\pm$ 1.4 & 50.5 $\pm$ 1.7 \\
& \multicolumn{1}{c|}{32B} & \textit{\textbf{70.6 $\pm$ 7.5}} & 60.3 $\pm$ 8.4 & \textbf{55.6 $\pm$ 4.2} & 56.0 $\pm$ 1.9 & \textbf{58.4 $\pm$ 6.0} & \textbf{55.0 $\pm$ 3.1} & \textbf{57.5 $\pm$ 5.1} & \textbf{58.8 $\pm$ 3.8} \\
\midrule
\multicolumn{2}{c}{\texttt{Chance performance}} & 50.0& 50.0& 50.0& 50.0& 50.0&50.0 & 50.0 & 50.0 \\
        \bottomrule
        \end{tabular}
    }
    \caption{Accuracy on each OOD dataset for models trained on MNLI with 32 examples. Measurements are taken using the checkpoint with the highest in-domain performance. $^{\ddagger}$ in-domain dataset.}
    \label{tab:acc_nli_32shot_mnli}
    \vspace{-5pt}
\end{table}

\begin{table}[H]
    \centering
    \resizebox{\textwidth}{!}{
        \begin{tabular}{lccccccccc}
        \toprule
        && \multicolumn{8}{c}{\textbf{\colouredsnli}} \\
        \cmidrule(lr){3-10}
        \textbf{Model} & \textbf{Size} &
        \textbf{\colouredsnli}$^{\ddagger}$ & \textbf{\colouredmnli} & \textbf{\colouredwnli}& \textbf{\colouredscitail} & \textbf{\colouredrte} & \textbf{\colouredhans} & \textbf{\colouredanli} & \textbf{\colouredpaws} \\
\midrule
\multirow{3}{*}{\texttt{OPT}}
& \multicolumn{1}{c|}{2.7b} & \textit{65.2 $\pm$ 5.1}& 58.3 $\pm$ 5.9 & 51.2 $\pm$ 0.3 & 61.2 $\pm$ 6.3 & 52.1 $\pm$ 2.0 & 52.0 $\pm$ 1.6 & 51.3 $\pm$ 1.3 & 51.9 $\pm$ 3.2 \\
& \multicolumn{1}{c|}{6.7b} & \textit{69.7 $\pm$ 3.8}& 59.3 $\pm$ 6.3 & 51.4 $\pm$ 0.9 & 64.4 $\pm$ 6.7 & 54.0 $\pm$ 2.7 & 53.3 $\pm$ 1.4 & 50.3 $\pm$ 0.7 & 51.5 $\pm$ 4.3 \\
& \multicolumn{1}{c|}{13b} & \textit{\textbf{82.9 $\pm$ 9.3}} & \textbf{71.9 $\pm$ 2.7} & 51.6 $\pm$ 0.6 & \textbf{66.4 $\pm$ 1.7} & \textbf{62.7 $\pm$ 2.8} & \textbf{57.1 $\pm$ 4.0} & 50.2 $\pm$ 1.6 & 53.1 $\pm$ 3.2 \\
& \multicolumn{1}{c|}{30b} & \textit{75.8 $\pm$ 8.2}& 62.5 $\pm$ 8.5 & 51.4 $\pm$ 1.2 & 60.9 $\pm$ 10.9 & 54.5 $\pm$ 7.5 & 53.4 $\pm$ 5.5 & 50.8 $\pm$ 1.3 & 50.9 $\pm$ 4.9 \\
\midrule
\multirow{3}{*}{\texttt{OLMo2}}
& \multicolumn{1}{c|}{7B} & \textit{56.7 $\pm$ 3.4}& 53.6 $\pm$ 0.2 & 51.0 $\pm$ 0.6 & 49.3 $\pm$ 5.7 & 52.1 $\pm$ 0.3 & 49.9 $\pm$ 0.6 & 51.0 $\pm$ 0.9 & 53.1 $\pm$ 1.5 \\
& \multicolumn{1}{c|}{13B} & \textit{52.6 $\pm$ 1.4}& 52.7 $\pm$ 5.5 & 51.3 $\pm$ 0.3 & 56.8 $\pm$ 1.1 & 51.3 $\pm$ 0.4 & 52.5 $\pm$ 1.2 & 50.6 $\pm$ 0.4 & 52.3 $\pm$ 0.4 \\
& \multicolumn{1}{c|}{32B} & \textit{67.7 $\pm$ 8.8}& 59.0 $\pm$ 5.5 & \textbf{54.2 $\pm$ 3.8} & 61.5 $\pm$ 1.8 & 53.9 $\pm$ 1.5 & 52.6 $\pm$ 2.7 & \textbf{52.9 $\pm$ 1.2} & \textbf{57.0 $\pm$ 1.7} \\
\midrule
\multicolumn{2}{c}{\texttt{Chance performance}} & 50.0& 50.0& 50.0& 50.0& 50.0&50.0 & 50.0 & 50.0 \\
        \bottomrule
        \end{tabular}
    }
    \caption{Accuracy on each OOD dataset for models trained on SNLI with 32 examples. Measurements are taken using the checkpoint with the highest in-domain performance. $^{\ddagger}$ in-domain dataset.}
    \label{tab:acc_nli_32shot_snli}
    \vspace{-5pt}
\end{table}

\subsection{Performance across Finetuning Runs}

\begin{figure}[H]

    \centering
    \begin{subfigure}[b]{0.5\textwidth}
        \includegraphics[width=\textwidth]{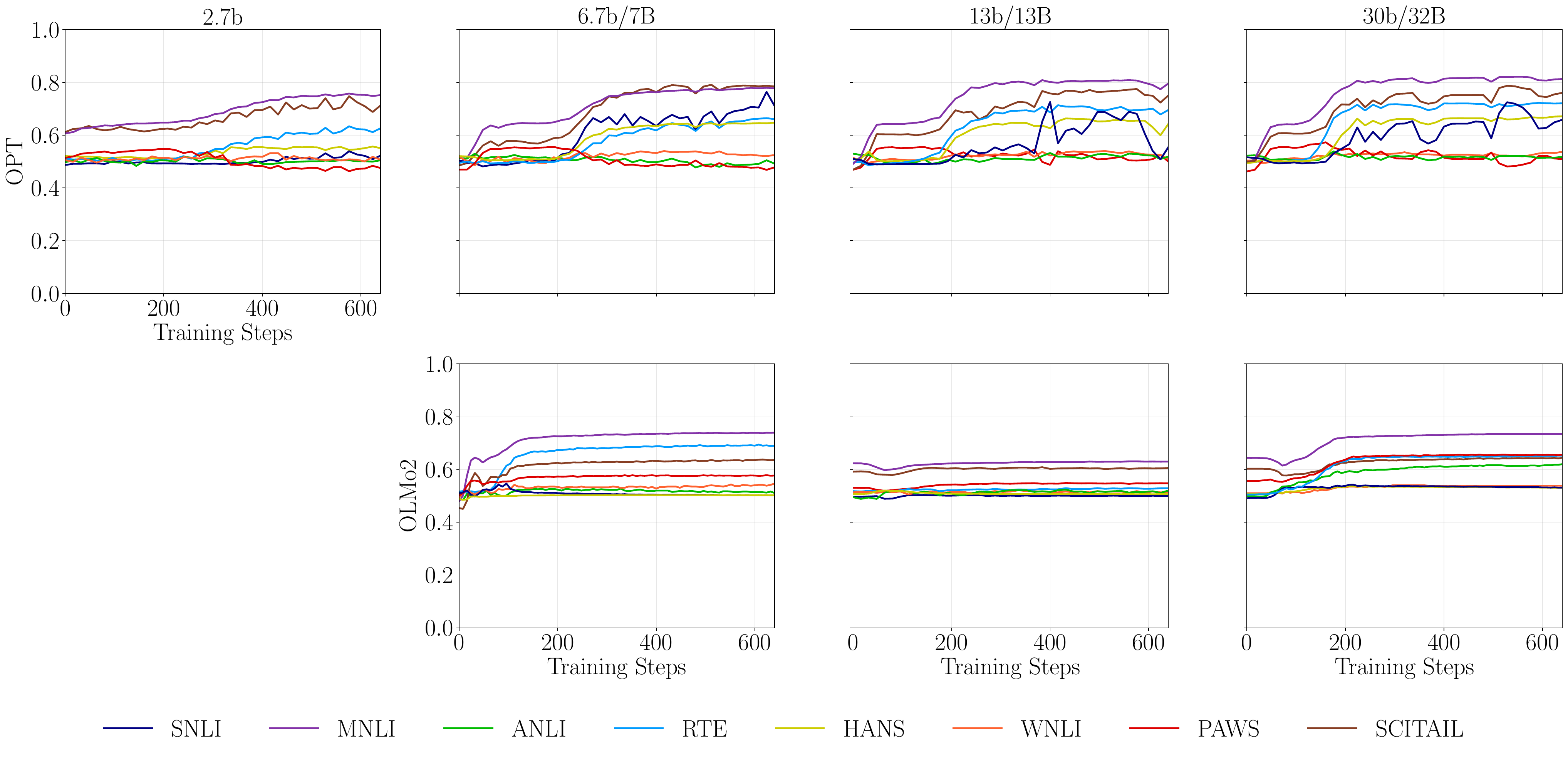}
    \caption{128-shot on MNLI}
    \end{subfigure}%
    ~
    \begin{subfigure}[b]{0.5\textwidth}
        \includegraphics[width=\textwidth]{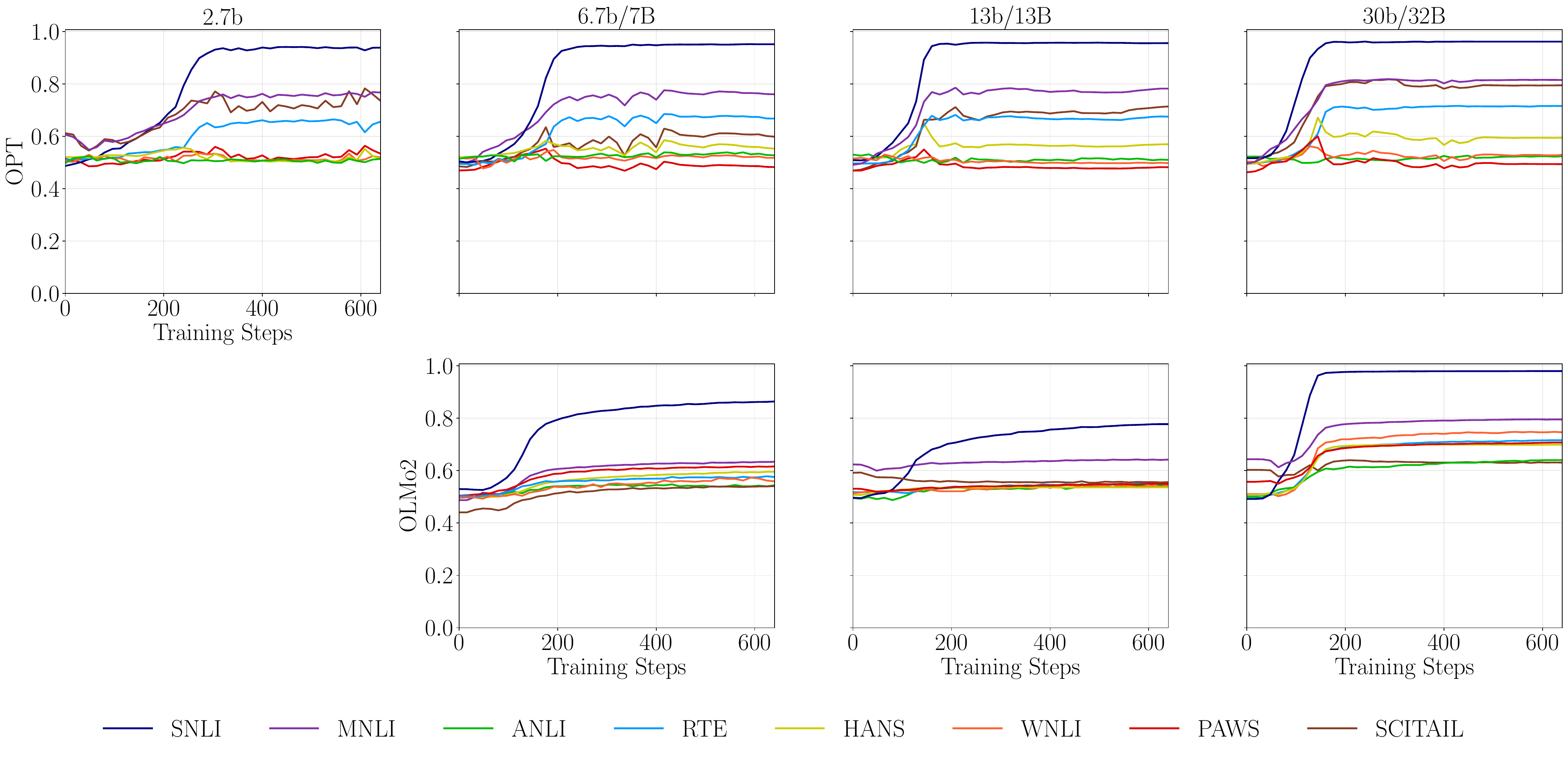}
    \caption{128-shot on SNLI}
    \end{subfigure}
    
    \begin{subfigure}[b]{0.5\textwidth}
    \includegraphics[width=\linewidth]{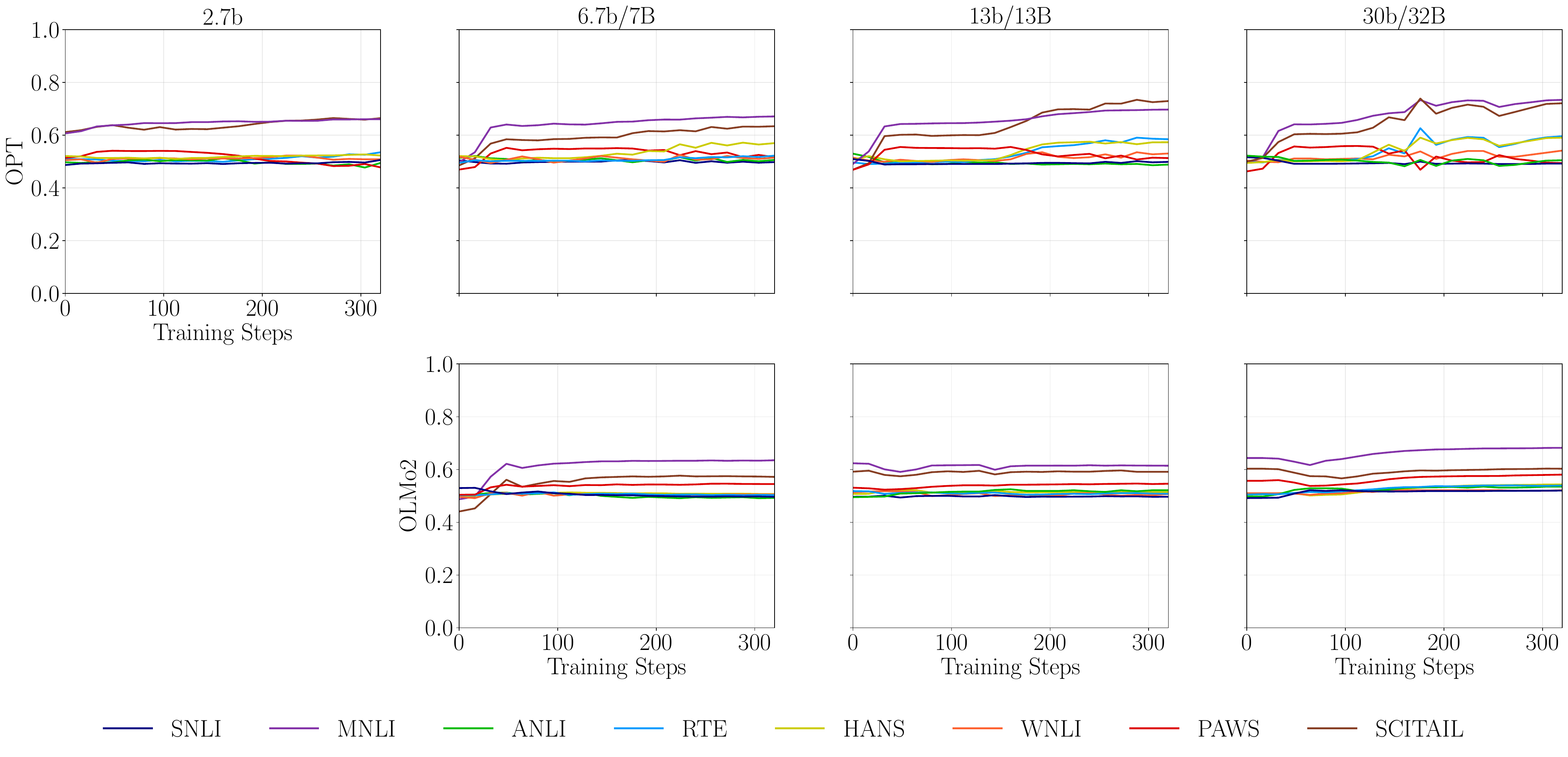}
    \caption{64-shot on MNLI}
    \end{subfigure}%
    ~
    \begin{subfigure}[b]{0.5\textwidth}
    \includegraphics[width=\linewidth]{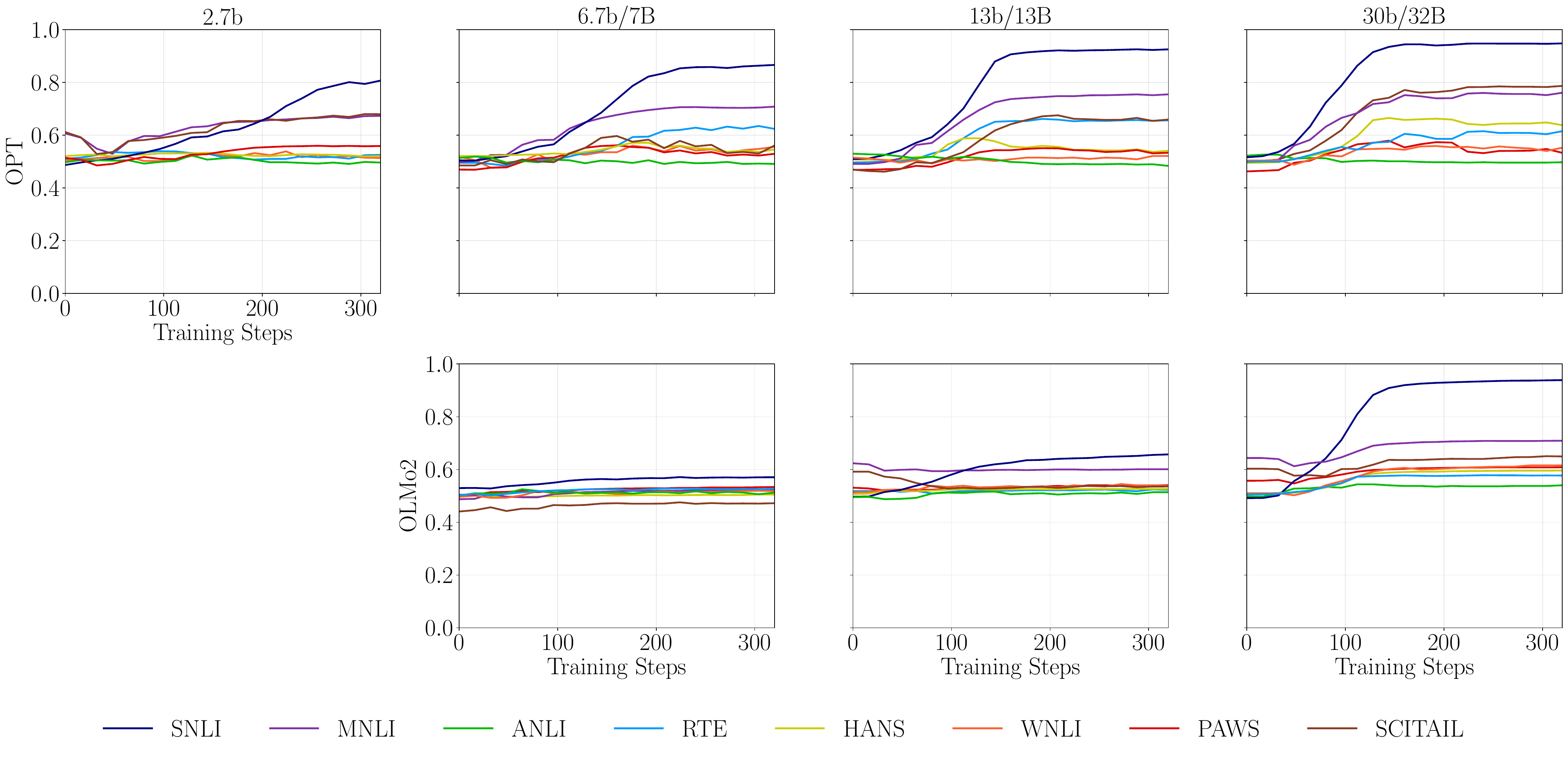}
    \caption{64-shot on SNLI}
    \end{subfigure}
    
    \begin{subfigure}[b]{0.5\textwidth}
        \includegraphics[width=\textwidth]{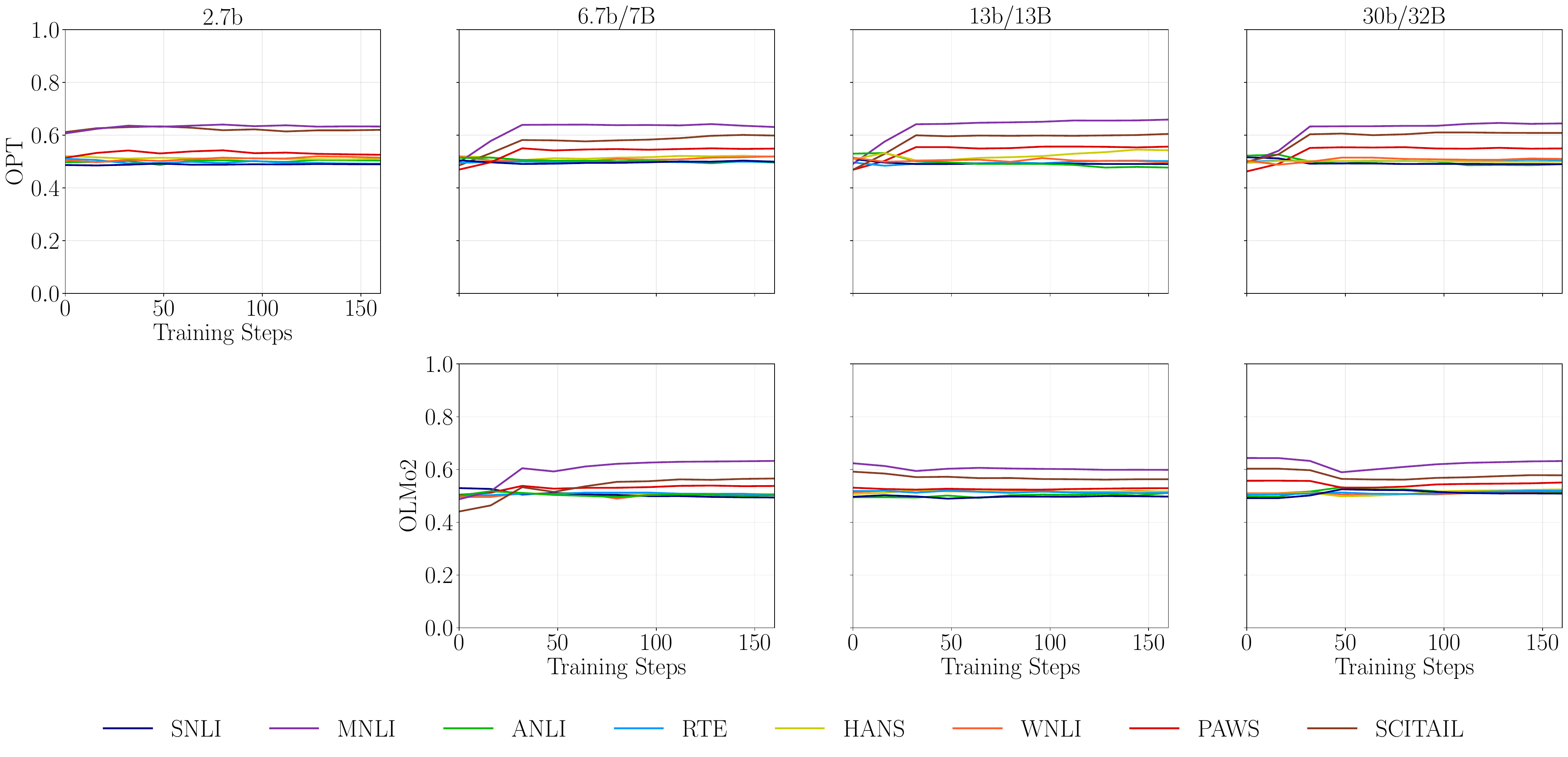}
    \caption{32-shot on MNLI}
    \end{subfigure}%
    ~
    \begin{subfigure}[b]{0.5\textwidth}
        \includegraphics[width=\textwidth]{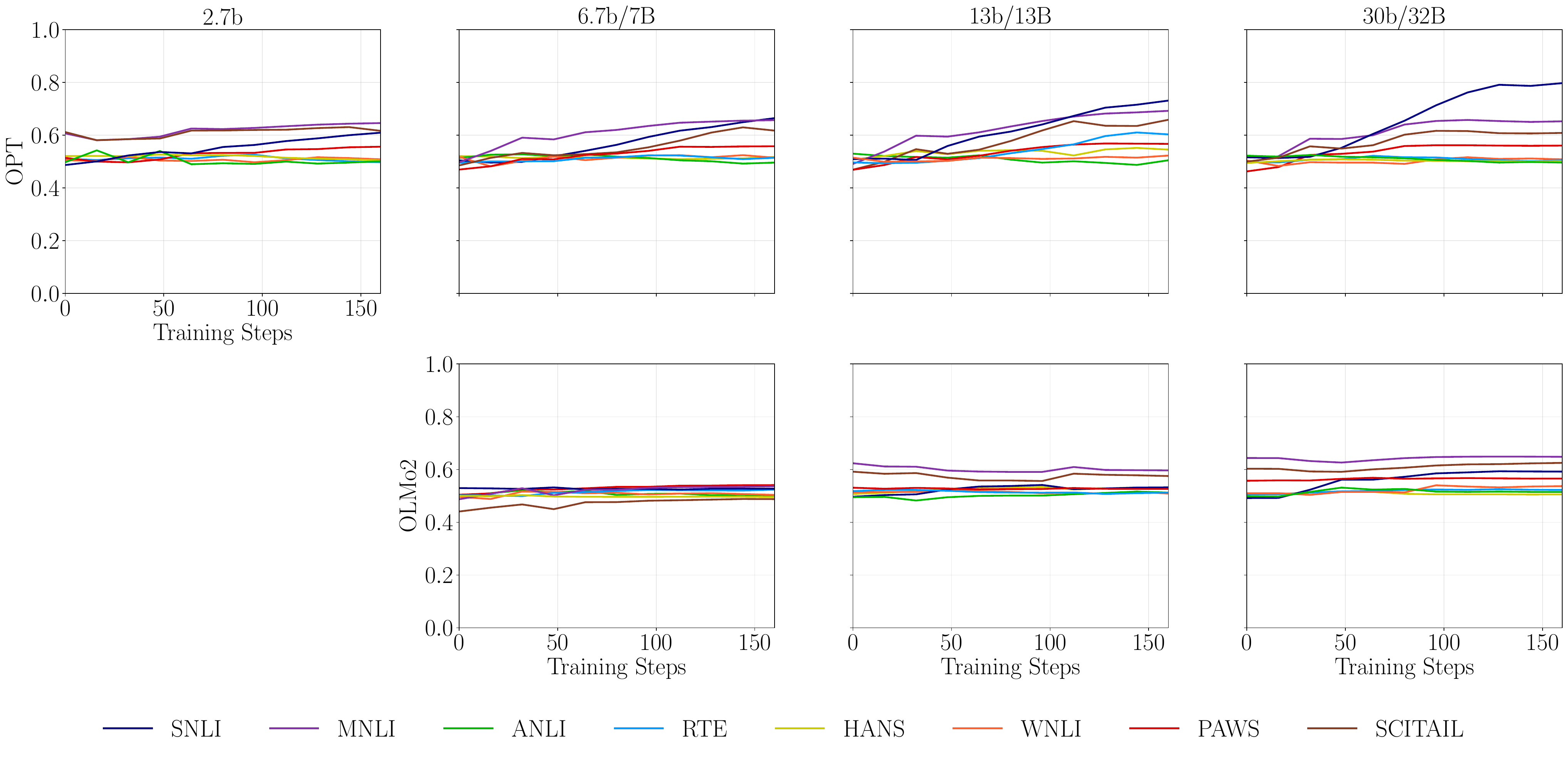}
    \caption{32-shot on SNLI}
    \end{subfigure}
    
    \caption{Few-shots results throughout a finetuning run on either MNLI or SNLI. \opt OOD performances (first rows) frequently oscillate during training; \olmo OOD performances (second rows) are relatively stable across training. Legend: \colouredmnli, \colouredsnli, \colouredwnli, \colouredrte, \colouredscitail, \colouredanli, \colouredhans and \colouredpaws \looseness=-1}
    \label{fig:odd_performance_across_train_full_128shots}
\end{figure}

\subsection{\opt's Partial OOD Correlation Graphs}

\begin{figure}[H]
  \centering
  \begin{subfigure}[b]{\textwidth}
    \centering
    \includegraphics[width=\textwidth]{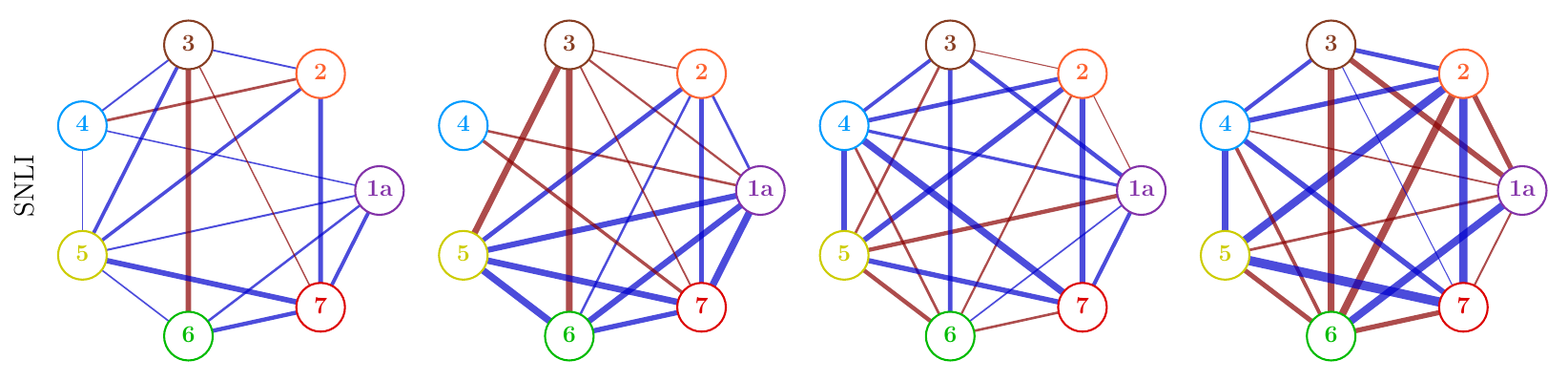}
  \end{subfigure}
  
  \begin{subfigure}[b]{\textwidth}
    \centering
    \includegraphics[width=\textwidth]{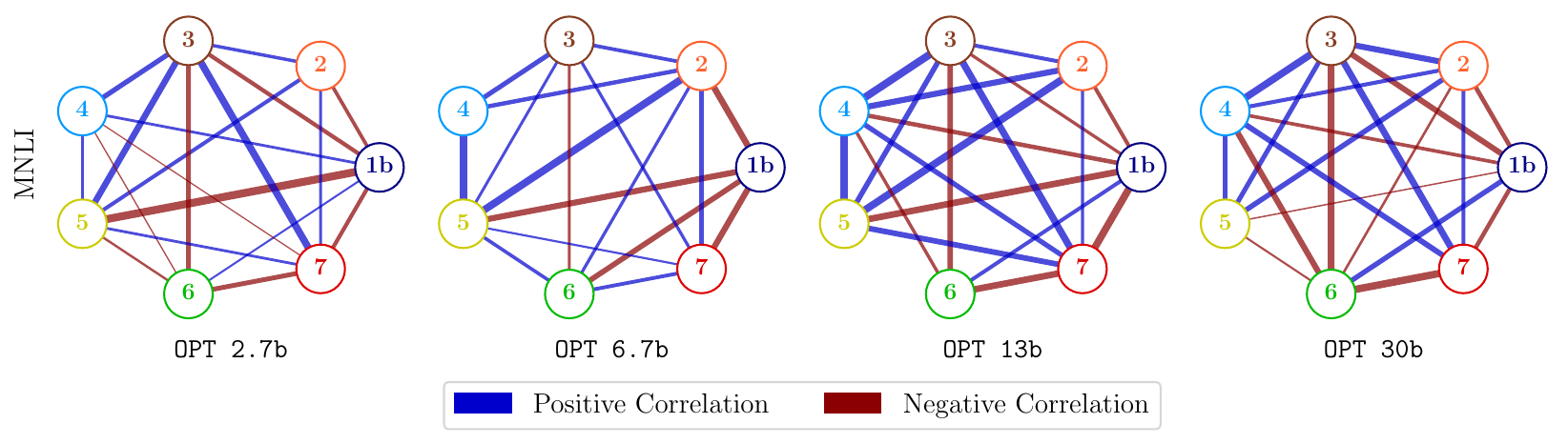}
  \end{subfigure}

  \caption{\opt~partial OOD correlation graphs on SNLI (top) and MNLI (bottom). Edge thickness increases with absolute correlation value. Legend: \colouredmnlin, \colouredsnlin, \colouredwnlin, \colouredrten, \colouredscitailn, \colouredanlin, \colouredhansn, and \colouredpawsn \looseness=-1}
  \label{fig:opt_graph}
\end{figure}

\subsection{Fit of Regressors Used when Computing Partial Correlations}

\newcommand{\colouredlinear}{\textcolor{blue}{\texttt{Linear}}\xspace}
\newcommand{\colouredridge}{\textcolor{orange}{\texttt{Ridge}}\xspace}
\newcommand{\colouredgam}{\textcolor{Green}{\texttt{GAM}}\xspace}

\begin{figure}[H]
    \centering
    \begin{subfigure}[b]{0.14\textwidth}
        \centering
        \includegraphics[width=\linewidth]{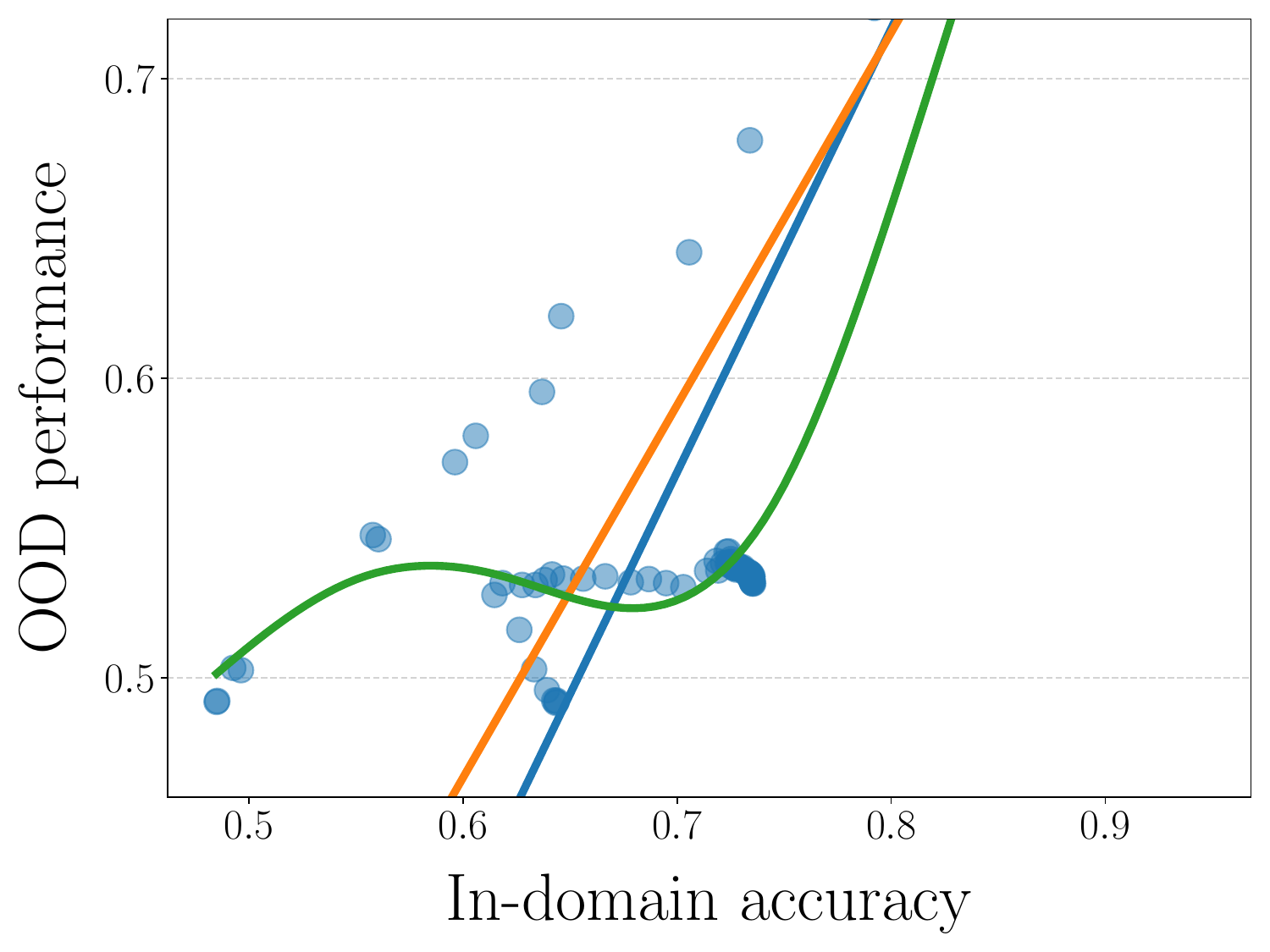}
    \end{subfigure}%
    \begin{subfigure}[b]{0.14\textwidth}
        \centering
        \includegraphics[width=\linewidth]{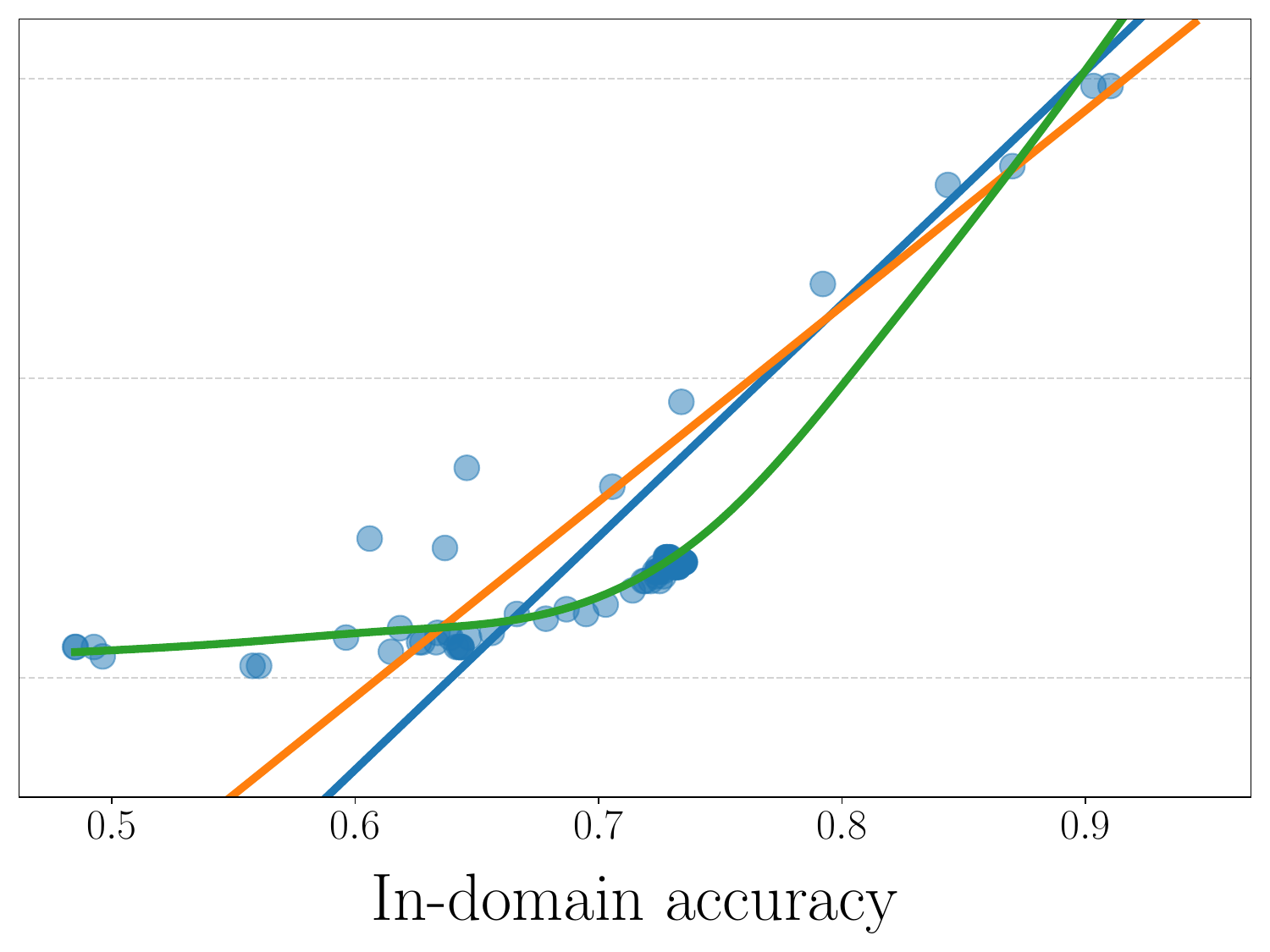}
    \end{subfigure}%
    \begin{subfigure}[b]{0.14\textwidth}
        \centering
        \includegraphics[width=\linewidth]{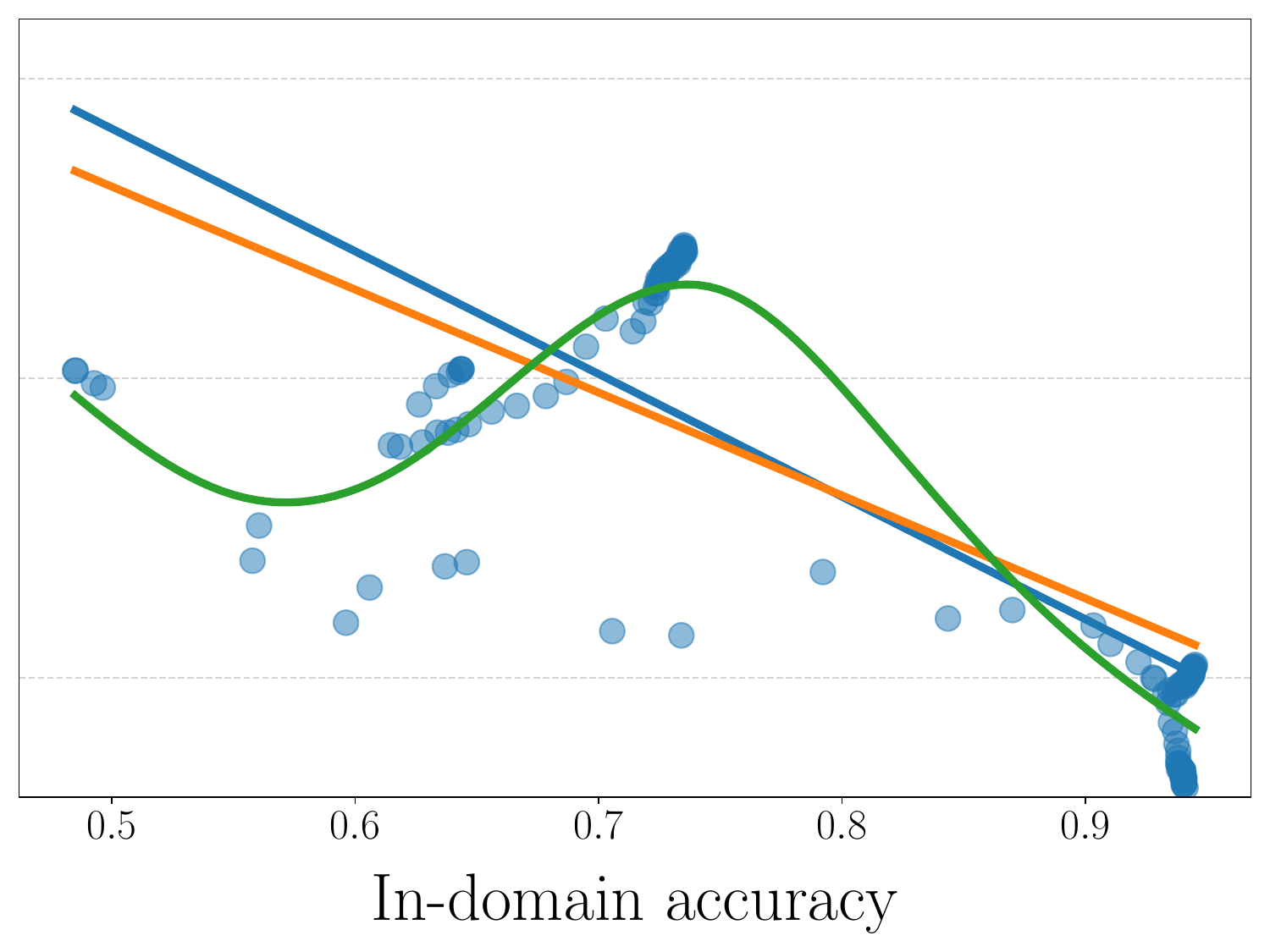}
    \end{subfigure}%
    \begin{subfigure}[b]{0.14\textwidth}
        \centering
        \includegraphics[width=\linewidth]{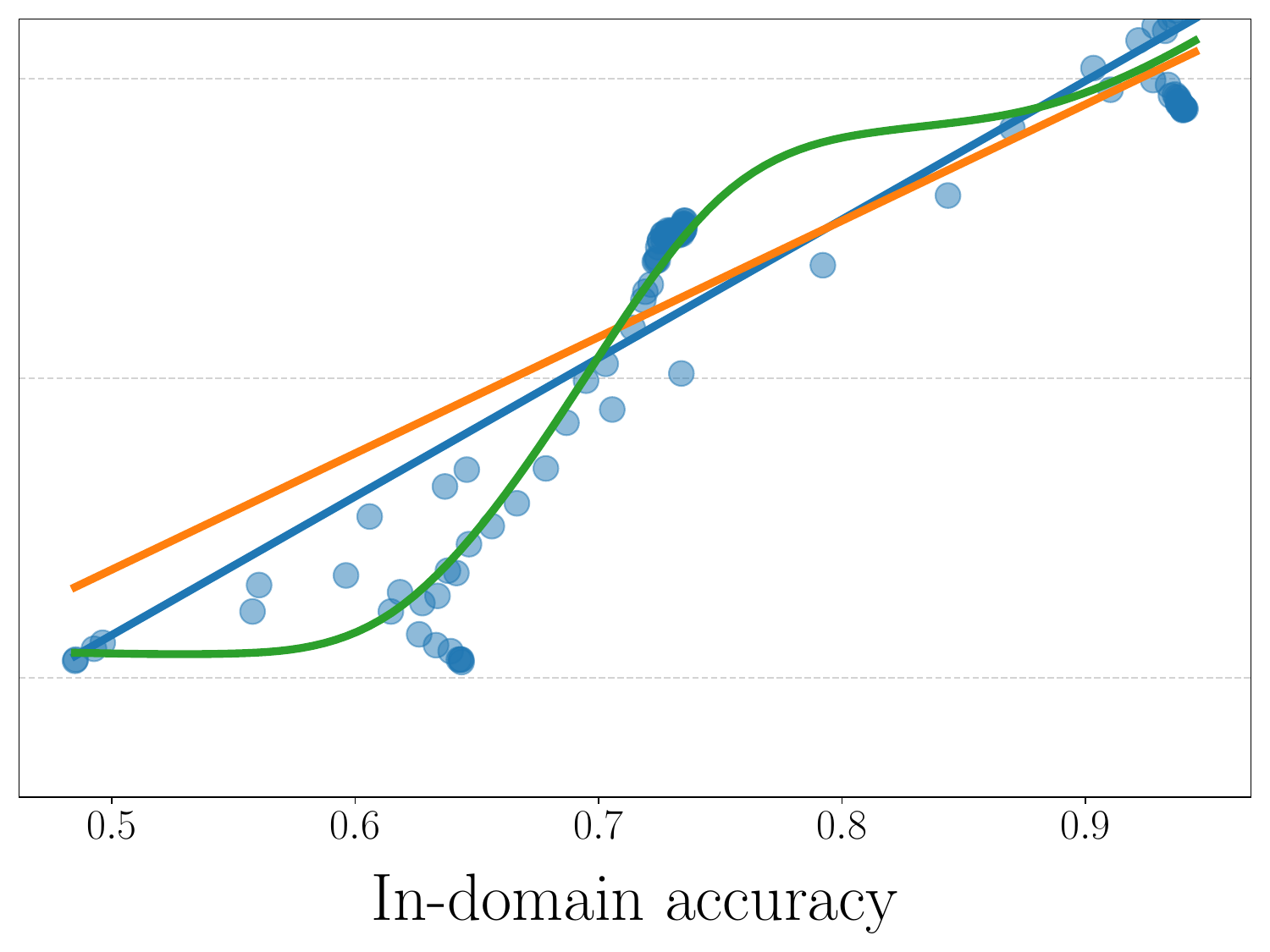}
    \end{subfigure}%
    \begin{subfigure}[b]{0.14\textwidth}
        \centering
        \includegraphics[width=\linewidth]{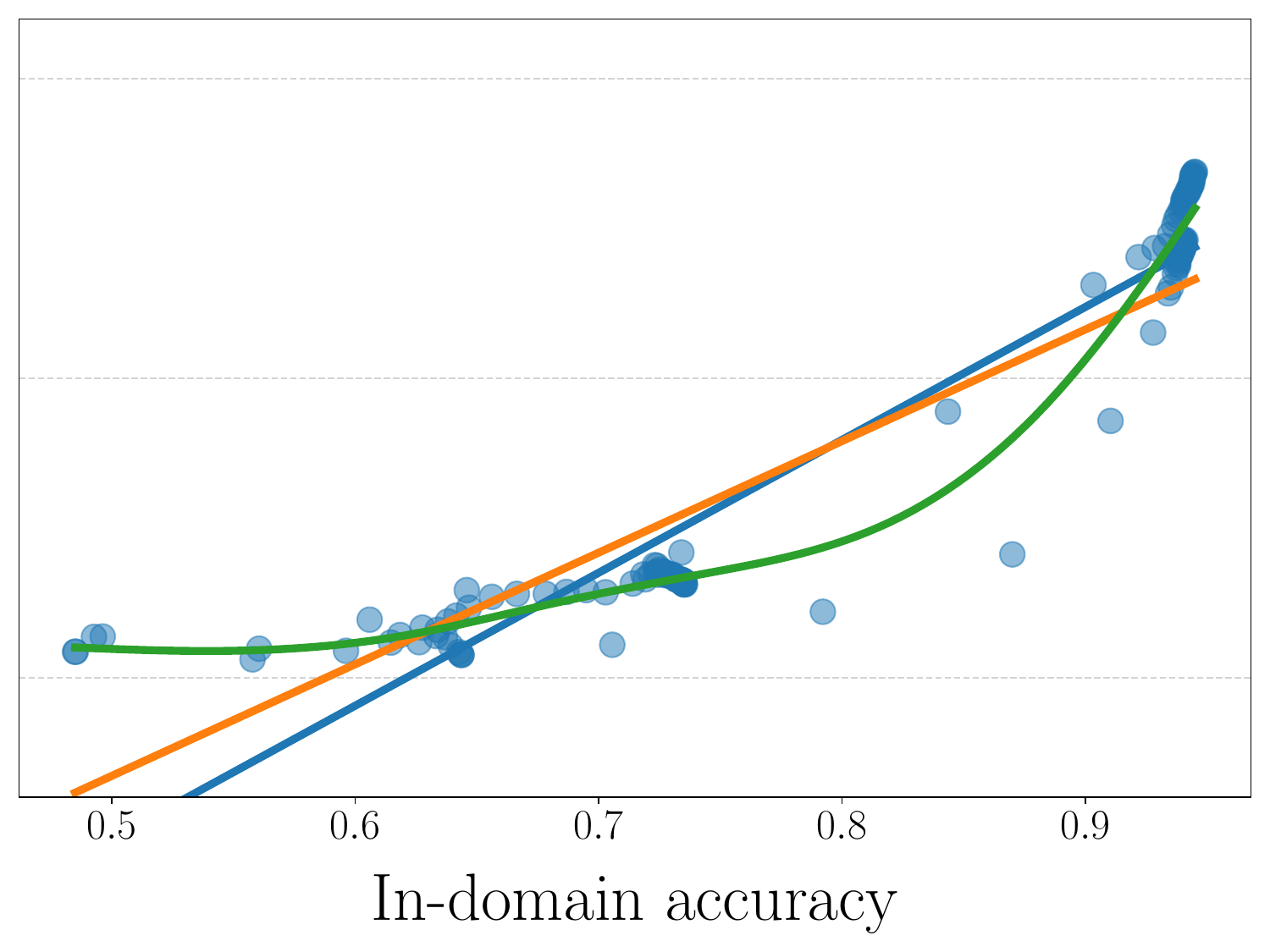}
    \end{subfigure}%
    \begin{subfigure}[b]{0.14\textwidth}
        \centering
        \includegraphics[width=\linewidth]{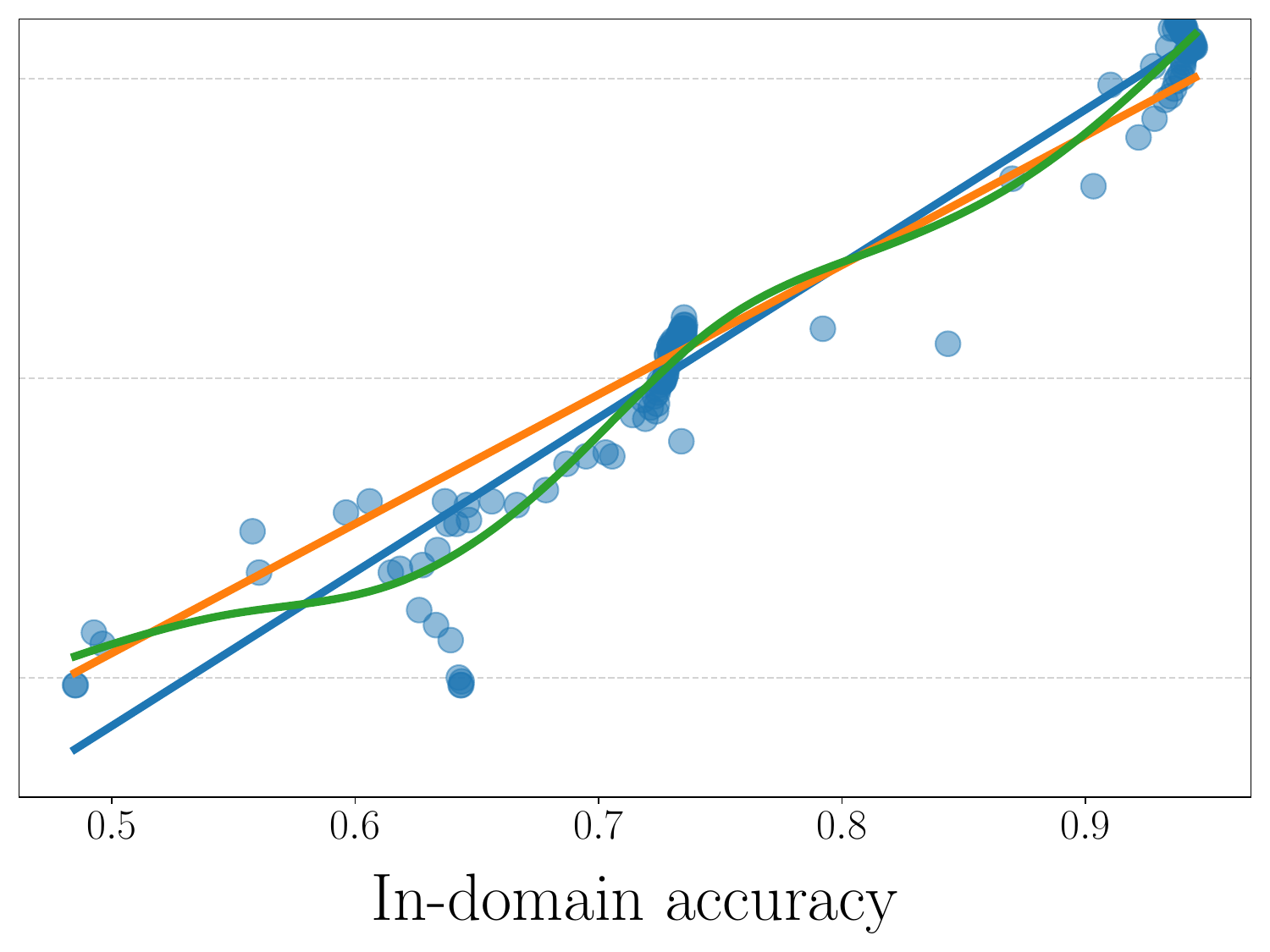}
    \end{subfigure}%
    \begin{subfigure}[b]{0.14\textwidth}
        \centering
        \includegraphics[width=\linewidth]{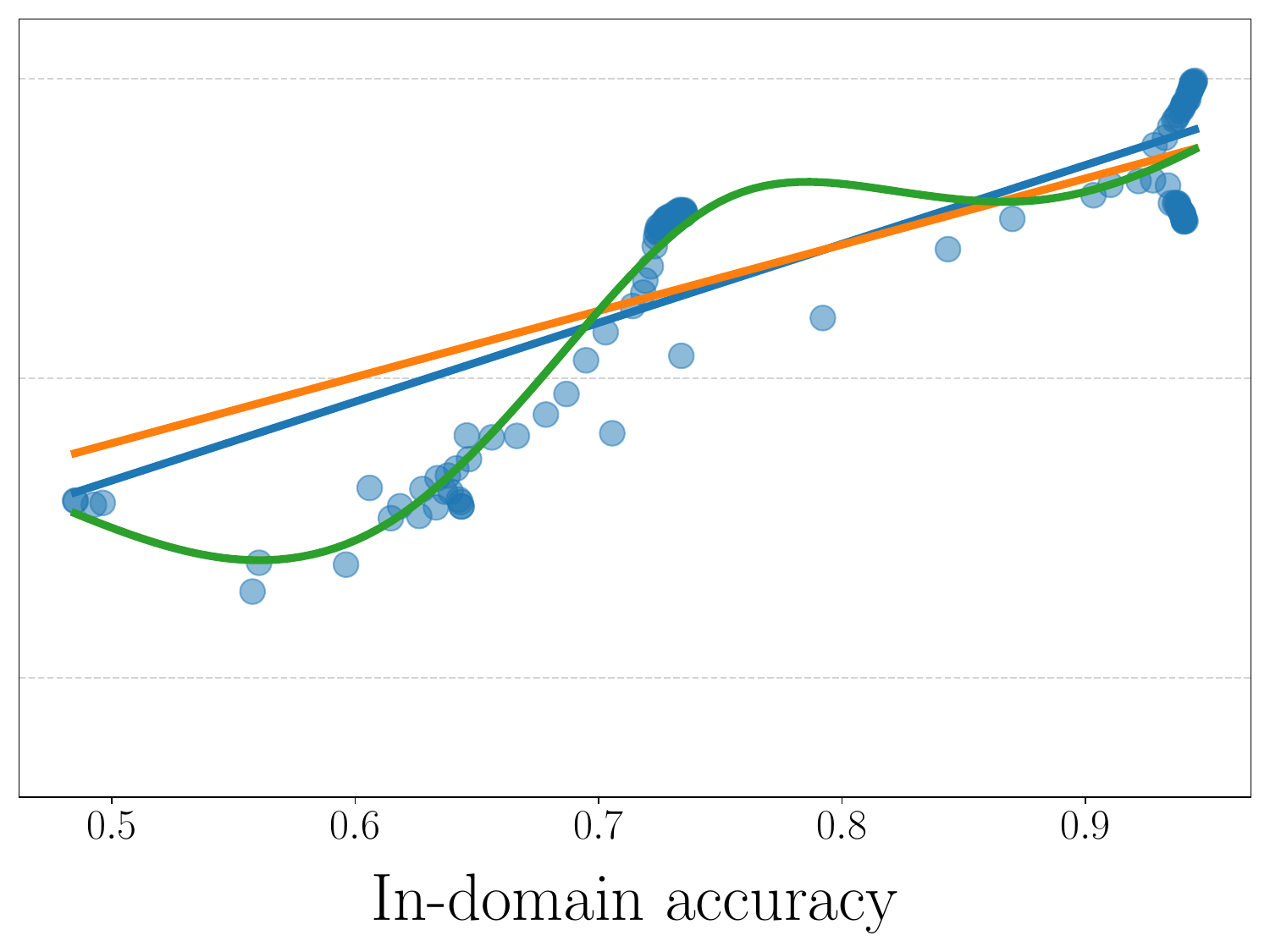}
    \end{subfigure}
    \begin{subfigure}[b]{0.14\textwidth}
        \centering
        \includegraphics[width=\linewidth]{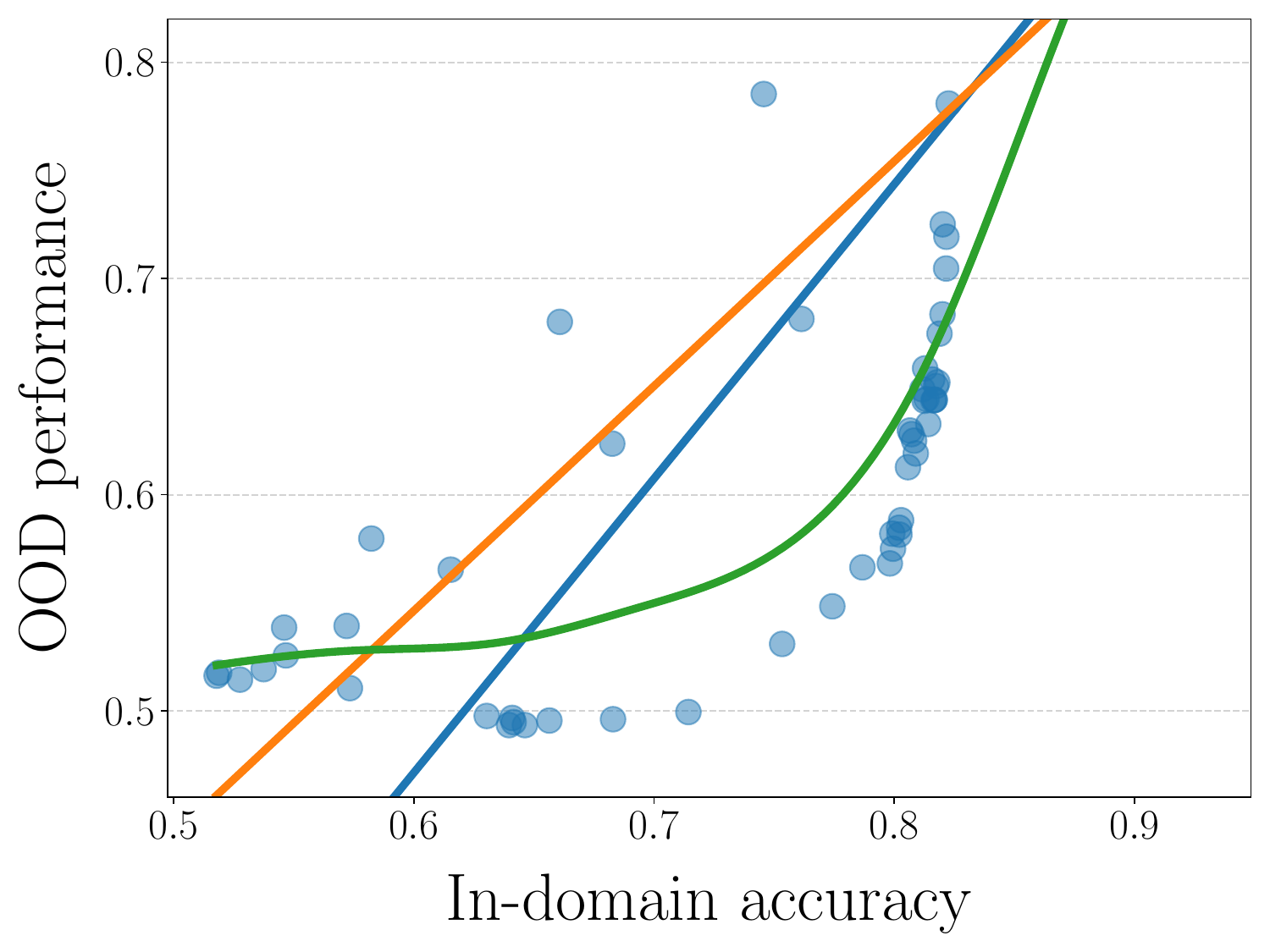}
        \scriptsize\colouredsnli
    \end{subfigure}%
    \begin{subfigure}[b]{0.14\textwidth}
        \centering
        \includegraphics[width=\linewidth]{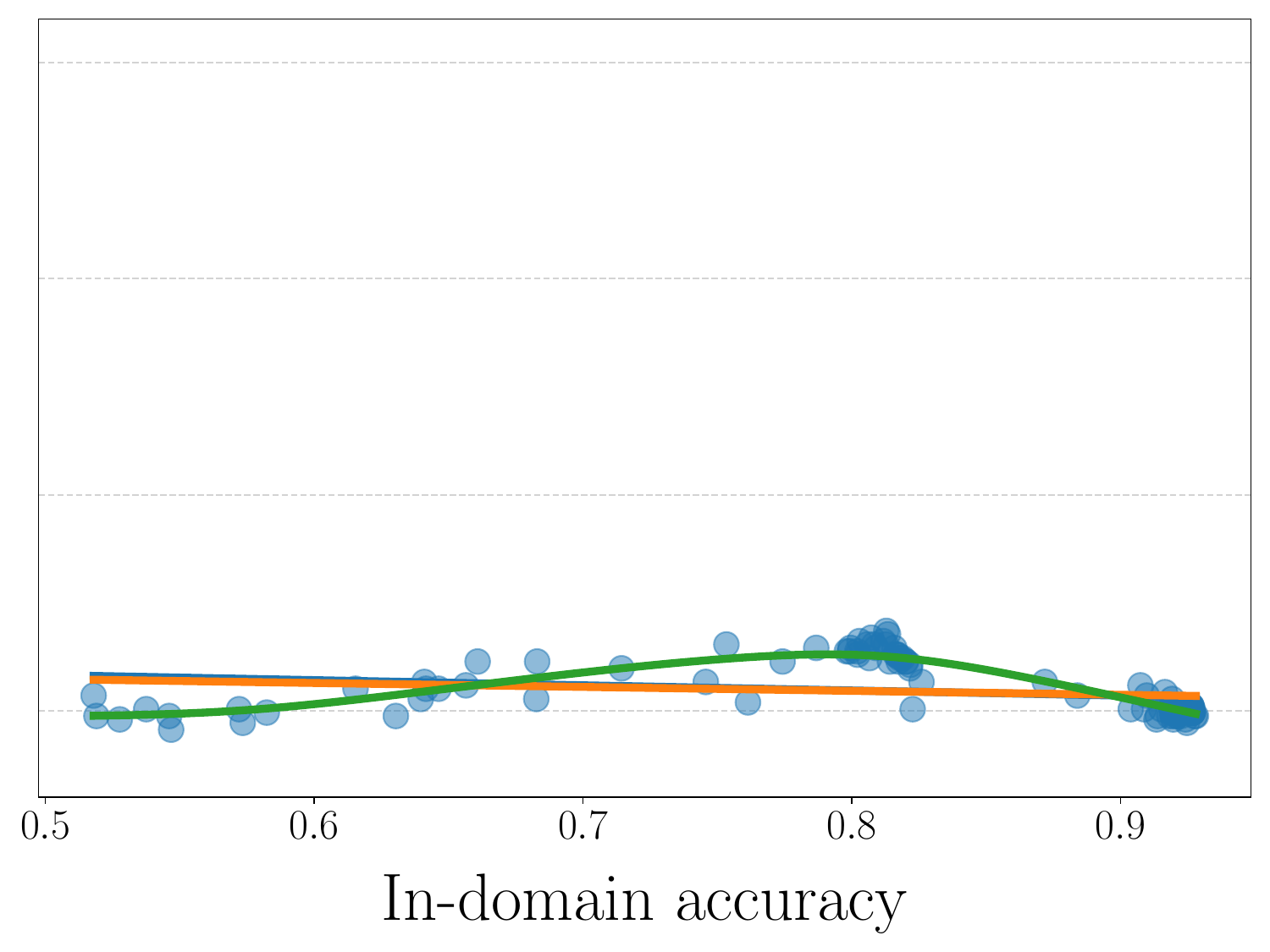}
        \scriptsize\colouredwnli
    \end{subfigure}%
    \begin{subfigure}[b]{0.14\textwidth}
        \centering
        \includegraphics[width=\linewidth]{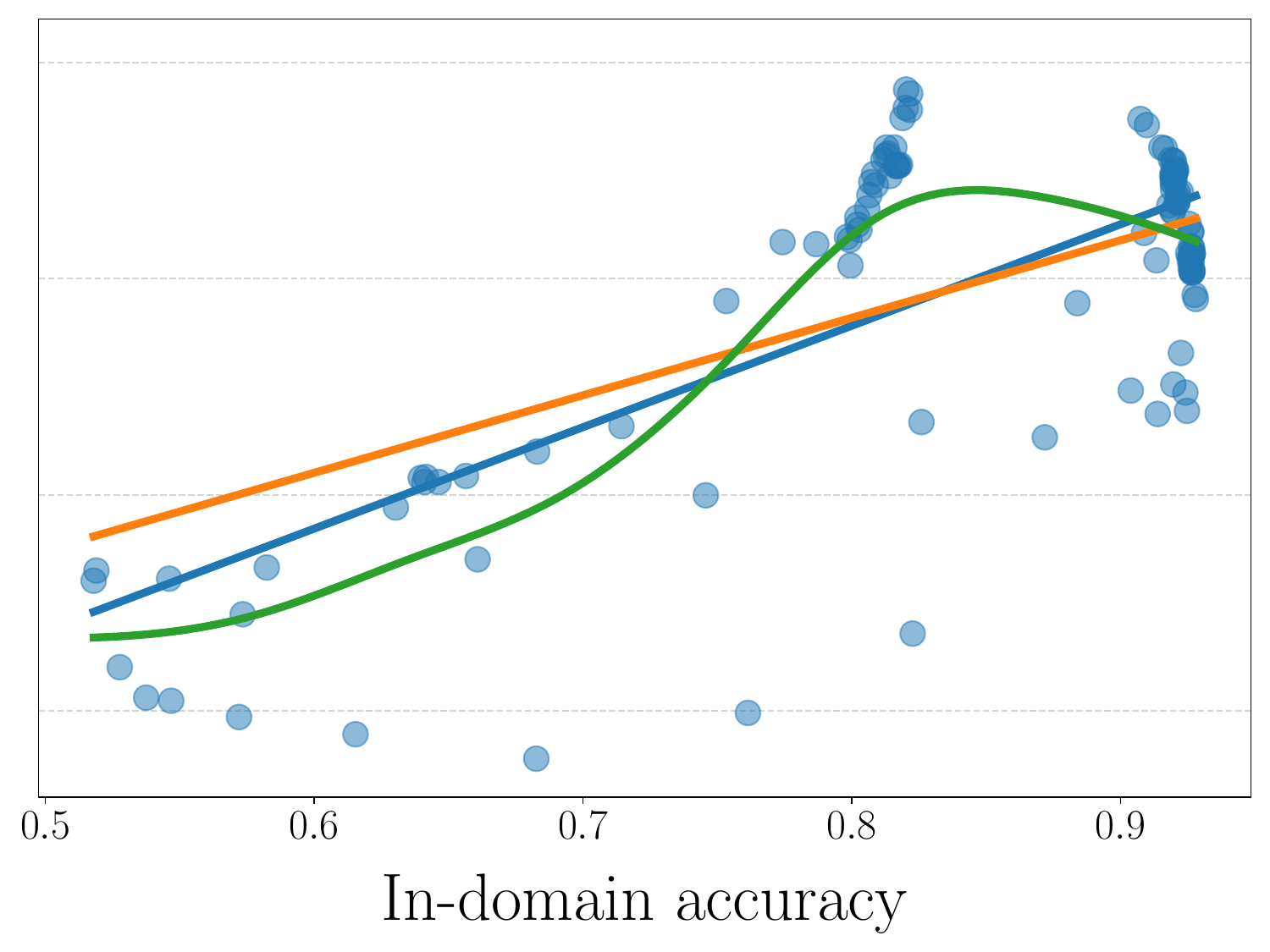}
        \scriptsize\colouredscitail
    \end{subfigure}%
    \begin{subfigure}[b]{0.14\textwidth}
        \centering
        \includegraphics[width=\linewidth]{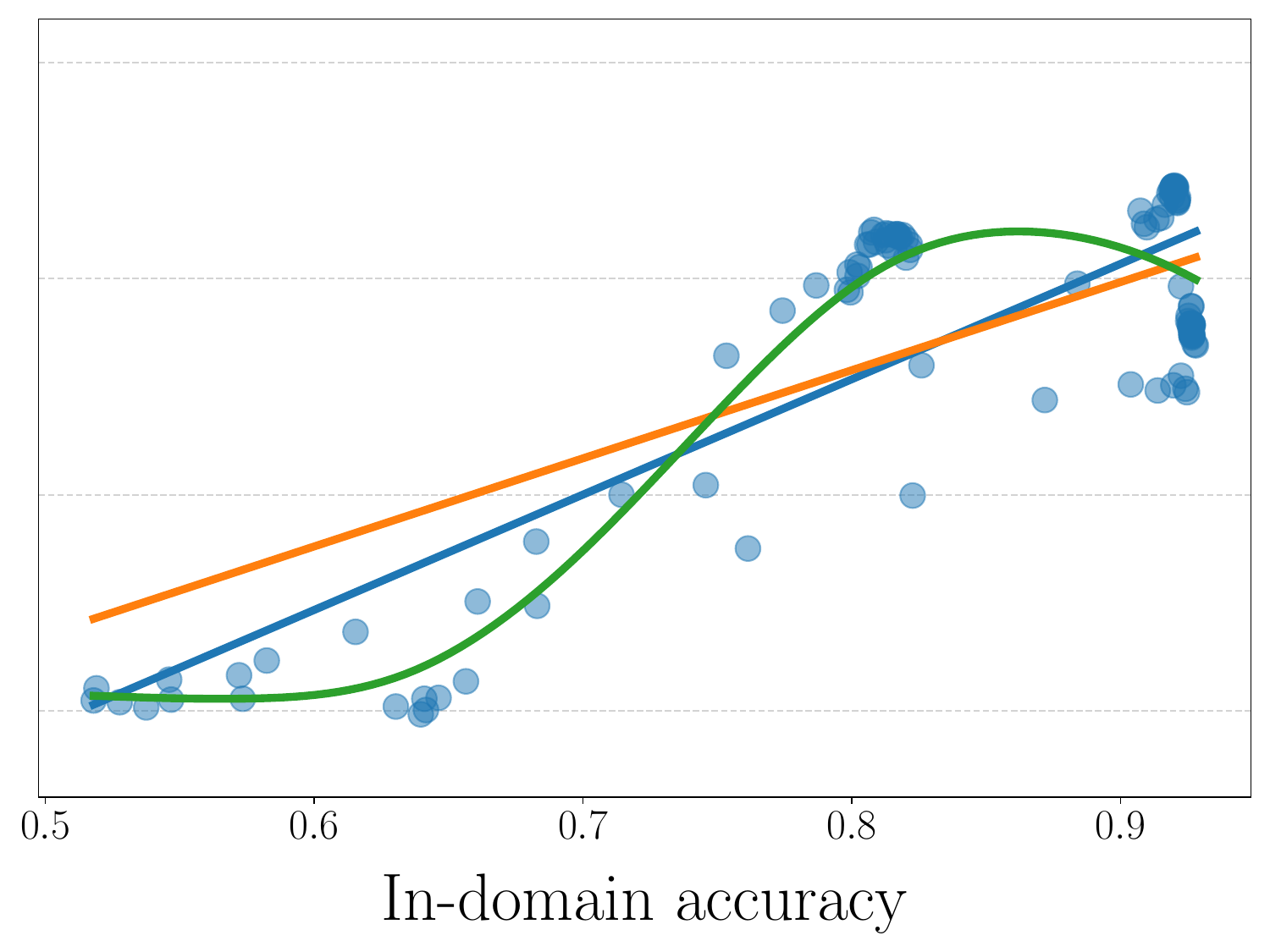}
        \scriptsize\colouredrte
    \end{subfigure}%
    \begin{subfigure}[b]{0.14\textwidth}
        \centering
        \includegraphics[width=\linewidth]{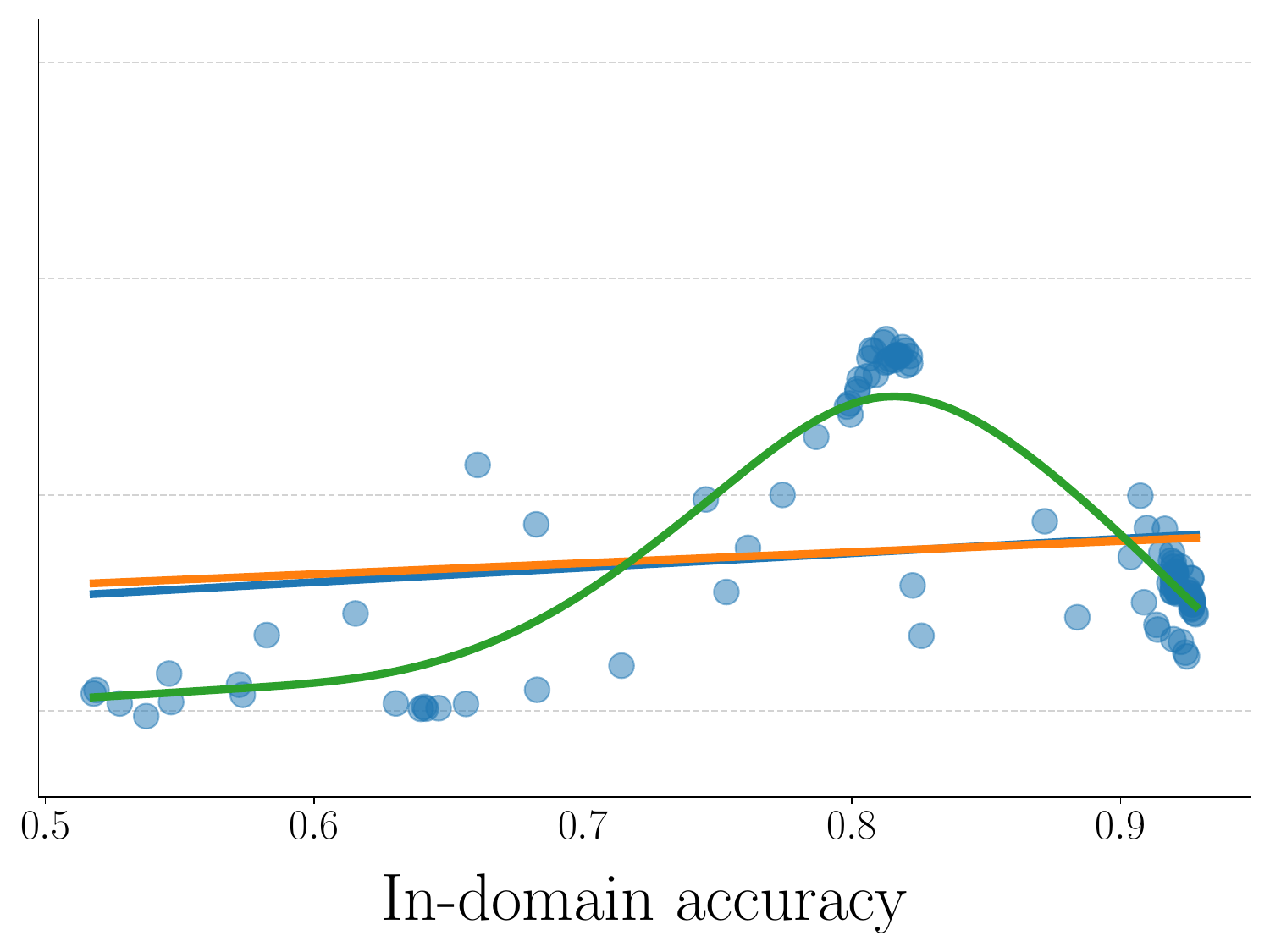}
        \scriptsize\colouredhans
    \end{subfigure}%
    \begin{subfigure}[b]{0.14\textwidth}
        \centering
        \includegraphics[width=\linewidth]{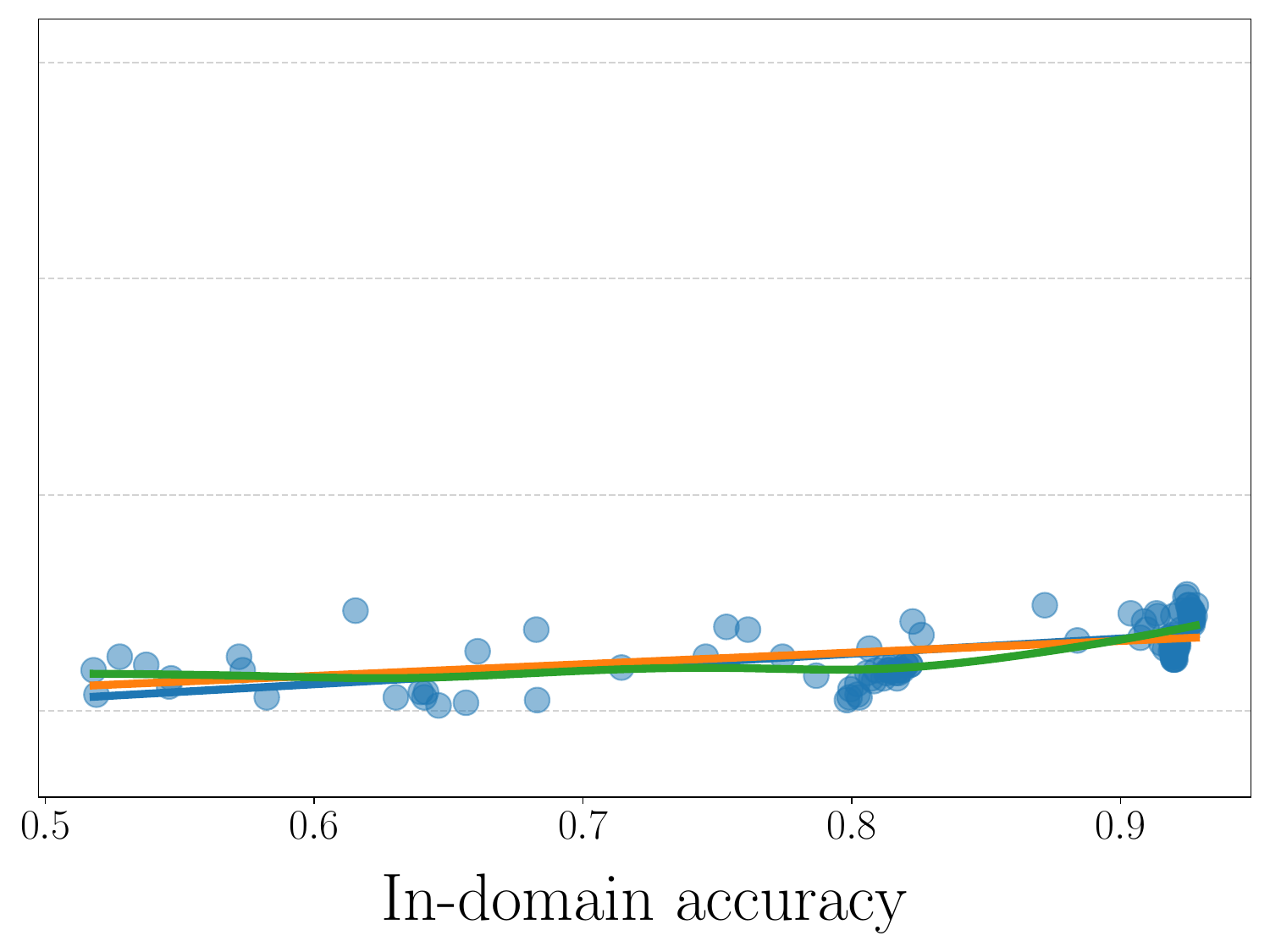}
        \scriptsize\colouredanli
    \end{subfigure}%
    \begin{subfigure}[b]{0.14\textwidth}
        \centering
        \includegraphics[width=\linewidth]{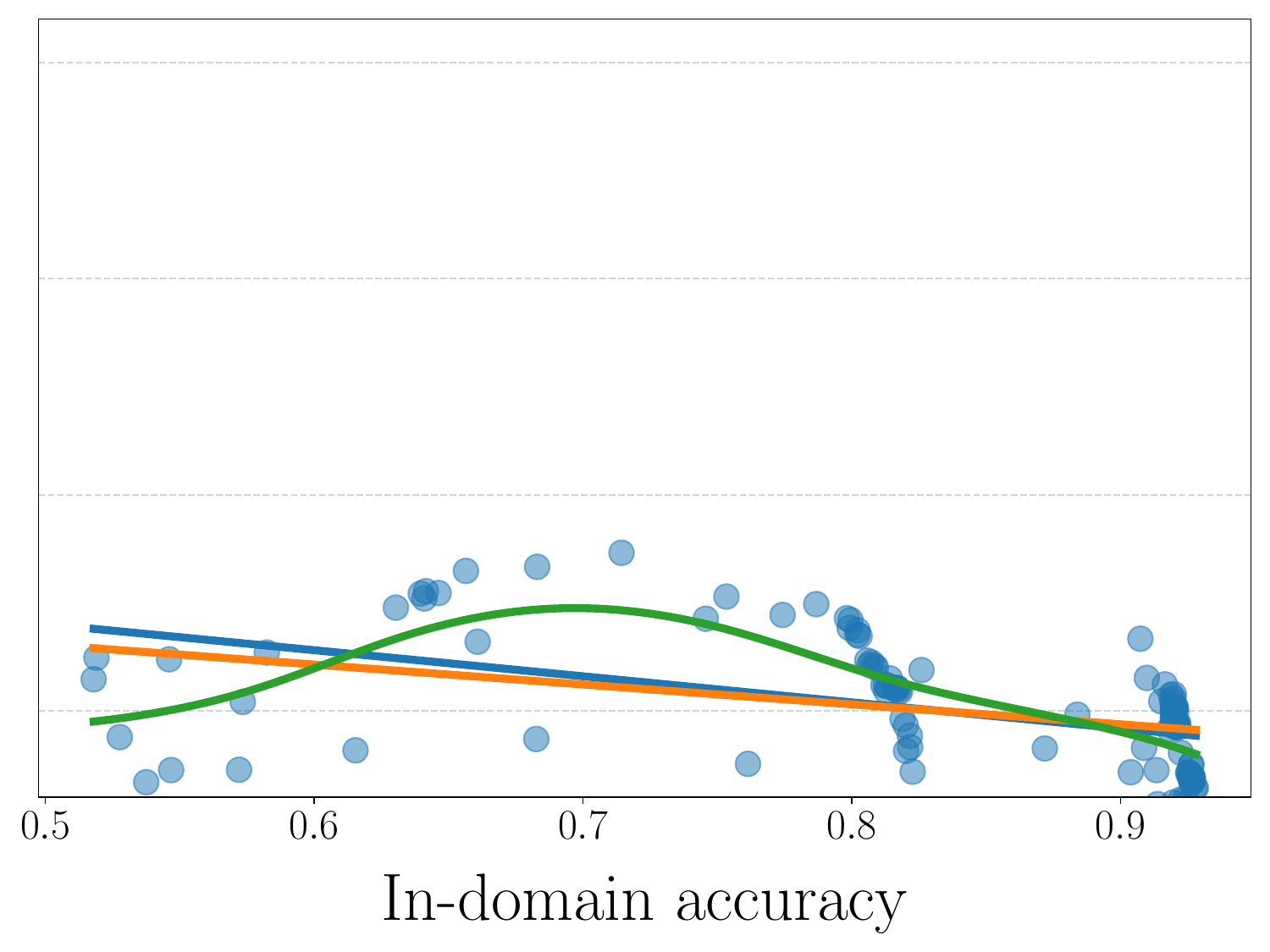}
        \scriptsize\colouredpaws
    \end{subfigure}
    \\[13pt]
    \begin{subfigure}[b]{0.14\textwidth}
        \centering
        \includegraphics[width=\linewidth]{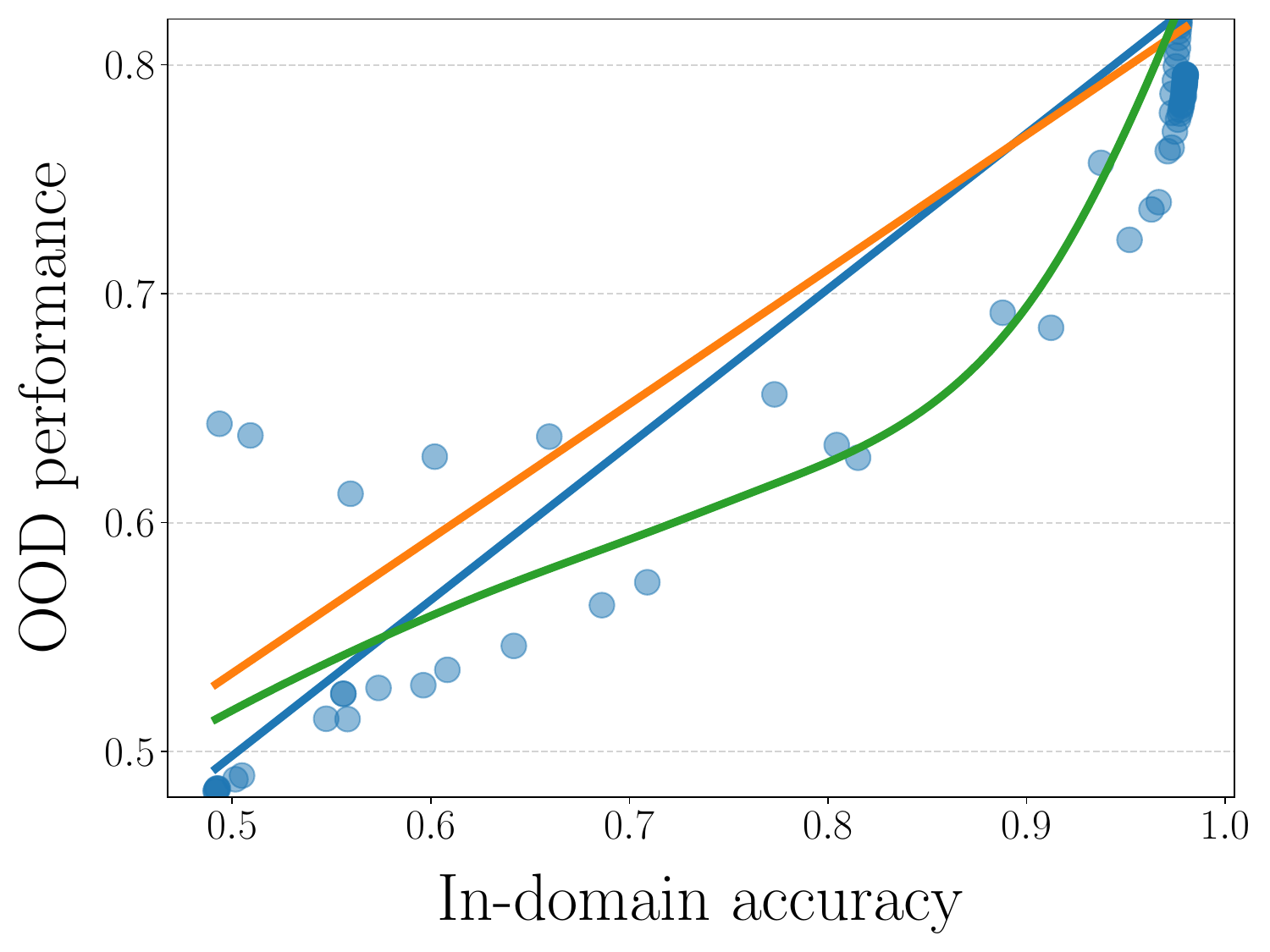}
    \end{subfigure}%
    \begin{subfigure}[b]{0.14\textwidth}
        \centering
        \includegraphics[width=\linewidth]{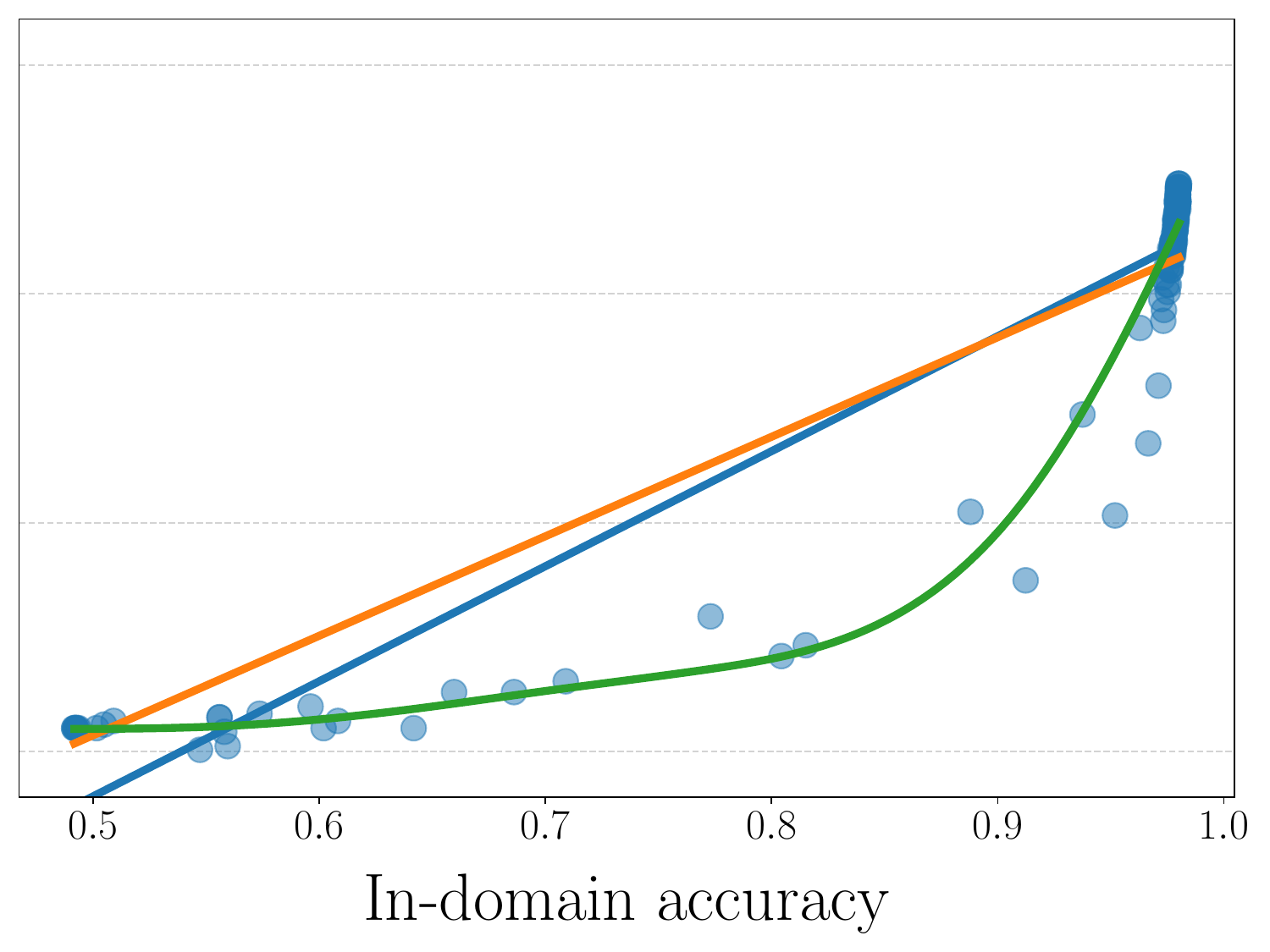}
    \end{subfigure}%
    \begin{subfigure}[b]{0.14\textwidth}
        \centering
        \includegraphics[width=\linewidth]{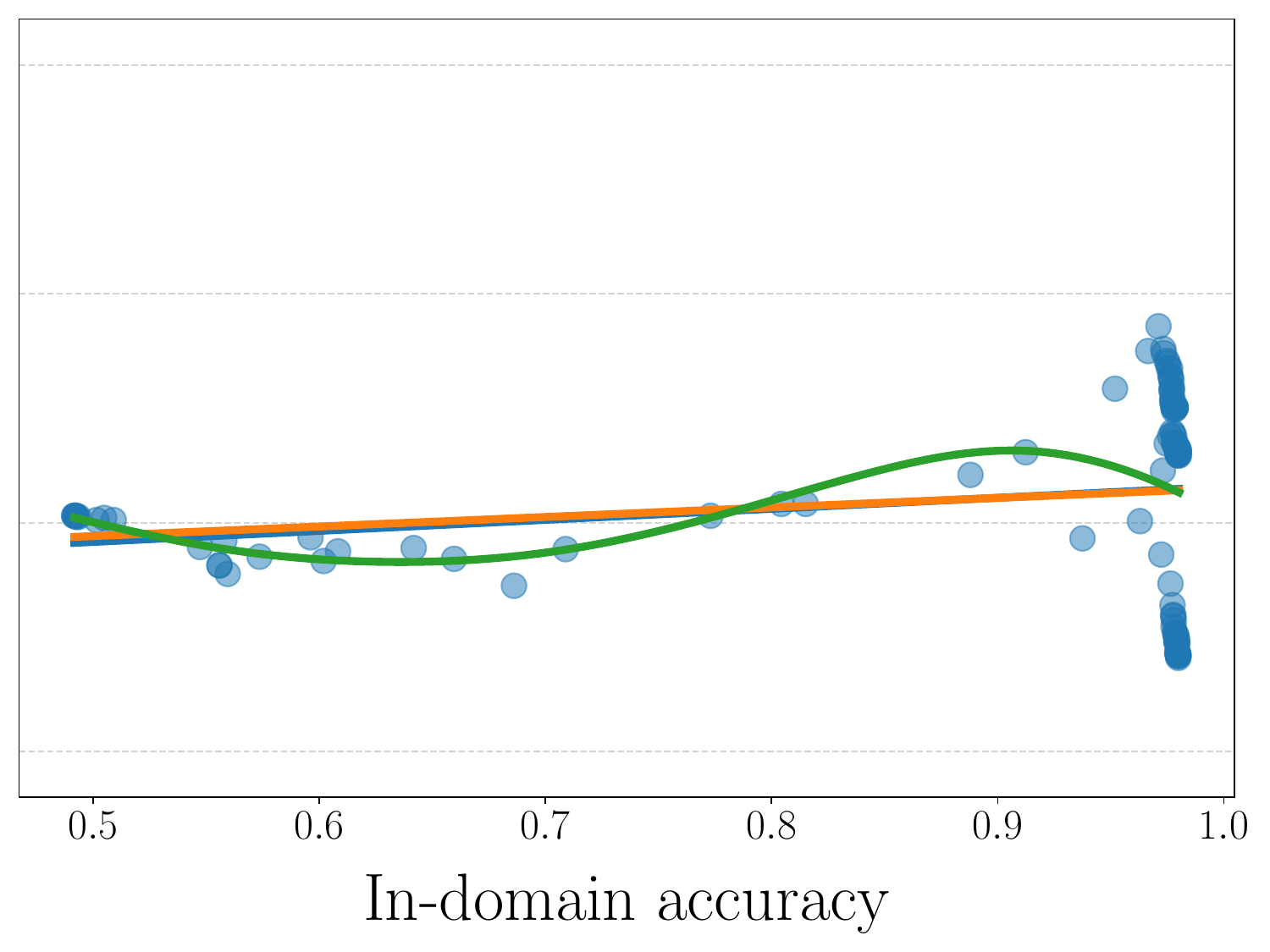}
    \end{subfigure}%
    \begin{subfigure}[b]{0.14\textwidth}
        \centering
        \includegraphics[width=\linewidth]{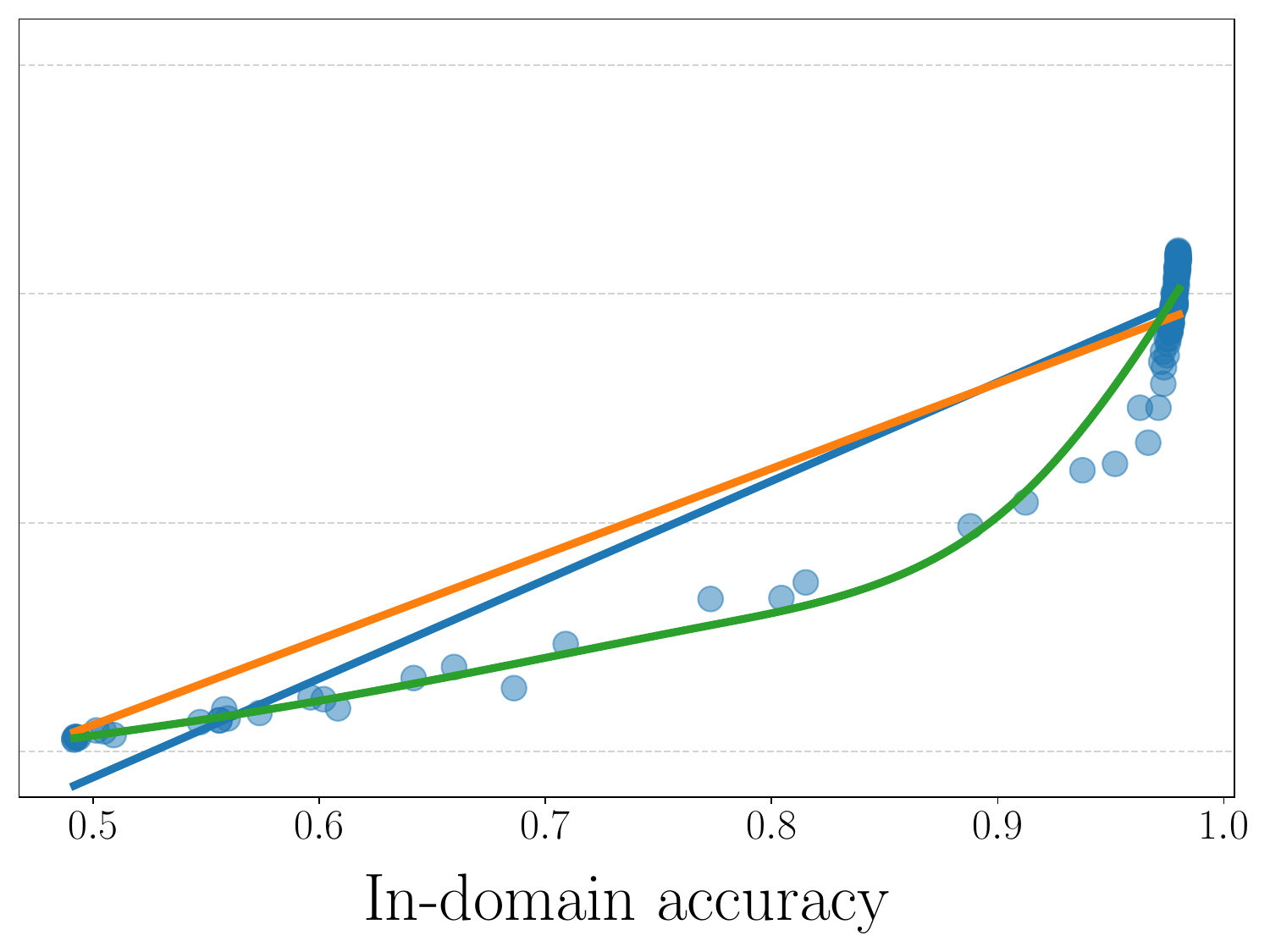}
    \end{subfigure}%
    \begin{subfigure}[b]{0.14\textwidth}
        \centering
        \includegraphics[width=\linewidth]{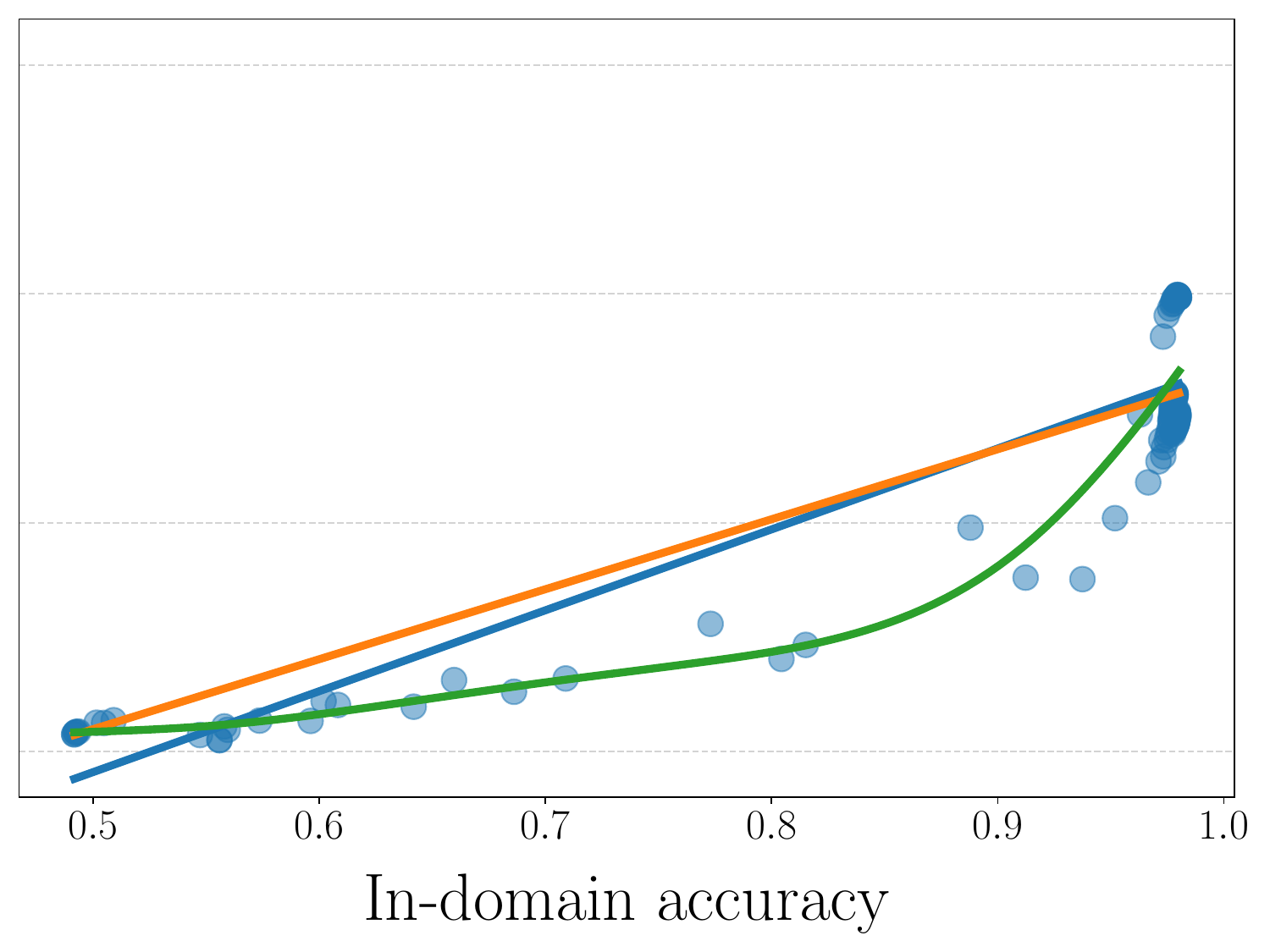}
    \end{subfigure}%
    \begin{subfigure}[b]{0.14\textwidth}
        \centering
        \includegraphics[width=\linewidth]{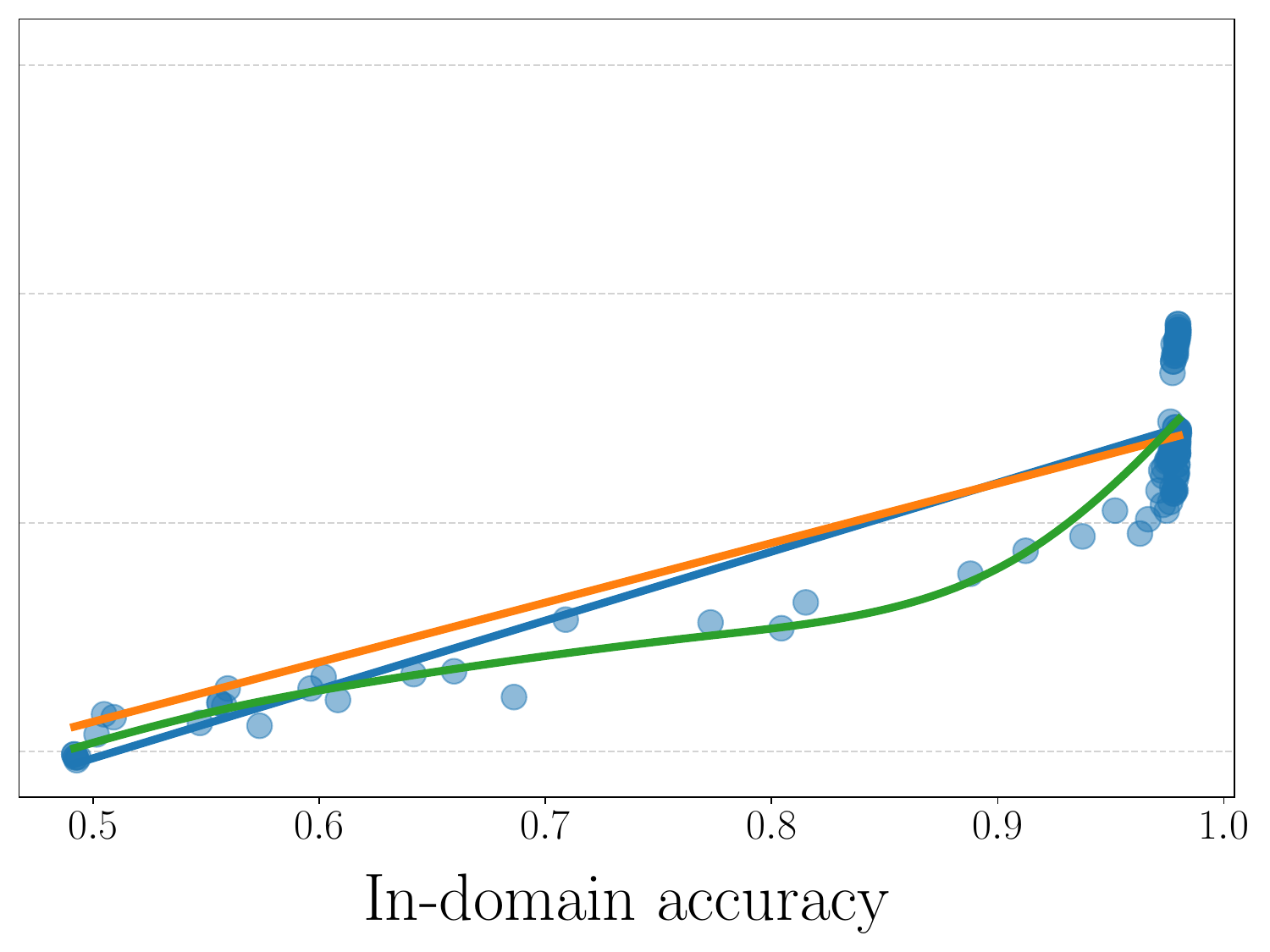}
        
    \end{subfigure}%
    \begin{subfigure}[b]{0.14\textwidth}
        \centering
        \includegraphics[width=\linewidth]{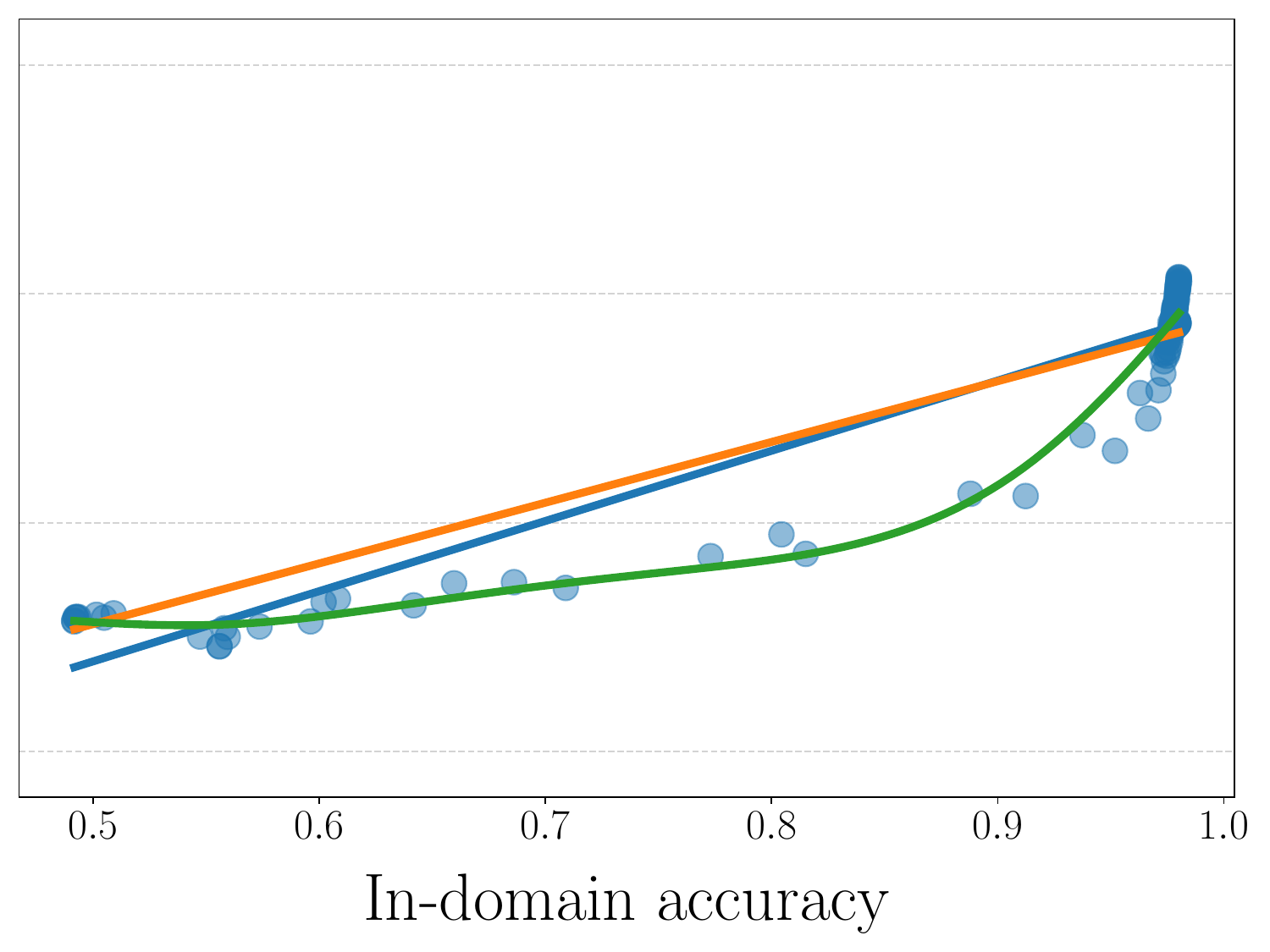}
    \end{subfigure}
    \begin{subfigure}[b]{0.14\textwidth}
        \centering
        \includegraphics[width=\linewidth]{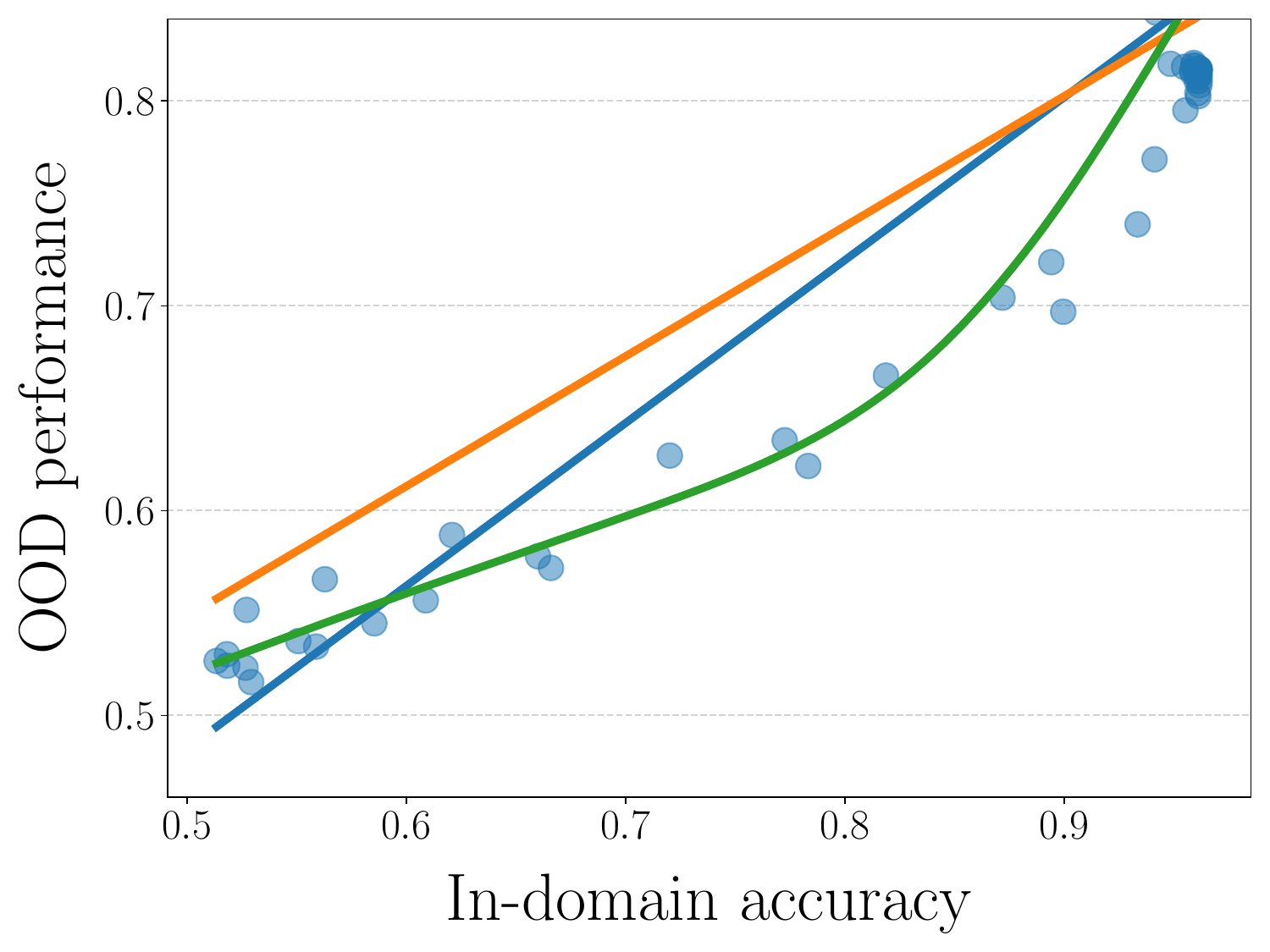}
        \scriptsize\colouredmnli
    \end{subfigure}%
    \begin{subfigure}[b]{0.14\textwidth}
        \centering
        \includegraphics[width=\linewidth]{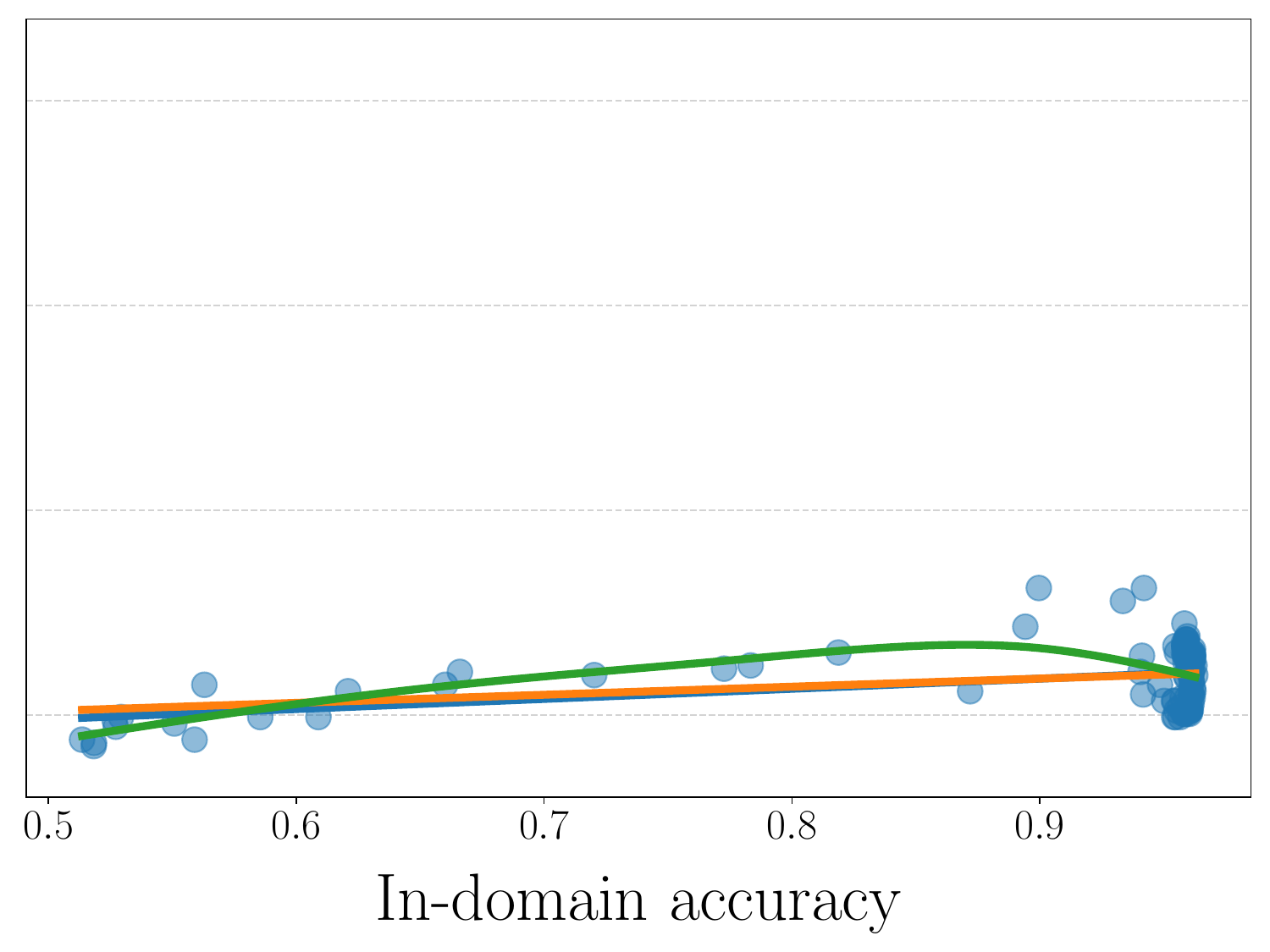}
        \scriptsize\colouredwnli
    \end{subfigure}%
    \begin{subfigure}[b]{0.14\textwidth}
        \centering
        \includegraphics[width=\linewidth]{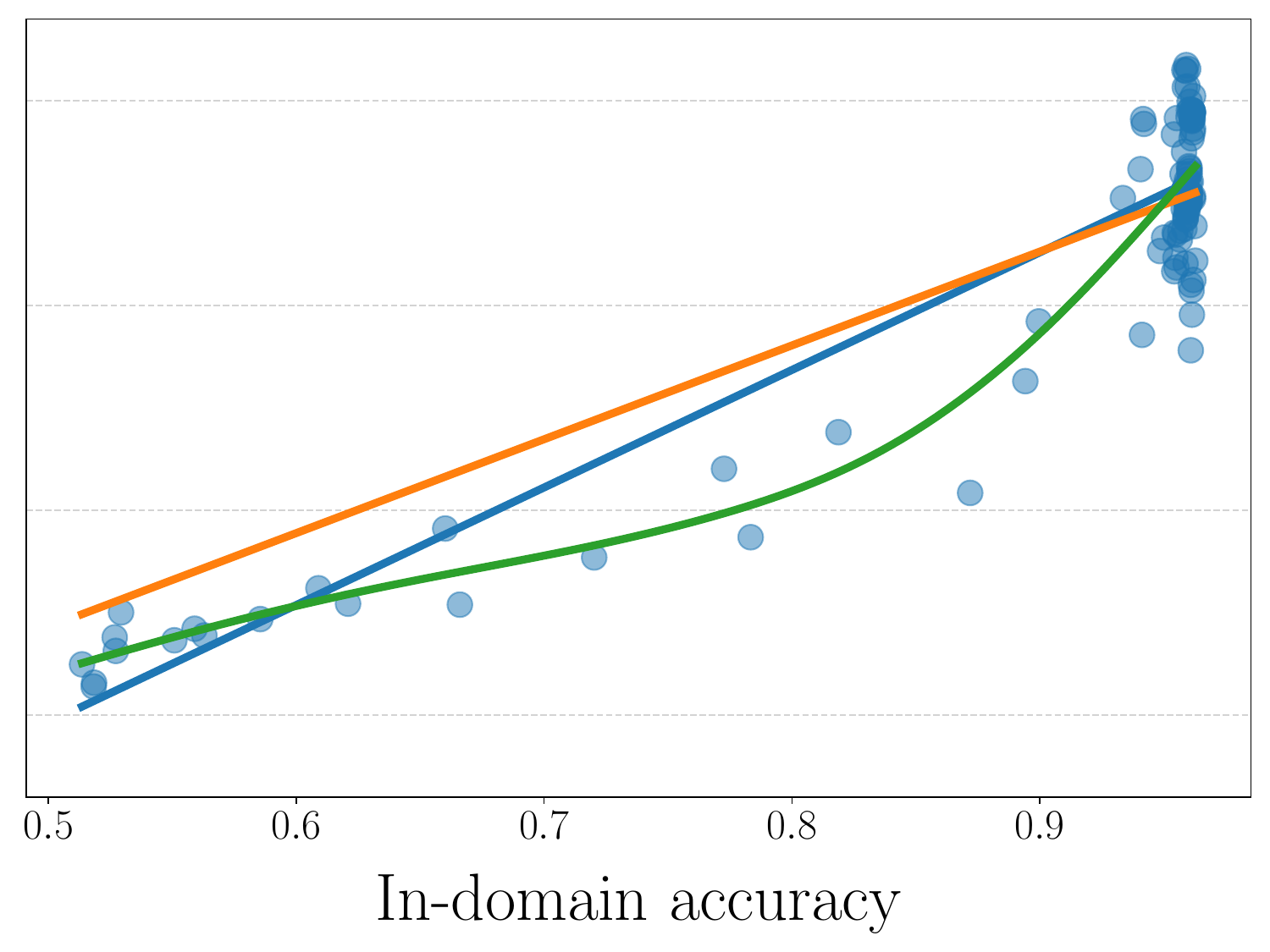}
        \scriptsize\colouredscitail
    \end{subfigure}%
    \begin{subfigure}[b]{0.14\textwidth}
        \centering
        \includegraphics[width=\linewidth]{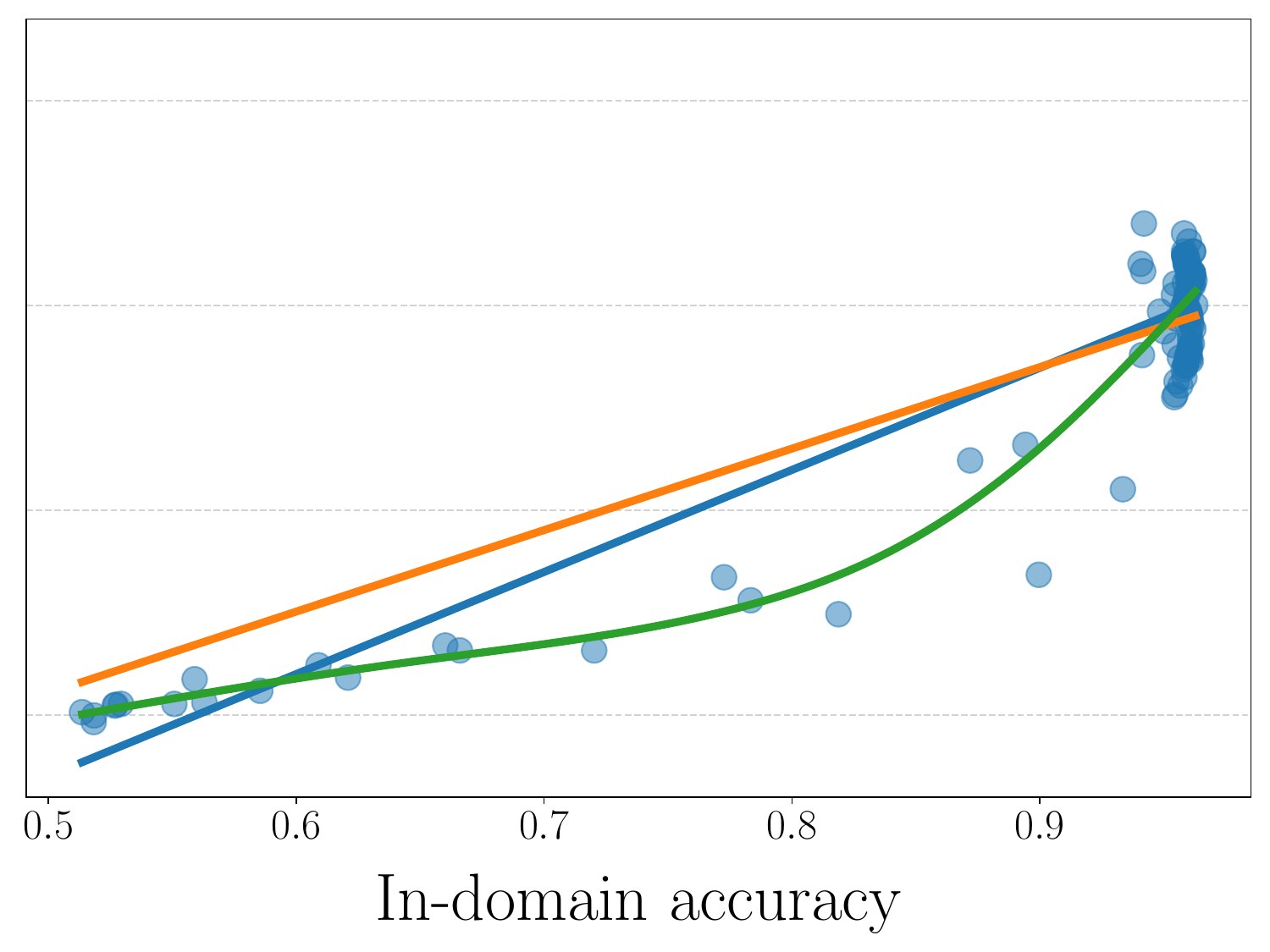}
        \scriptsize\colouredrte
    \end{subfigure}%
    \begin{subfigure}[b]{0.14\textwidth}
        \centering
        \includegraphics[width=\linewidth]{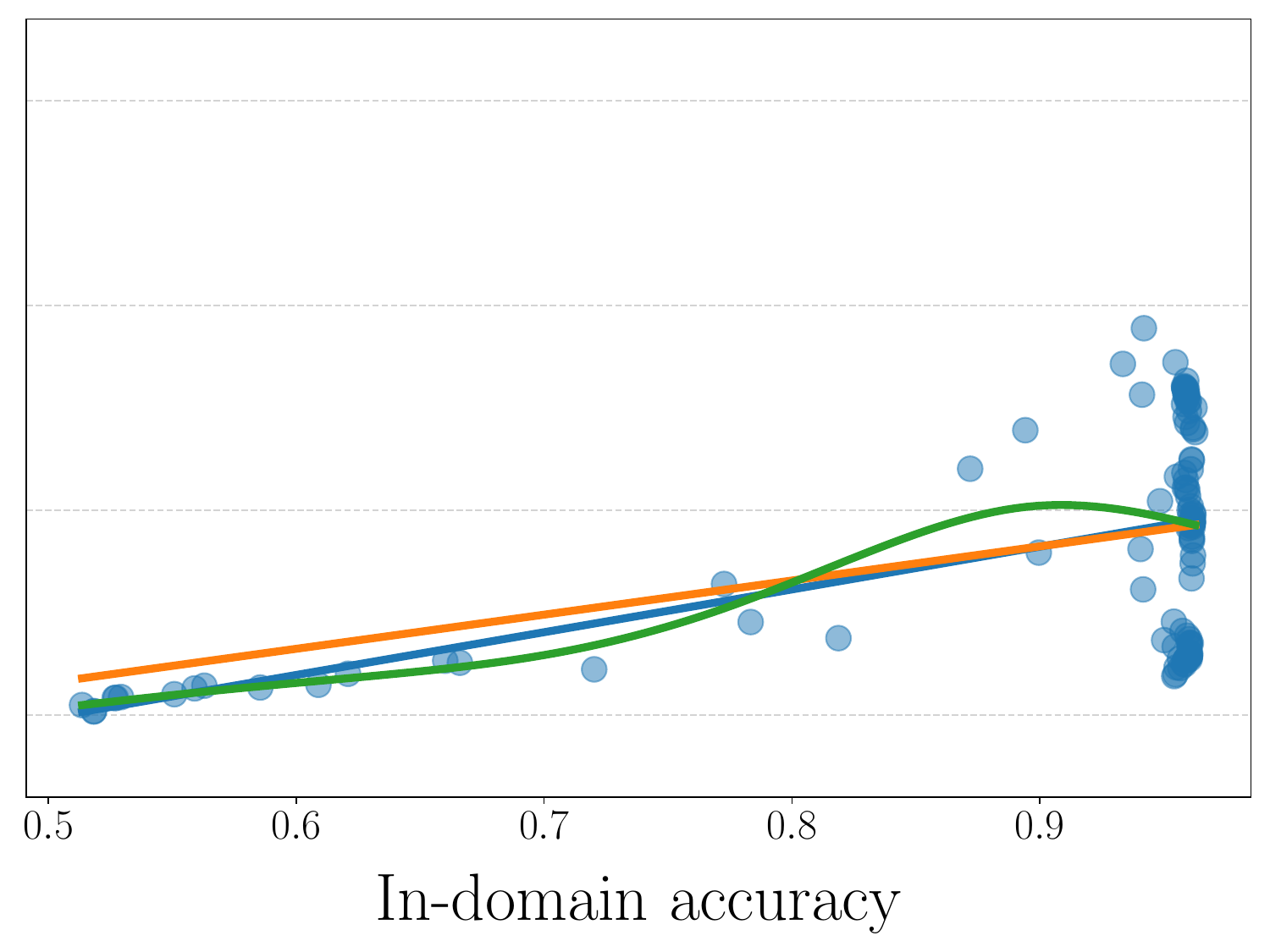}
        \scriptsize\colouredhans
    \end{subfigure}%
    \begin{subfigure}[b]{0.14\textwidth}
        \centering
        \includegraphics[width=\linewidth]{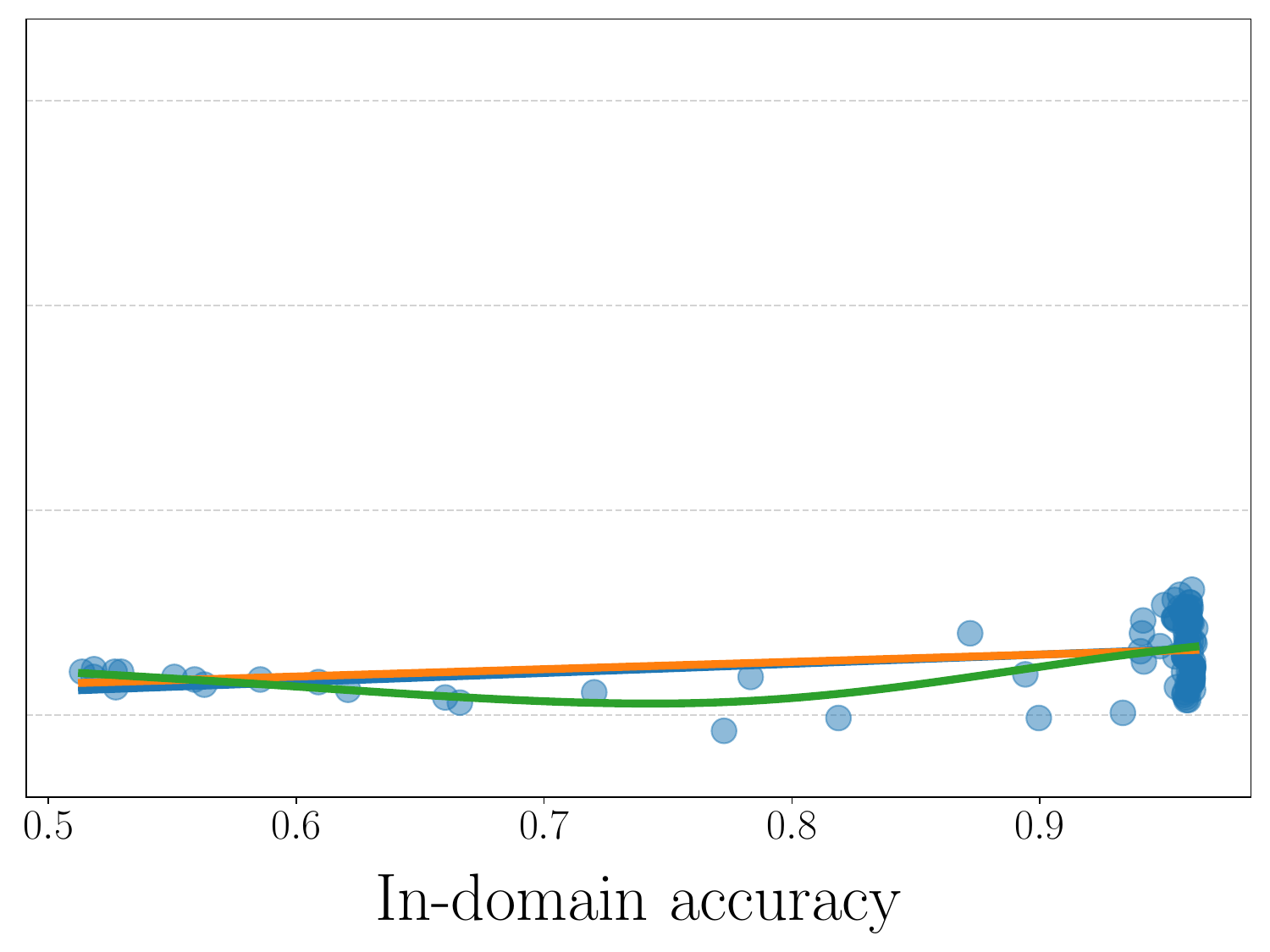}
        \scriptsize\colouredanli
    \end{subfigure}%
    \begin{subfigure}[b]{0.14\textwidth}
        \centering
        \includegraphics[width=\linewidth]{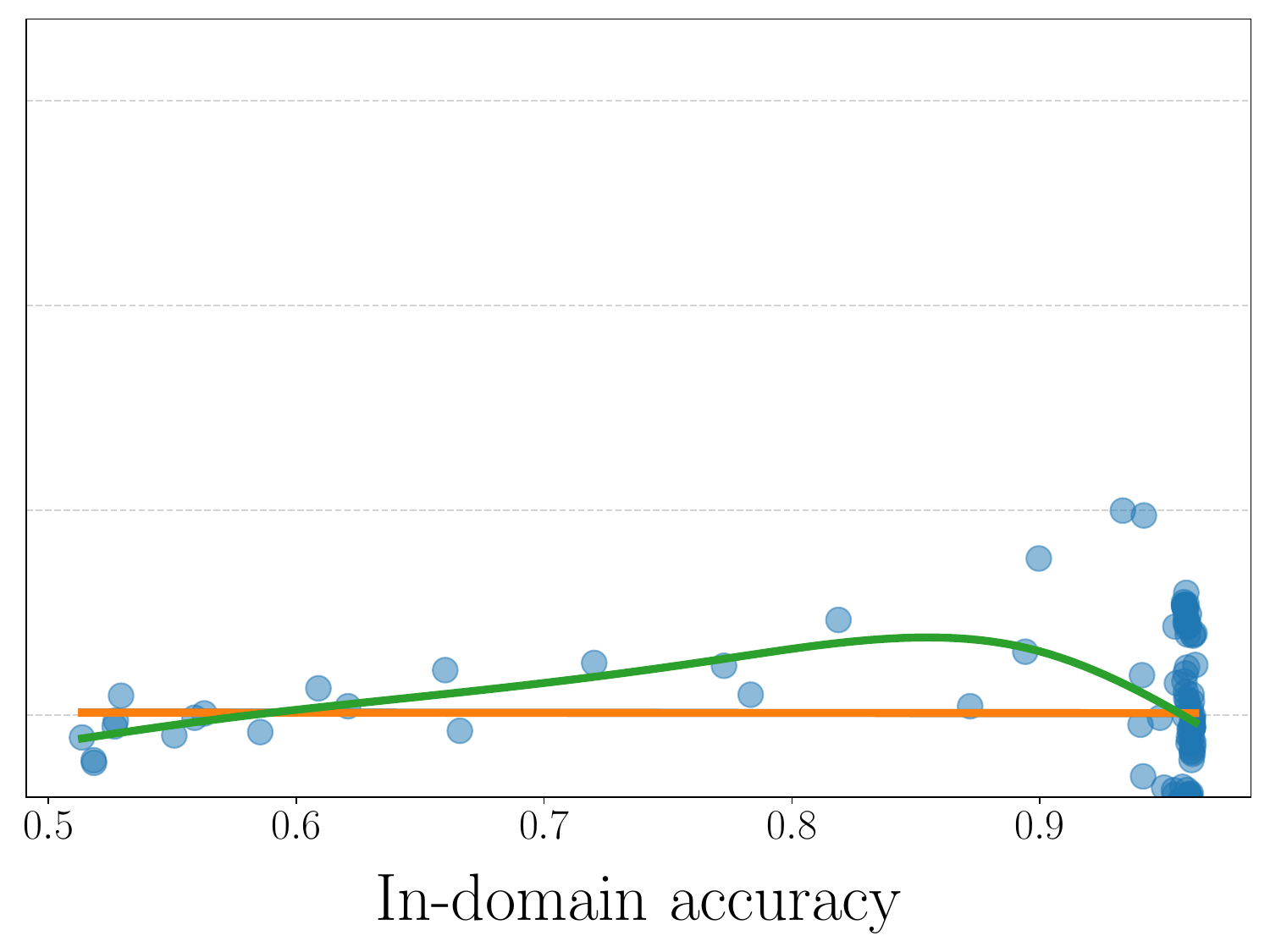}
        \scriptsize\colouredpaws
    \end{subfigure}
    \caption{Regressors trained to predict OOD performance for 128-shots models.
    Models were finetuned on MNLI (top) and SNLI (bottom). Results for \olmothirty on first and third rows, \optthirty on second and fourth rows. Legend: \colouredlinear, \colouredridge and \colouredgam.\looseness=-1}
    \label{fig:ood_regression_full_128shots}
\end{figure}

\subsection{Heatmaps with Partial OOD Correlations using Linear Regressors}

\begin{figure}[H]
    \centering
    \includegraphics[width=\textwidth]{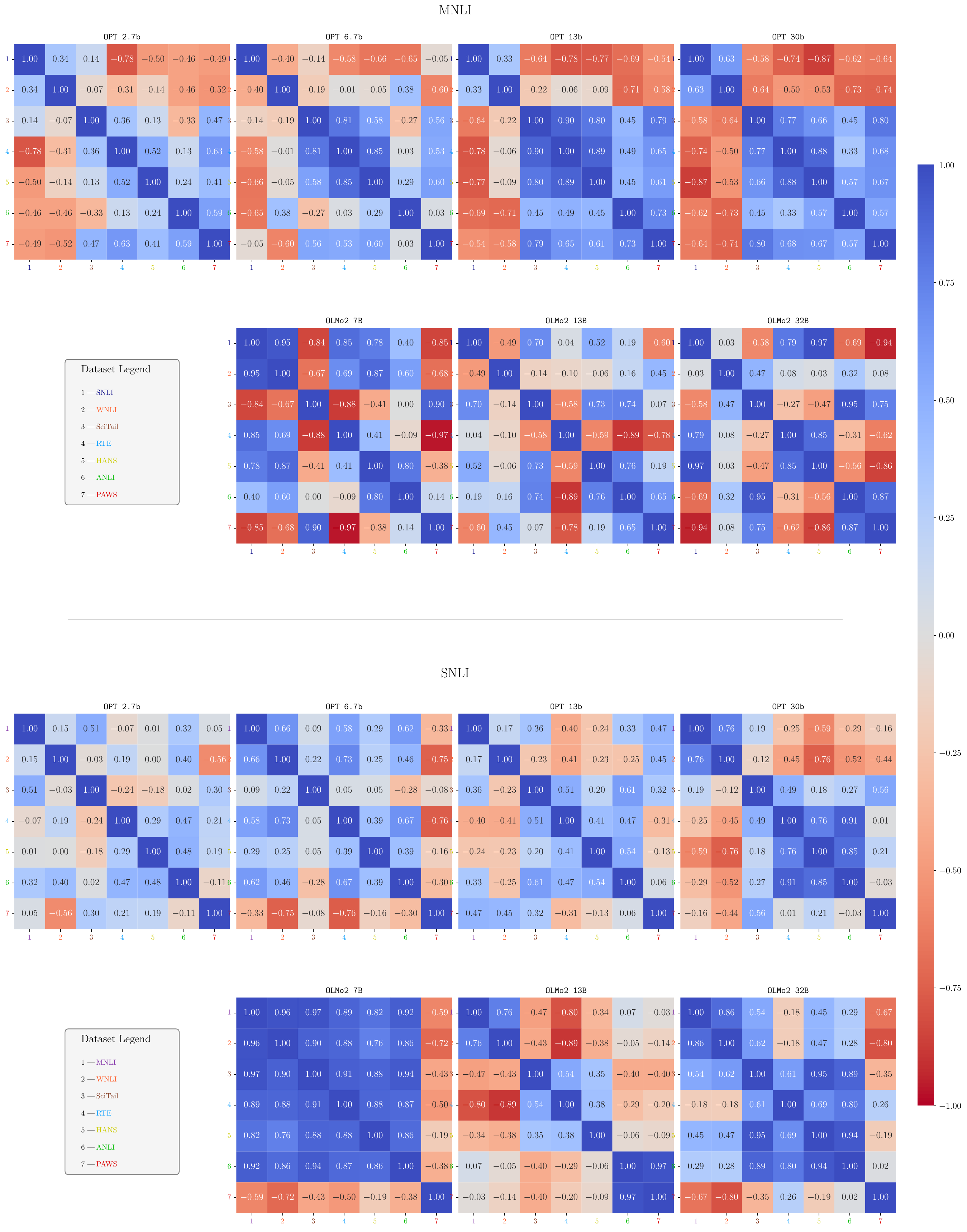}
    \caption{Partial correlations taken with linear regressors of \opt (first and third rows) and \olmo (second and forth rows) across model sizes (ordered from left to right) trained on MNLI (top) and SNLI (bottom). Models were fine-tuned with 128-shots.}
    \label{fig:corr_linear_4x4_128}
\end{figure}

\begin{figure}[H]
    \centering
    \includegraphics[width=\textwidth]{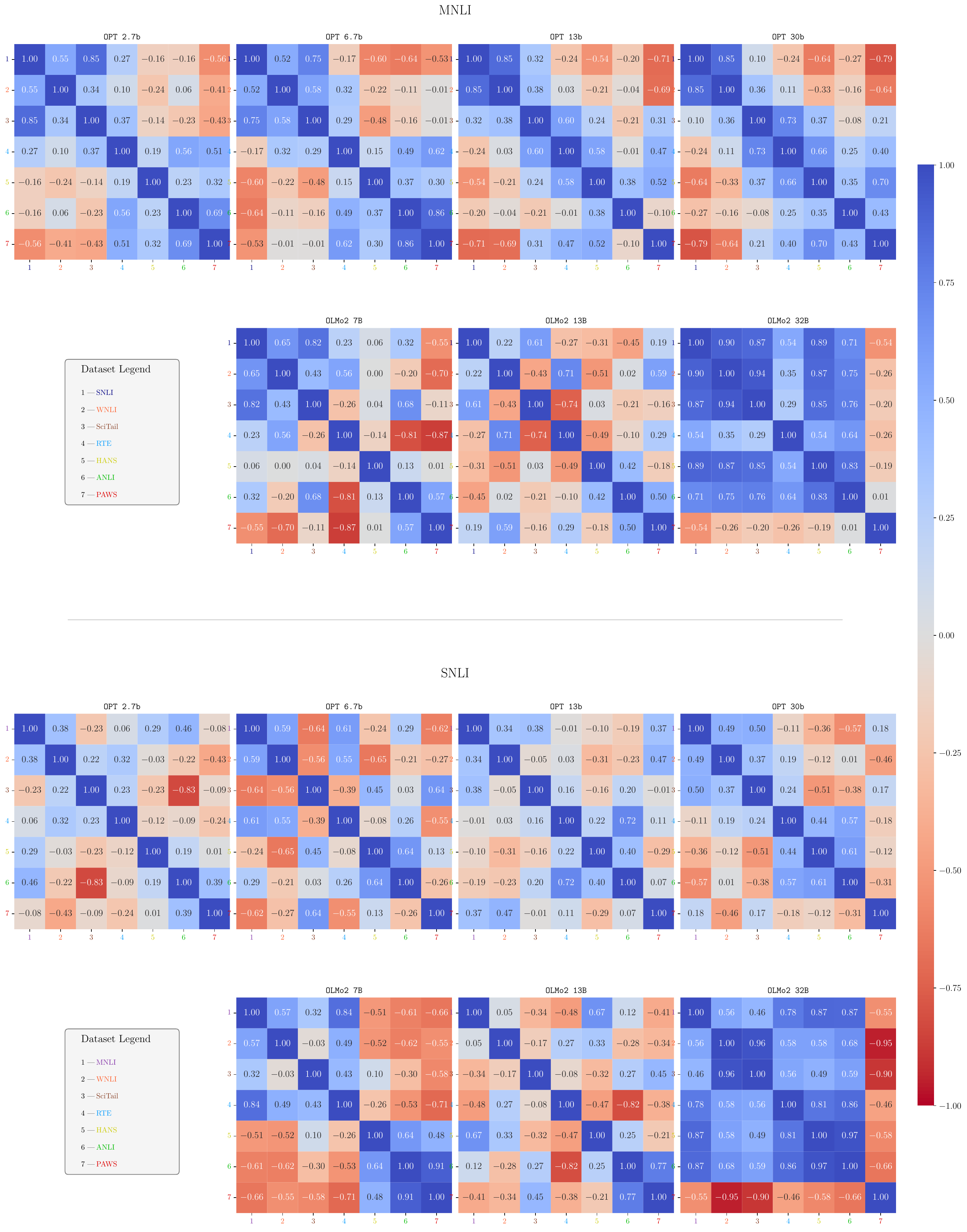}
    \caption{Partial correlations taken with linear regressors of \opt (first and third rows) and \olmo (second and forth rows) across model sizes (ordered from left to right) trained on MNLI (top) and SNLI (bottom). Models were fine-tuned with 64-shots.}
    \label{fig:corr_linear_4x4_64}
\end{figure}

\begin{figure}[H]
    \centering
    \includegraphics[width=\textwidth]{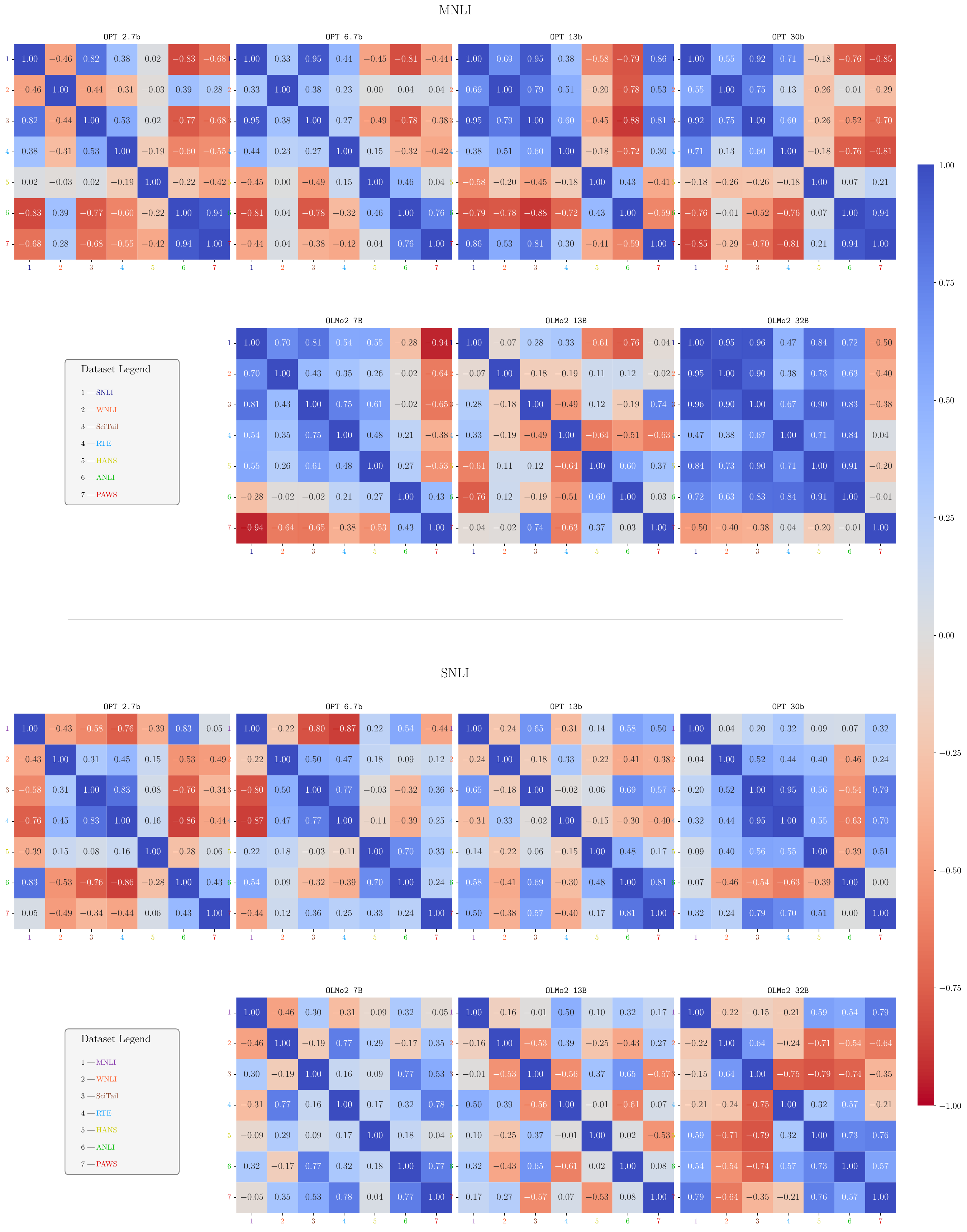}
    \caption{Partial correlations taken with linear regressors of \opt (first and third rows) and \olmo (second and forth rows) across model sizes (ordered from left to right) trained on MNLI (top) and SNLI (bottom). Models were fine-tuned with 32-shots.}
    \label{fig:corr_linear_4x4_32}
\end{figure}

\subsection{Heatmaps with Partial OOD Correlations using GAM Regressors}

\begin{figure}[H]
    \centering
    \includegraphics[width=\textwidth]{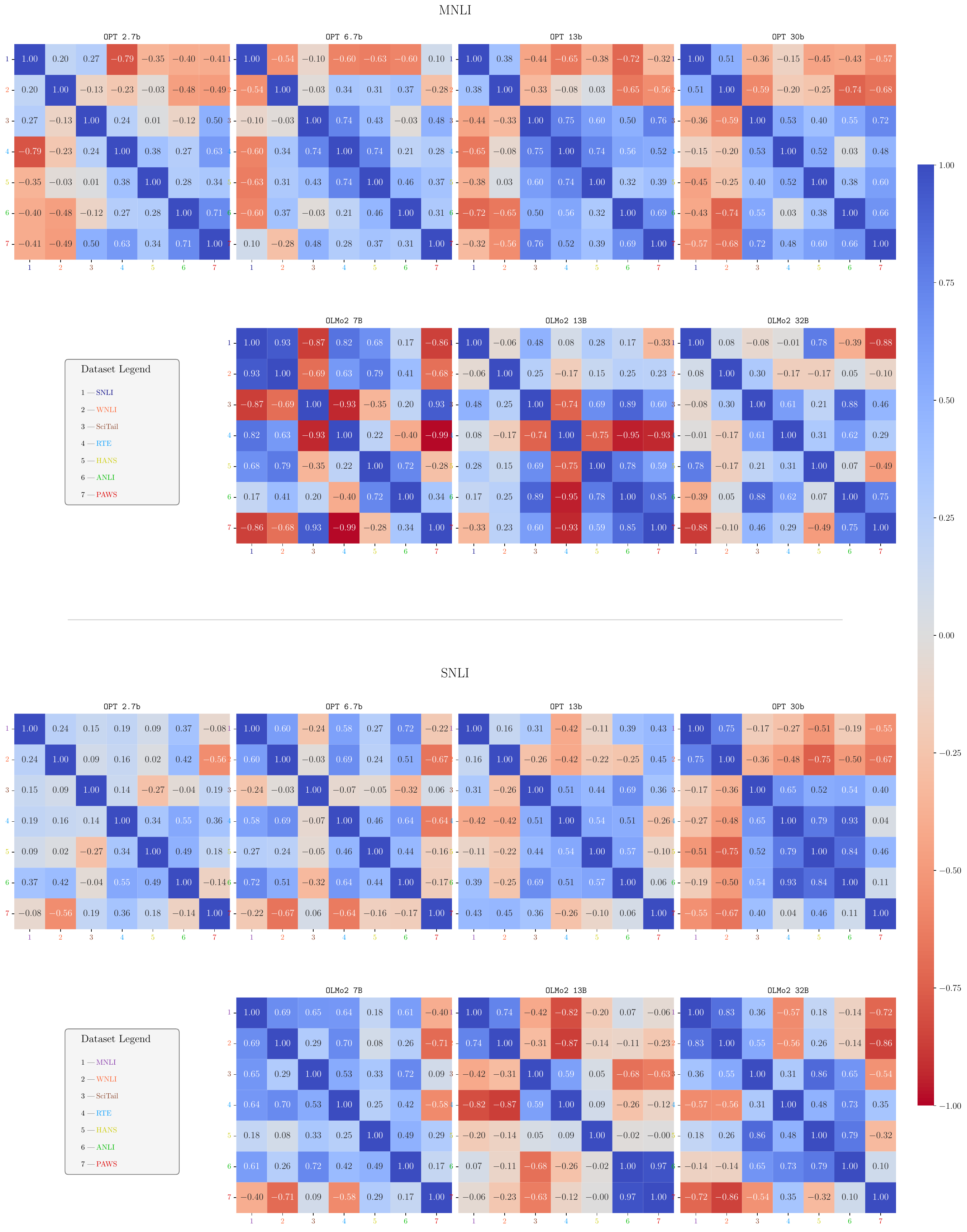}
    \caption{Partial correlations taken with GAM regressors of \opt (first and third rows) and \olmo (second and forth rows) across model sizes (ordered from left to right) trained on MNLI (top) and SNLI (bottom). Models were fine-tuned with 128-shots.}
    \label{fig:corr_gam_4x4_128}
\end{figure}

\begin{figure}[H]
    \centering
    \includegraphics[width=\textwidth]{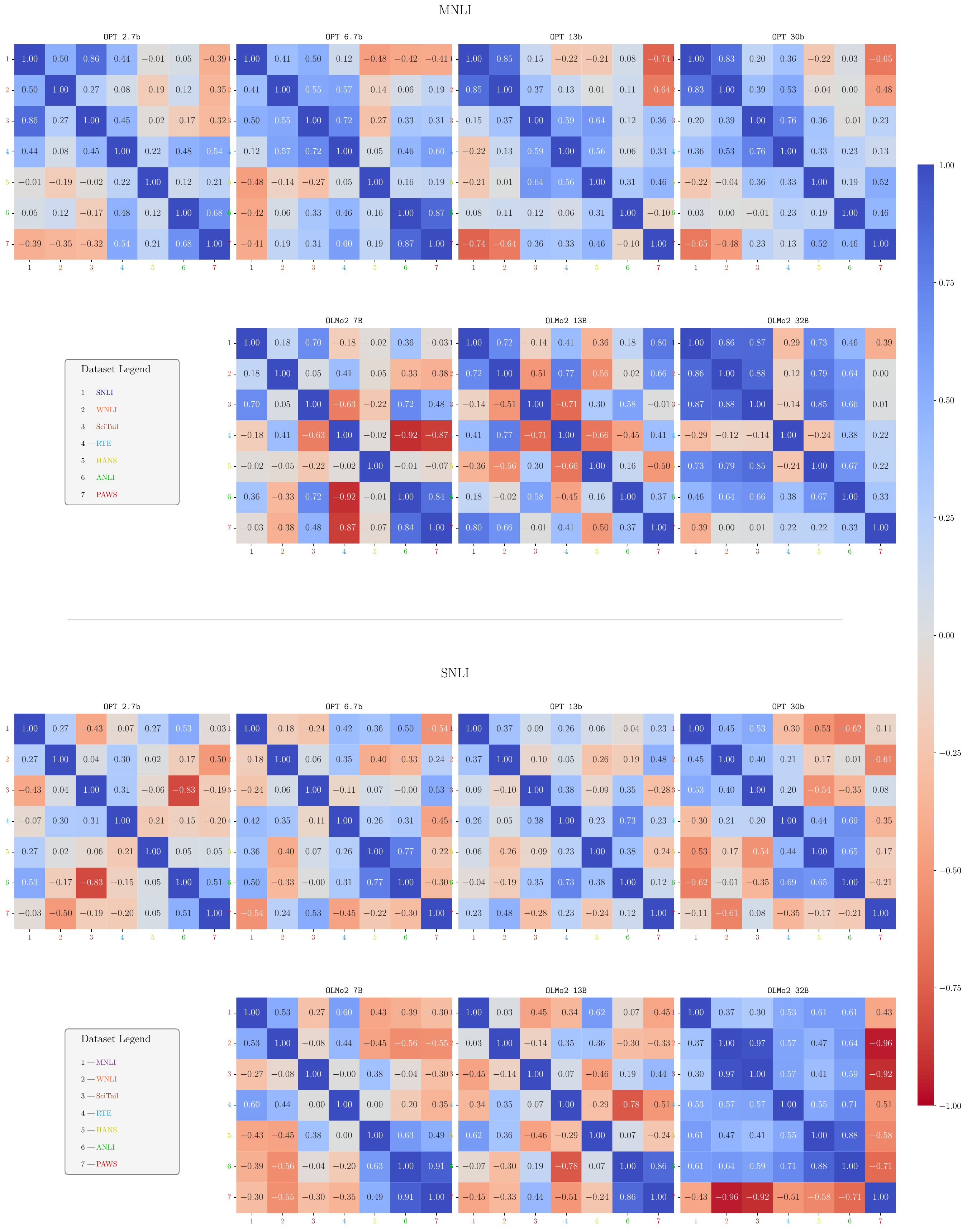}
    \caption{Partial correlations taken with GAM regressors of \opt (first and third rows) and \olmo (second and fourth rows) across model sizes (ordered from left to right) trained on MNLI (top) and SNLI (bottom). Models were fine-tuned with 64-shots.}
    \label{fig:corr_gam_4x4_64}
\end{figure}

\begin{figure}[H]
    \centering
    \includegraphics[width=\textwidth]{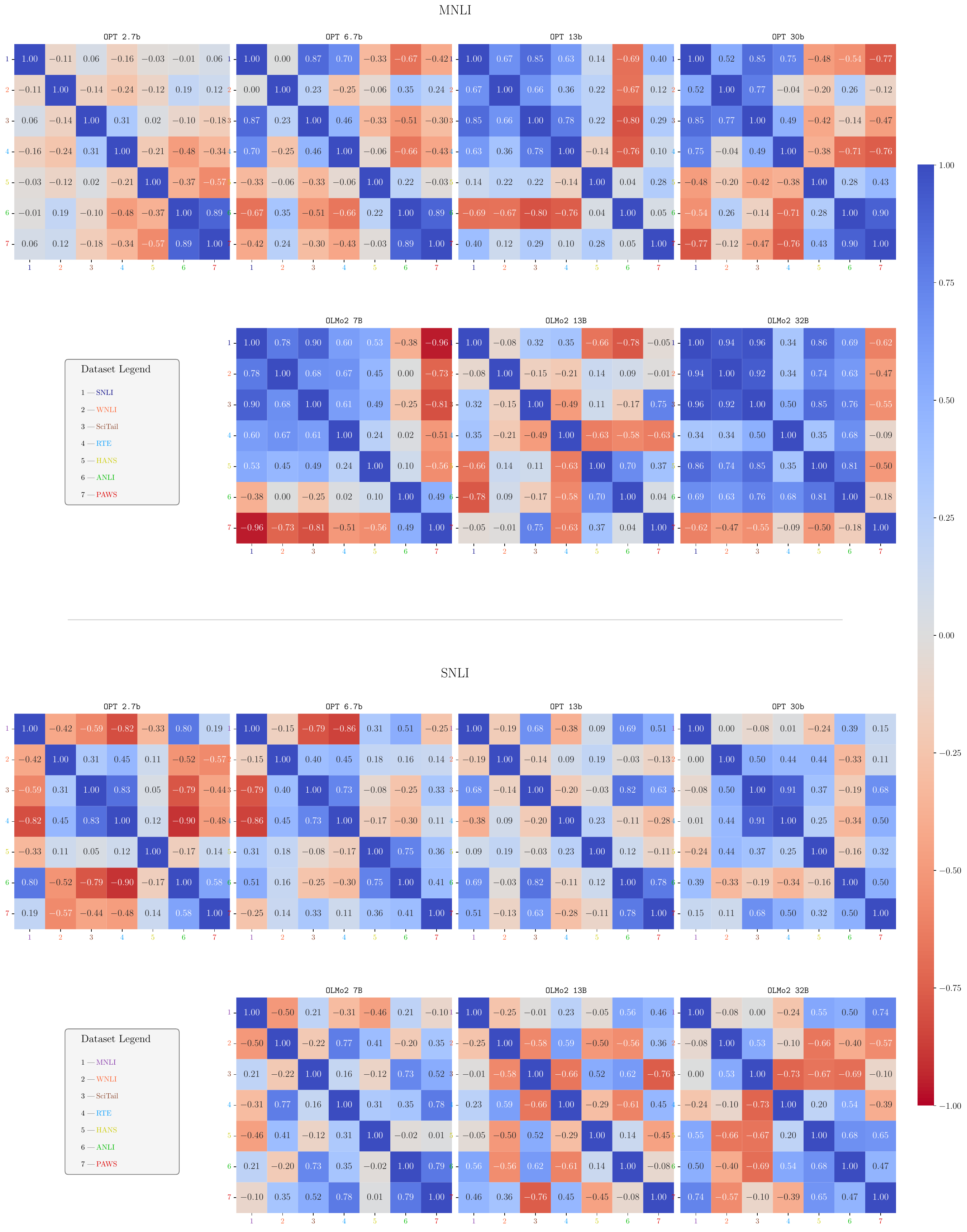}
    \caption{Partial correlations taken with GAM regressors of \opt (first and third rows) and \olmo (second and fourth rows)  across model sizes (ordered from left to right) trained on MNLI (top) and SNLI (bottom). Models were fine-tuned with 32-shots.}
    \label{fig:corr_gam_4x4_32}
\end{figure}

\subsection{Average Correlationg across Model Sizes}

\begin{figure}[H]
    \centering
    \includegraphics[width=\textwidth]{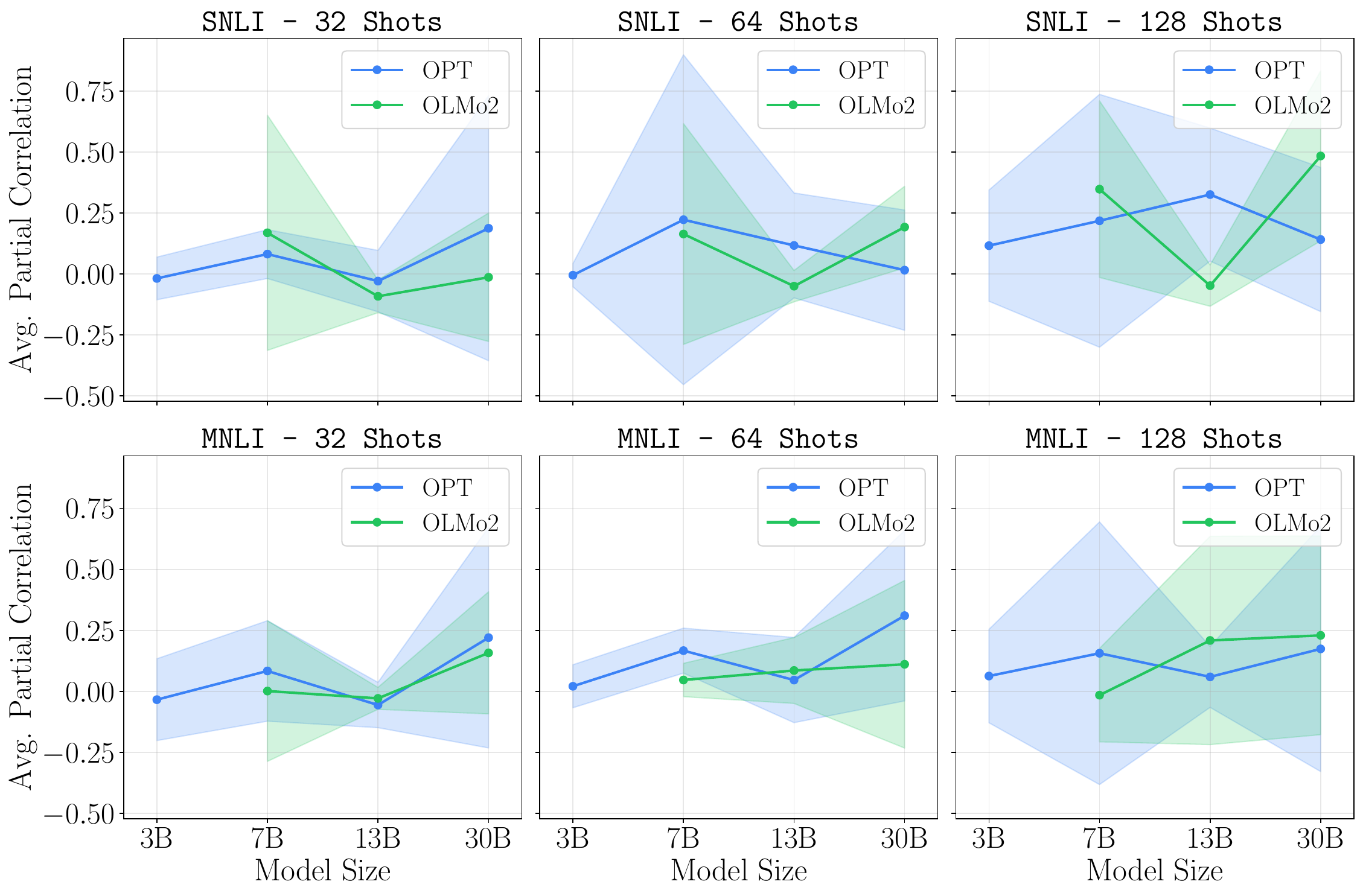}
    \caption{Average partial correlations  across model sizes between \opt and \olmo generalisation results taken with GAM regressors.}
    \label{fig:corr_across_sizes}
\end{figure}

\end{document}